%% file: voxopt.tex
\documentclass[10pt,twocolumn,letterpaper]{article}

\usepackage[pagenumbers]{cvpr} 

\usepackage{graphicx}
\usepackage{amsmath}
\usepackage{amssymb}
\usepackage{booktabs}
\usepackage{mathtools}
\usepackage[export]{adjustbox}
\usepackage[table]{xcolor}
\usepackage{subcaption}
\usepackage{microtype}
\usepackage{color}
\definecolor{turquoise}{cmyk}{0.65,0,0.1,0.3}
\definecolor{purple}{rgb}{0.65,0,0.65}
\definecolor{dark_green}{rgb}{0, 0.5, 0}
\definecolor{orange}{rgb}{0.8, 0.6, 0.2}
\definecolor{red}{rgb}{0.8, 0.2, 0.2}
\definecolor{darkred}{rgb}{0.6, 0.1, 0.05}
\definecolor{blueish}{rgb}{0.0, 0.3, .6}
\definecolor{light_gray}{rgb}{0.7, 0.7, .7}
\definecolor{pink}{rgb}{1, 0, 1}
\definecolor{greyblue}{rgb}{0.25, 0.25, 1}
\definecolor{green}{rgb}{0.72, 1, 0.7}
\definecolor{yellow}{rgb}{1, 1, 0.7}

\usepackage[pagebackref,breaklinks,colorlinks]{hyperref}
\usepackage{siunitx}

\usepackage[capitalize]{cleveref}
\crefname{section}{Sec.}{Secs.}
\Crefname{section}{Section}{Sections}
\Crefname{table}{Table}{Tables}
\crefname{table}{Tab.}{Tabs.}

\makeatletter
\newcommand{\printfnsymbol}[1]{%
        \textsuperscript{\@fnsymbol{#1}}%
}
\makeatother

\DeclarePairedDelimiter{\norm}{\lVert}{\rVert}

\makeatletter
\newcommand{\definetrim}[2]{%
  \define@key{Gin}{#1}[]{\setkeys{Gin}{trim=#2,clip}}%
}
\makeatother

\def\cvprPaperID{9439} 
\def\confName{CVPR}
\def\confYear{2022}

\begin{document}

\title{Plenoxels: Radiance Fields without Neural Networks}

\author{
Alex Yu$^*$
\qquad
Sara Fridovich-Keil$^*$
\qquad
Matthew Tancik
\qquad
Qinhong Chen\\
\qquad
Benjamin Recht
\qquad
Angjoo Kanazawa 
\vspace{1em}
\\
UC Berkeley
}
\maketitle

\begin{abstract}
We introduce Plenoxels (plenoptic voxels), a system for photorealistic view synthesis. Plenoxels represent a scene as a sparse 3D grid with spherical harmonics. This representation can be optimized from calibrated images via gradient methods and regularization without any neural components. On standard, benchmark tasks, Plenoxels are optimized two orders of magnitude faster than Neural Radiance Fields with no loss in visual quality. 
For video and code, please see \url{https://alexyu.net/plenoxels}.

\end{abstract}

{\let\thefootnote\relax\footnote{* Authors contributed equally to this work. }}\par

\section{Introduction}
\label{sec:intro}

A recent body of research has capitalized on implicit, coordinate-based neural networks as the 3D representation to optimize 3D volumes from calibrated 2D image supervision. In particular, Neural Radiance Fields (NeRF)~\cite{mildenhall2020nerf} demonstrated photorealistic novel viewpoint rendering, capturing scene geometry as well as view-dependent effects. This impressive quality, however, requires extensive computation time for both training and rendering, with training lasting more than a day and rendering requiring ~30 seconds per frame, on a single GPU. Multiple subsequent papers~\cite{yu2021plenoctrees, reiser2021kilonerf, rebain2020derf, liu2021neural, garbin2021fastnerf, hedman2021baking} reduced this computational cost, particularly for rendering, but single GPU training still requires multiple hours, a bottleneck that limits the practical application of photorealistic volumetric reconstruction.

\input{figures_tex/teaser}

In this paper, we show that we can train a radiance field from scratch, without neural networks, while maintaining NeRF quality and reducing optimization time by two orders of magnitude. We provide a custom CUDA \cite{cuda} implementation that capitalizes on the model simplicity to achieve substantial speedups. Our typical optimization time on a single Titan RTX GPU is 11 minutes on bounded scenes (compared to roughly 1 day for NeRF, more than a $100\times$ speedup) and 27 minutes on unbounded scenes (compared to roughly 4 days for NeRF++~\cite{zhang2020nerf}, again more than a $100\times$ speedup). Although our implementation is not optimized for fast rendering, we can render novel viewpoints at interactive rates $15$ fps. If faster rendering is desired, our optimized Plenoxel model can be converted into a PlenOctree \cite{yu2021plenoctrees}.

Specifically, we propose an explicit volumetric representation, based on a view-dependent sparse voxel grid without any neural networks. Our model can render photorealistic novel viewpoints and be optimized end-to-end from calibrated 2D photographs, using the differentiable rendering loss on training views as well as a total variation regularizer. We call our model Plenoxel for plenoptic volume elements, as it consists of a sparse voxel grid in which each voxel stores opacity and spherical harmonic coefficients. These coefficients are interpolated to model the full plenoptic function continuously in space. To achieve high resolution on a single GPU, we prune empty voxels and follow a coarse to fine optimization strategy. Although our core model is a bounded voxel grid, we can model unbounded scenes by using normalized device coordinates (for forward-facing scenes) or by surrounding our grid with multisphere images to encode the background (for $360^{\circ}$ scenes).

Our method reveals that photorealistic volumetric reconstruction can be approached using standard tools from inverse problems: a data representation, a forward model, a regularization function, and an optimizer. Our method shows that each of these components can be simple and state of the art results can still be achieved. Our experiments suggest the key element of Neural Radiance Fields is not the neural network but the differentiable volumetric renderer.

\section{Related Work}
\label{sec:references}


\paragraph{Classical Volume Reconstruction.}

We begin with a brief overview of classical methods for volume reconstruction, focusing on those which find application in our work. In particular, the most common classical methods for volume rendering are voxel grids~\cite{choy20163dr2n2, seitz1997, seitz2000, kar2017learning, tulsiani2017multiview, sitzmann2019deepvoxels, lombardi2019neural} and multi-plane images (MPIs)~\cite{Szeliski2004StereoMW, Penner2017Soft3R, zhou2018stereo, srinivasan2019pushing, mildenhall2019local, wizadwongsa2021nex}. Voxel grids are capable of representing arbitrary topologies but can be memory limited at high resolution. One approach for reducing the memory requirement for voxel grids is to encode hierarchical structure, for instance using octrees~\cite{riegler2017octnet, hane2017hierarchical, tatarchenko2017octree, wang2017} (see~\cite{Knoll_asurvey} for a survey); we use an even simpler sparse array structure.
Using these grid-based representations combined with some form of interpolation produces a continuous representation that can be arbitrarily resized using standard signal processing methods (see~\cite{oppenheim} for reference). 
This combination of sparsity and interpolation enables even a simple grid-based model to represent 3D scenes at high resolution without prohibitive memory requirements.
We combine this classical sampling and interpolation paradigm with the forward volume rendering formula introduced by Max~\cite{max1995} (based on work from Kajiya and Von Herzen~\cite{kajiya1984} and used in NeRF) to directly optimize a 3D model from indirect 2D observations. 
We further extend these classical approaches by modeling view dependence, which we accomplish by optimizing spherical harmonic coefficients for each color channel at each voxel. Spherical harmonics are a standard basis for functions over the sphere, and have been used previously to represent view dependence \cite{ramamoorthi2001relationship, sloan2002, basri2003, yu2021plenoctrees}.


\paragraph{Neural Volume Reconstruction.}
Recently, dramatic improvements in neural volume reconstruction have renewed interest in this direction. 
Neural implicit representations were first used to model occupancy~\cite{mescheder2019occupancy, chen2019learning, martel2021acorn} and signed distance to an object's surface~\cite{park2019deepsdf, takikawa2021neural}, and perform novel view synthesis  from 3D point clouds~\cite{aliev2020neural,ruckert2021adop, wiles2020synsin, Lassner_pulsar}.
Several papers extended this idea of neural implicit 3D modeling to model a scene using only calibrated 2D image supervision via a differentiable volume rendering formulation~\cite{sitzmann2019deepvoxels, sitzmann2020scene, lombardi2019neural, mildenhall2020nerf}.

NeRF~\cite{mildenhall2020nerf} in particular uses a differentiable volume rendering formula to train a coordinate-based multilayer perceptron (MLP) to directly predict color and opacity from 3D position and 2D viewing direction. NeRF produces impressive results but requires several days for full training, and about half an minute to render a full image, because every rendered pixel requires evaluating the coordinate-based MLP at hundreds of sample locations along the corresponding ray. Many papers have since extended the capabilities of NeRF, including modeling the background in $360^{\circ}$ views~\cite{zhang2020nerf} and incorporating anti-aliasing for multiscale rendering~\cite{barron2021mipnerf}. We extend our Plenoxel method to unbounded $360^{\circ}$ scenes using a background model inspired by NeRF++ \cite{zhang2020nerf}.

Of these methods, Neural Volumes \cite{lombardi2019neural} is the most similar to ours in that it uses a voxel grid with interpolation, but optimizes this grid through a convolutional neural network and applies a learned warping function to improve the effective resolution (of a $128^3$ grid). We show that the voxel grid can be optimized directly and high resolution can be achieved by pruning and coarse to fine optimization, without any neural networks or warping functions.


\input{figures_tex/overview}

\paragraph{Accelerating NeRF.}

In light of the substantial computational requirements of NeRF for both training and rendering, many recent papers have proposed methods to improve efficiency, particularly for rendering. 
Among these methods are many that achieve speedup by subdividing the 3D volume into regions that can be processed more efficiently \cite{rebain2020derf, liu2021neural}. 
Other speedup approaches have focused on a range of computational and pre- or post-processing methods to remove bottlenecks in the original NeRF formulation.
JAXNeRF~\cite{jaxnerf2020github}, a JAX~\cite{jax2018github} reimplementation of NeRF offers a speedup for both training and rendering via parallelization across many GPUs or TPUs.
AutoInt~\cite{lindell2021autoint} restructures the coordinate-based MLP to compute ray integrals exactly, for more than $10\times$ faster rendering with a small loss in quality. 
Learned Initializations~\cite{tancik2021learned} employs meta-learning on many scenes to start from a better MLP initialization, for both $>10\times$ faster training and better priors when per-scene data is limited.
Other methods \cite{donerf2021, piala2021terminerf, kellnhofer2021neural} achieve speedup by predicting a surface or sampling near the surface, reducing the number of samples necessary for rendering each ray.

Another approach is to pretrain a NeRF (or similar model) and then extract it into a different data structure that can support fast inference~\cite{garbin2021fastnerf, hedman2021baking, reiser2021kilonerf, yu2021plenoctrees}.
In particular, PlenOctrees~\cite{yu2021plenoctrees} extracts a NeRF variant into a sparse voxel grid in which each voxel represents view-dependent color using spherical harmonic coefficients. Because the extracted PlenOctree can be further optimized, this method can speed up training by roughly $3\times$, and because it uses an efficient GPU octree implementation without any MLP evaluations, it achieves $>3000\times$ rendering speedup. 
Our method extends PlenOctrees to perform end-to-end optimization of a sparse voxel representation with spherical harmonics, offering much faster training (two orders of magnitude speedup compared to NeRF). Our Plenoxel model is a generalization of PlenOctrees to support sparse plenoptic voxel grids of arbitrary resolution (not necessary powers of two) with the ability to perform trilinear interpolation, which is easier to implement with this sparse voxel structure. 

\section{Method}
\label{sec:methods}

Our model is a sparse voxel grid in which each occupied voxel corner stores a scalar opacity $\sigma$ and a vector of spherical harmonic (SH) coefficients for each color channel. From here on we refer to this representation as Plenoxel. The opacity and color at an arbitrary position and viewing direction are determined by trilinearly interpolating the values stored at the neighboring voxels and evaluating the spherical harmonics at the appropriate viewing direction. 
Given a set of calibrated images, we optimize our model directly using the rendering loss on training rays. Our model is illustrated in \cref{fig:overview} and described in detail below.

\subsection{Volume Rendering}
We use the same differentiable model for volume rendering as in NeRF, where the color of a ray is approximated by integrating over samples taken along the ray:
\begin{align}
\label{eq:max}
    \hat C(\textbf{r}) = \sum_{i=1}^N T_i \big(1 - \exp(-\sigma_i \delta_i)\big)\textbf{c}_i \\
\text{where } \hspace{2em}
T_i = \exp \left(-\sum_{j=1}^{i-1} \sigma_j \delta_j \right)
\end{align}
$T_i$ represents how much light is transmitted through ray $\textbf{r}$ to sample $i$ (versus contributed by preceding samples), $\left(1 - \exp(-\sigma_i \delta_i)\right)$ denotes how much light is contributed by sample $i$,  $\sigma_i$ denotes the opacity of sample $i$, and $\textbf{c}_i$ denotes the color of sample $i$, with distance $\delta_i$ to the next sample. Although this formula is not exact (it assumes single-scattering \cite{kajiya1984} and constant values between samples \cite{max1995}), it is differentiable and enables updating the 3D model based on the error of each training ray.

\subsection{Voxel Grid with Spherical Harmonics}
\label{sec:grid}

Similar to PlenOctrees~\cite{yu2021plenoctrees}, we use a sparse voxel grid for our geometry model. However, for simplicity and ease of implementing trilinear interpolation, we do not use an octree for our data structure. Instead, we store a dense 3D index array with pointers into a separate data array containing values for occupied voxels only.
Like PlenOctrees, each occupied voxel stores a scalar opacity $\sigma$ and a vector of spherical harmonic coefficients for each color channel. 
Spherical harmonics form an orthogonal basis for functions defined over the sphere, with low degree harmonics encoding smooth (more Lambertian) changes in color and higher degree harmonics encoding higher-frequency (more specular) effects.
The color of a sample $\textbf{c}_i$ is simply the sum of these harmonic basis functions for each color channel, weighted by the corresponding optimized coefficients and evaluated at the appropriate viewing direction. 
We use spherical harmonics of degree 2, which requires 9 coefficients per color channel for a total of 27 harmonic coefficients per voxel. We use degree 2 harmonics because PlenOctrees found that higher order harmonics confer only minimal benefit.

Our Plenoxel grid uses trilinear interpolation to define a continuous plenoptic function throughout the volume. This is in contrast to PlenOctrees, which assumes that the opacity and spherical harmonic coefficients remain constant inside each voxel. This difference turns out to be an important factor in successfully optimizing the volume, as we discuss below.
All coefficients (for opacity and spherical harmonics) are optimized directly, without any special initialization or pretraining with a neural network.

\subsection{Interpolation}
The opacity and color at each sample point along each ray are computed by trilinear interpolation of opacity and harmonic coefficients stored at the nearest 8 voxels. 
We find that trilinear interpolation significantly outperforms a simpler nearest neighbor interpolation; an ablation is presented in \cref{tab:interpablate}. 
The benefits of interpolation are twofold: interpolation increases the effective resolution by representing sub-voxel variations in color and opacity, and interpolation produces a continuous function approximation that is critical for successful optimization.
Both of these effects are evident in \cref{tab:interpablate}: doubling the resolution of a nearest-neighbor-interpolating Plenoxel closes much of the gap between nearest neighbor and trilinear interpolation at a fixed resolution, yet some gap remains due to the difficulty of optimizing a discontinuous model. 
Indeed, we find that trilinear interpolation is more stable with respect to variations in learning rate compared to nearest neighbor interpolation (we tuned the learning rates separately for each interpolation method in \cref{tab:interpablate}, to provide close to the best number possible for each setup).

\renewcommand{\tabcolsep}{5pt}
\begin{table}[h]
\centering
\begin{tabular}{@{}lcccc@{}}
\toprule
& & PSNR $\uparrow$ & SSIM $\uparrow$ & LPIPS $\downarrow$ \\ \cmidrule(r){1-1} \cmidrule(l){3-5} 
Trilinear, $256^3$ &  & 30.57 & 0.950 & 0.065 \\
Trilinear, $128^3$ &  & 28.46 & 0.926 & 0.100 \\
Nearest Neighbor, $256^3$ &  & 27.17 & 0.914 & 0.119 \\
Nearest Neighbor, $128^3$ &  & 23.73 & 0.866 & 0.176 \\
 \bottomrule
\end{tabular}
\caption{\textbf{Ablation over interpolation method.} Results are averaged over the 8 NeRF synthetic scenes. We find that trilinear interpolation provides dual benefits of improving effective resolution and improving optimization, such that trilinear interpolation at resolution $128^3$ outperforms nearest neighbor interpolation at $256^3$.}
\label{tab:interpablate}
\end{table}

\subsection{Coarse to Fine}
We achieve high resolution via a coarse-to-fine strategy that begins with a dense grid at lower resolution, optimizes, prunes unnecessary voxels, refines the remaining voxels by subdividing each in half in each dimension, and continues optimizing. For example, in the synthetic case, we begin with $256^3$ resolution and upsample to $512^3$. We use trilinear interpolation to initialize the grid values after each voxel subdivision step. In fact, we can resize between arbitrary resolutions using trilinear interpolation. Voxel pruning is performed using the method from PlenOctrees~\cite{yu2021plenoctrees}, which applies a threshold to the maximum weight $T_i (1- \exp(-\sigma_i \delta_i))$ of each voxel over all training rays (or, alternatively, to the density value in each voxel).
Due to trilinear interpolation, naively pruning can adversely impact the 
the color and density near surfaces since values at these points interpolate with the voxels in the immediate exterior.
To solve this issue, we perform a dilation operation so that a voxel is only pruned if both itself and its neighbors are deemed unoccupied.


\subsection{Optimization}

We optimize voxel opacities and spherical harmonic coefficients with respect to the mean squared error (MSE) over rendered pixel colors, with total variation (TV) regularization~\cite{rudin1994total}. Specifically, our base loss function is:
\begin{equation}
   \mathcal{L} = \mathcal{L}_{recon} + \lambda_{TV}\, \mathcal{L}_{TV}
\label{eq:loss}
\end{equation}
Where the MSE reconstruction loss $\mathcal{L}_{recon}$
and the total variation regularizer $\mathcal{L}_{TV}$ are:
\begin{equation*}
\begin{aligned}
   \mathcal{L}_{recon}  &= \frac1{|\mathcal{R}|}\sum_{\textbf{r} \in \mathcal{R}} \norm{C(\textbf{r}) - \hat C(\textbf{r})}_2^2 \\
\mathcal{L}_{TV} &=
\frac{1}{|\mathcal{V}|}
\sum_{\substack{\mathbf{v} \in \mathcal{V}\\d \in [D]}}
    \sqrt{\Delta_x^2(\mathbf{v},d) + \Delta_y^2(\mathbf{v},d) + \Delta_z^2(\mathbf{v},d)}
\end{aligned}
\end{equation*}
with $\Delta_x^2(\mathbf{v},d)$ shorthand for the squared difference between the $d$th value in voxel $\mathbf{v}:=(i,j,k)$ and the $d$th value in voxel $(i+1,j,k)$ normalized by the resolution, and analogously for $\Delta_y^2(\mathbf{v},d)$ and $\Delta_z^2(\mathbf{v},d)$.
Note in practice we use different weights for SH coefficients and $\sigma$ values. These weights are fixed for each scene type (bounded, forward-facing, and $360^{\circ}$). 

For faster iteration, we use a stochastic sample of the rays $\mathcal{R}$ to evaluate the MSE term and a stochastic sample of the voxels $\mathcal{V}$ to evaluate the TV term in each optimization step. We use the same learning rate schedule as JAXNeRF and Mip-NeRF~\cite{jaxnerf2020github, barron2021mipnerf}, but tune the initial learning rate separately for opacity and harmonic coefficients. The learning rate is fixed for all scenes in all datasets in the main experiments. 

Directly optimizing voxel coefficients is a challenging problem for several reasons: there are many values to optimize (the problem is high-dimensional), the optimization objective is nonconvex due to the rendering formula, and the objective is poorly conditioned. Poor conditioning is typically best resolved by using a second order optimization algorithm (\eg as recommended in~\cite{NoceWrig06}), but this is practically challenging to implement for a high-dimensional optimization problem because the Hessian is too large to easily compute and invert in each step. Instead, we use RMSProp~\cite{hinton} to ease the ill-conditioning problem without the full computational complexity of a second-order method.

\subsection{Unbounded Scenes}
We show that Plenoxels can be optimized for a wide range of settings beyond the synthetic scenes from the original NeRF paper.

With minor modifications, Plenoxels extend to real, unbounded scenes, both forward-facing and $360^{\circ}$. For forward-facing scenes, we use the same sparse voxel grid structure with normalized device coordinates, as defined in the original NeRF paper \cite{mildenhall2020nerf}.

\paragraph{Background model.} For $360^{\circ}$ scenes, we augment our sparse voxel grid foreground representation with a multi-sphere image (MSI) background model, which also uses learned voxel colors and opacities with trilinear interpolation within and between spheres.
Note that this is effectively the same as our foreground model, except the voxels are warped into spheres using
the simple equirectangular projection (voxels index over sphere angles $\theta$ and $\phi$).
We place 64 spheres linearly in inverse radius from $1$ to $\infty$ (we pre-scale the inner scene to be approximately  contained in the unit sphere).
To conserve memory, we store only rgb channels for the colors (only zero-order SH) and store all layers sparsely by using opacity thresholding as in our main model.
This is similar to the background model in NeRF++ \cite{zhang2020nerf}. 

\subsection{Regularization}

We illustrate the importance of TV regularization in \cref{fig:tv}. In addition to TV regularization, which encourages smoothness and is used on all scenes, for certain types of scenes we also use additional regularizers. 

\input{figures_tex/tv}

On the real, forward-facing and $360^\circ$ scenes, we use a sparsity prior based on a Cauchy loss following SNeRG~\cite{hedman2021baking}:
\begin{equation}
    \mathcal{L}_s = \lambda_s \sum_{i,k}\log \left(1 + 2\sigma(\mathbf{r}_i(t_k))^2 \right)
\end{equation}
where $\sigma(\mathbf{r}_i(t_k))$ denotes the opacity of sample $k$ along training ray $i$. In each minibatch of optimization on forward-facing scenes, we evaluate this loss term at each sample on each active ray.
This is also similar to the sparsity loss used in PlenOctrees~\cite{yu2021plenoctrees} and encourages voxels to be empty,
which helps to save memory and reduce quality loss when upsampling.

On the real, $360^{\circ}$ scenes, we also use a beta distribution regularizer on the accumulated foreground transmittance of each ray in each minibatch. This loss term, following Neural Volumes~\cite{lombardi2019neural}, promotes a clear foreground-background decomposition by encouraging the foreground to be either fully opaque or empty. This beta loss is:
\begin{equation}
    \mathcal{L}_{\beta} = \lambda_{\beta}\sum_{\mathbf{r}}\left(\log(T_{FG}(\mathbf{r})) + \log(1 - T_{FG}(\mathbf{r})) \right)
\end{equation}
where $\mathbf{r}$ are the training rays and $T_{FG}(\mathbf{r})$ is the accumulated foreground transmittance (between 0 and 1) of ray $\mathbf{r}$. 


\subsection{Implementation}

Since sparse voxel volume rendering is not well-supported in modern
autodiff libraries, we created a custom PyTorch CUDA \cite{cuda} extension library to achieve fast differentiable volume rendering; we hope practitioners will find this implementation useful in their applications. We also provide a slower, higher-level JAX~\cite{jax2018github} implementation. Both implementations will be released to the public. 


The speed of our implementation is possible in large part because the gradient of our Plenoxel model becomes very sparse very quickly, as shown in \cref{fig:sparsegrad}. Within the first 1-2 minutes of optimization, fewer than 10\% of the voxels have nonzero gradients.

\input{figures_tex/sparsegrad}

\section{Results}
\label{sec:results}

\input{figures_tex/1epoch}

We present results on synthetic, bounded scenes; real, unbounded, forward-facing scenes; and real, unbounded, $360^{\circ}$ scenes. We include time trial comparisons with prior work, showing dramatic speedup in training compared to all prior methods (alongside real-time rendering). Quantitative comparisons are presented in \cref{tab:results}, and visual comparisons are shown in \cref{fig:synthetic}, \cref{fig:forward-facing}, and \cref{fig:360}. Our method achieves quality results after even the first epoch of optimization, less than 1.5 minutes, as shown in \cref{fig:1epoch}.

We also present the results from various ablation studies of our method. In the main text we present average results (PSNR, SSIM~\cite{ssim}, and VGG LPIPS~\cite{lpips}) over all scenes of each type; full results on each scene individually are included in the supplement. We include full experimental details (hyperparameters, etc.) in the supplement.

\subsection{Synthetic Scenes}

Our synthetic experiments use the 8 scenes from NeRF: chair, drums, ficus, hotdog, lego, materials, mic, and ship. Each scene includes 100 ground truth training views with 800 $\times$ 800 resolution, from known camera positions distributed randomly in the upper hemisphere facing the object, which is set against a plain white background. Each scene is evaluated on 200 test views, also with resolution 800 $\times$ 800 and known inward-facing camera positions in the upper hemisphere. We provide quantitative comparisons in \cref{tab:results} and visual comparisons in \cref{fig:synthetic}.

\renewcommand{\tabcolsep}{2pt}
\begin{table}[t]
\begin{tabular}{@{}llccccc@{}}
\toprule
 &  & PSNR $\uparrow$ & SSIM $\uparrow$ & LPIPS $\downarrow$ & & Train Time  \\ \cmidrule(r){1-1} \cmidrule(l){3-5}  \cmidrule(l){7-7}
 Ours  &  & 31.71 & \textbf{0.958} & \textbf{0.049} &  & \textbf{11 mins} \\
NV~\cite{lombardi2019neural} &  & 26.05 & 0.893 & 0.160 &  & $>$1 day \\
JAXNeRF~\cite{mildenhall2020nerf, jaxnerf2020github} &  & \textbf{31.85} & 0.954 & 0.072 &  & 1.45 days \\ \cmidrule(r){1-1} \cmidrule(l){3-5}  \cmidrule(l){7-7}
Ours & & 26.29 & \textbf{0.839} & \textbf{0.210} &  & \textbf{24 mins} \\
LLFF~\cite{mildenhall2019local} &  & 24.13 & 0.798 & 0.212 &  & ---* \\
JAXNeRF~\cite{mildenhall2020nerf, jaxnerf2020github} &  & \textbf{26.71} & 0.820 & 0.235 &  & 1.62 days \\ \cmidrule(r){1-1} \cmidrule(l){3-5}  \cmidrule(l){7-7}
Ours &  &  20.40 & \textbf{0.696} & \textbf{0.420} &  & \textbf{27 mins} \\
NeRF++~\cite{zhang2020nerf} &  & \textbf{20.49} & 0.648 &  0.478 &  & $\sim$4 days \\
\bottomrule
\end{tabular}
\caption{\textbf{Results.} \textit{Top:} average over the 8 synthetic scenes from NeRF; \textit{Middle:} the 8 real, forward-facing scenes from NeRF; \textit{Bottom:} the 4 real, $360^\circ$ scenes from Tanks and Temples~\cite{tanks}. 4 of the synthetic scenes train in under 10 minutes. *LLFF requires pretraining a network to predict MPIs for each view, and then can render novel scenes without further training; this pretraining is amortized across all scenes so we do not include it in the table.}
\label{tab:results}
\end{table}

\input{figures_tex/synthetic}

We compare our method to Neural Volumes (NV)~\cite{lombardi2019neural} (as a prior method that predicts a grid for each scene, using a 3D convolutional network), and JAXNeRF~\cite{mildenhall2020nerf, jaxnerf2020github}. For Neural Volumes we use values reported in~\cite{mildenhall2020nerf}; for JAXNeRF we report results from our own rerunning, fixing the centered pixel bug. Our method achieves comparable quality compared to the best baseline, while training in an average of 11 minutes per scene on a single GPU and supporting interactive rendering.

\subsection{Real Forward-Facing Scenes}

We extend our method to unbounded, forward-facing scenes by using normalized device coordinates (NDC), as derived in NeRF~\cite{mildenhall2020nerf}. Our method is otherwise identical to the version we use on bounded, synthetic scenes, except that we use TV regularization (with a stronger weight) throughout the optimization. This change is likely necessary because of the reduced number of training views for these scenes, as described in \cref{sec:ablate}.

Our forward-facing experiments use the same 8 scenes as in NeRF, 5 of which are originally from LLFF~\cite{mildenhall2019local}. Each scene consists of 20 to 60 forward-facing images captured by a handheld cell phone with resolution 1008 $\times$ 756, with $\frac{7}{8}$ of the images used for training and the remaining $\frac{1}{8}$ of the images reserved as a test set. 

We compare our method to Local Light Field Fusion (LLFF)~\cite{mildenhall2019local} (a prior method that uses a 3D convolutional network to predict a grid for each input view) and JAXNeRF. We provide quantitative comparisons in \cref{tab:results} and visual comparisons in \cref{fig:forward-facing}.


\input{figures_tex/forward_facing}


\subsection{Real $360^{\circ}$ Scenes}

We extend our method to real, unbounded, $360^{\circ}$ scenes by surrounding our sparse voxel grid with an multi-sphere image (MSI, based on multi-plane images introduced by~\cite{zhou2018stereo}) background model, in which each background sphere is also a simple voxel grid with trilinear interpolation (both within each sphere and between adjacent background sphere layers). 

Our $360^{\circ}$ experiments use 4 scenes from the Tanks and Temples dataset~\cite{tanks}: M60, playground, train, and truck. For each scene, we use the same train/test split as~\cite{riegler2020free}.

We compare our method to NeRF++~\cite{zhang2020nerf}, which augments NeRF with a background model to represent unbounded scenes. We present quantitative comparisons in \cref{tab:results} and visual comparisons in \cref{fig:360}.


\input{figures_tex/360}

\subsection{Ablation Studies}
\label{sec:ablate}

In this section, we perform extensive ablation studies of our method to understand which features are core to its success, with such a simple model. In \cref{tab:interpablate}, we show that continuous (in our case, trilinear) interpolation is responsible for dramatic improvement in fidelity compared to nearest neighbor interpolation (\ie constant within each voxel)~\cite{yu2021plenoctrees}. 

In \cref{tab:ntrainablate}, we consider how our method handles a dramatic reduction in training data, from 100 views to 25 views, on the 8 synthetic scenes. We compare our method to NeRF and find that, despite its lack of complex neural priors, by increasing TV regularization our method can outperform NeRF even in this limited data regime. This ablation also sheds light on why our model performs better with higher TV regularization on the real forward-facing scenes compared to the synthetic scenes: the real scenes have many fewer training images, and the stronger regularizer helps our optimization extend smoothly to sparsely-supervised regions.

We also ablate over the resolution of our Plenoxel grid in \cref{tab:resoablate} and the rendering formula in \cref{tab:renderingcompare}. The rendering formula from Max \cite{max1995} yields a substantial improvement compared to that of Neural Volumes \cite{lombardi2019neural}, perhaps because it is more physically accurate (as discussed further in the supplement). The supplement also includes ablations over the learning rate schedule and optimizer demonstrating  Plenoxel optimization to be robust to these hyperparameters.

\begin{table}[]
\begin{tabular}{@{}llccc@{}}
\toprule
 &  & PSNR $\uparrow$ & SSIM $\uparrow$ & LPIPS $\downarrow$ \\ \cmidrule(r){1-1} \cmidrule(l){3-5} 
Ours: 100 images (low TV) &  & 31.71 & 0.958 & 0.050 \\
NeRF: 100 images~\cite{mildenhall2020nerf} &  & 31.01 & 0.947 & 0.081 \\ \cmidrule(r){1-1} \cmidrule(l){3-5} 
Ours: 25 images (low TV) &  & 26.88 & 0.911 & 0.099 \\
Ours: 25 images (high TV) &  & 28.25 & 0.932 & 0.078 \\
NeRF: 25 images~\cite{mildenhall2020nerf} &  & 27.78 & 0.925 & 0.108 \\
\bottomrule
\end{tabular}
\caption{\textbf{Ablation over the number of views.} By increasing our TV regularization, we exceed NeRF fidelity even when the number of training views is only a quarter of the full dataset. Results are averaged over the 8 synthetic scenes from NeRF.}
\label{tab:ntrainablate}
\end{table}

\begin{table}[]
\renewcommand{\tabcolsep}{11.5pt}
\begin{tabular}{@{}llccc@{}}
\toprule
Resolution &  & PSNR $\uparrow$ & SSIM $\uparrow$ & LPIPS $\downarrow$ \\ \cmidrule(r){1-1} \cmidrule(l){3-5} 
$512^3$ &  & 31.71 & 0.958 & 0.050 \\
$256^3$ &  & 30.57 & 0.950 & 0.065 \\
$128^3$ &  & 28.46 & 0.926 & 0.100 \\
$64^3$ &  & 26.11 & 0.892 & 0.139 \\
$32^3$ &  & 23.49 & 0.859 & 0.174 \\
\bottomrule
\end{tabular}
\caption{\textbf{Ablation over the Plenoxel grid resolution.} Results are averaged over the 8 synthetic scenes from NeRF.}
\label{tab:resoablate}
\end{table}

\begin{table}[]
\renewcommand{\tabcolsep}{3pt}
\begin{tabular}{@{}llccc@{}}
\toprule
Rendering Formula &  & PSNR $\uparrow$ & SSIM $\uparrow$ & LPIPS $\downarrow$ \\ \cmidrule(r){1-1} \cmidrule(l){3-5} 
Max \cite{max1995}, used in NeRF \cite{mildenhall2020nerf} &  & 30.57 & 0.950 & 0.065 \\
Neural Volumes \cite{lombardi2019neural} &  & 27.54 & 0.906 & 0.201 \\
\bottomrule
\end{tabular}
\caption{\textbf{Comparison of different rendering formulas.} We compare the rendering formula from Max \cite{max1995} (used in NeRF and our main method) to the one used in Neural Volumes \cite{lombardi2019neural}, which uses absolute instead of relative transmittance. Results are averaged over the 8 synthetic scenes from NeRF.}
\label{tab:renderingcompare}
\end{table}



    



\section{Discussion}
We present a method for photorealistic scene modeling and novel viewpoint rendering that produces results with comparable fidelity to the state-of-the-art, while taking orders of magnitude less time to train. Our method is also strikingly straightforward, shedding light on the core elements that are necessary for solving 3D inverse problems: a differentiable forward model, a continuous representation (in our case, via trilinear interpolation), and appropriate regularization. We acknowledge that the ingredients for this method have been available for a long time, however nonlinear optimization with tens of millions of variables has only recently become accessible to the computer vision practitioner.




\paragraph{Limitations and Future Work.}
As with any underdetermined inverse problem, our method is susceptible to artifacts. Our method exhibits different artifacts than neural methods, as shown in \cref{fig:artifacts}, but both methods achieve similar quality in terms of standard metrics (as presented in \cref{sec:results}). Future work may be able to adjust or mitigate these remaining artifacts by studying different regularization priors and/or more physically accurate differentiable rendering functions.

Although we report all of our results for each dataset with a fixed set of hyperparameters, there is no optimal a priori setting of the TV weight $\lambda_{TV}$. In practice better results may be obtained by tuning this parameter on a scene-by-scene basis, which is possible due to our fast training time. This is expected because the scale, smoothness, and number of training views varies between scenes. We note that NeRF also has hyperparameters to be set such as the length of positional encoding, learning rate, and number of layers, and tuning these may also increase performance on a scene-by-scene basis.


\input{figures_tex/drumfail}

Our method should extend naturally to support multiscale rendering with proper anti-aliasing through voxel cone-tracing, similar to the modifications in Mip-NeRF~\cite{barron2021mipnerf}. Another easy addition is tone-mapping to account for white balance and exposure changes, which we expect would help especially in the real $360^\circ$ scenes.
A hierarchical data structure (such as an octree) may provide additional speedup compared to our sparse array implementation, provided that differentiable interpolation is preserved.

Since our method is two orders of magnitude faster than NeRF, we believe that
it may enable downstream applications currently bottlenecked by the performance of NeRF--for example, multi-bounce lighting
and 3D generative models across large databases of scenes.
By combining our method with additional components such as camera optimization and
large-scale voxel hashing, it may enable a 
practical pipeline for end-to-end photorealistic 3D reconstruction.

\section*{Acknowledgements}

    We note that Utkarsh Singhal and Sara Fridovich-Keil tried a related idea with point clouds
some time prior to this project.
Additionally, we would like to thank Ren Ng for helpful suggestions
and Hang Gao for reviewing a draft of the paper.
The project is generously supported in part by the CONIX Research Center, one of six centers in JUMP, a Semiconductor Research Corporation (SRC) program sponsored by DARPA; a Google research award to Angjoo Kanazawa;
Benjamin Recht's ONR awards N00014-20-1-2497 and N00014-18-1-2833, NSF CPS award 1931853, and the DARPA Assured Autonomy program (FA8750-18-C-0101).
Sara Fridovich-Keil and Matthew Tancik are supported by the NSF GRFP.

{\small
\bibliographystyle{ieee_fullname}
\bibliography{references}
}

\include{supp}

\end{document}

%% file: figures_tex/teaser.tex
\begin{figure}[]
  \centering
  \includegraphics[width=.8\linewidth]{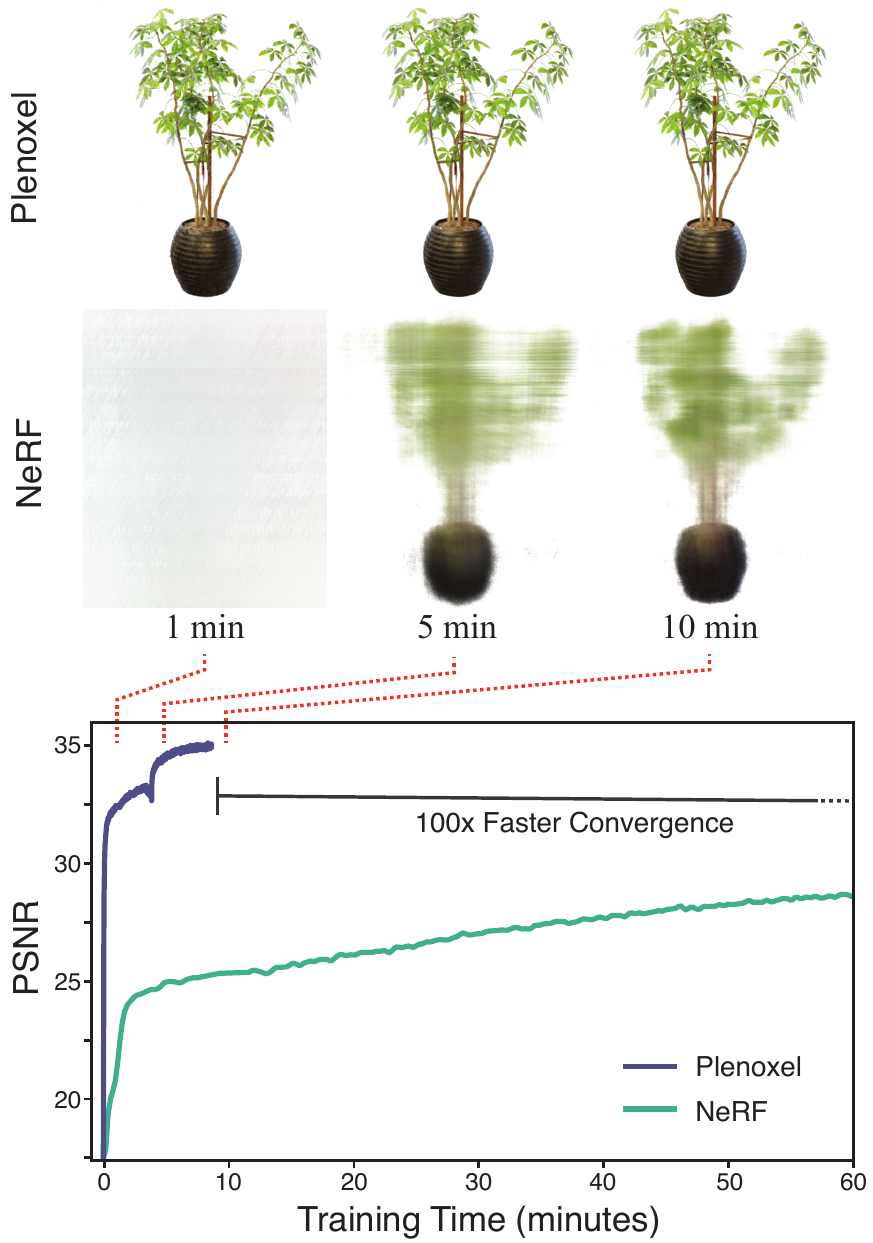}
  \caption{\textbf{Plenoxel: Plenoptic Volume Elements} for fast optimization of radiance fields. We show that direct optimization of a fully explicit 3D model can match the rendering quality of modern neural based approaches such as NeRF while optimizing over two orders of magnitude faster.}
  \label{fig:teaser}
\end{figure}

%% file: figures_tex/overview.tex
\begin{figure*}[t]
  \centering
  \includegraphics[width=\linewidth]{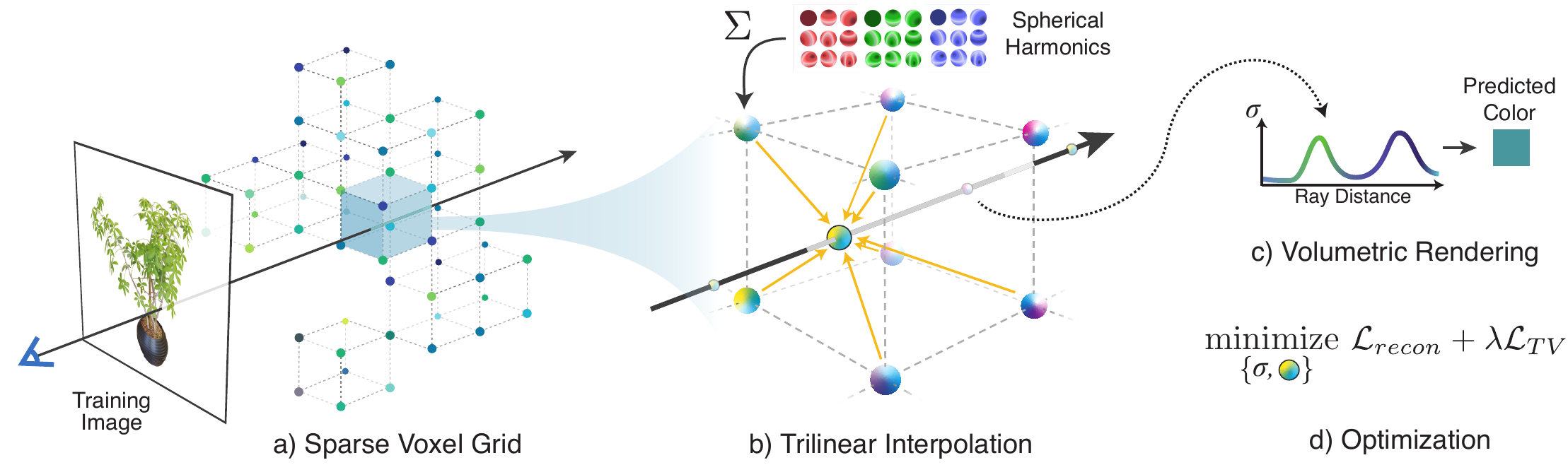}
  \caption{
    \textbf{Overview of our sparse Plenoxel model.} 
    Given a set of images of an object or scene, 
    we reconstruct a (a) sparse voxel (``Plenoxel'') grid with 
    density and spherical harmonic coefficients at each voxel.
    To render a ray, we (b) compute the color and opacity of each sample point via trilinear interpolation of the neighboring voxel coefficients.
    We integrate the color and opacity of these samples using (c) differentiable volume rendering, following the recent success of NeRF~\cite{mildenhall2020nerf}.
    The voxel coefficients can then be (d) optimized using
    the standard MSE reconstruction loss 
    relative to the training images, along with a total variation regularizer.
  }
  \label{fig:overview}
\end{figure*}

%% file: figures_tex/tv.tex
\newcommand{\tvwidth}{0.24\linewidth}

\newcommand{\tvleft}{.8\width}
\newcommand{\tvright}{0.01\width}
\newcommand{\tvbottom}{.25\height}
\newcommand{\tvtop}{.4\height}

\begin{figure}[]
  \captionsetup[subfigure]{labelformat=empty}
  \centering
    \begin{subfigure}[b]{\tvwidth}
        \adjincludegraphics[trim={{\tvleft} {\tvbottom} {\tvright} {\tvtop}}, clip, width=\linewidth]{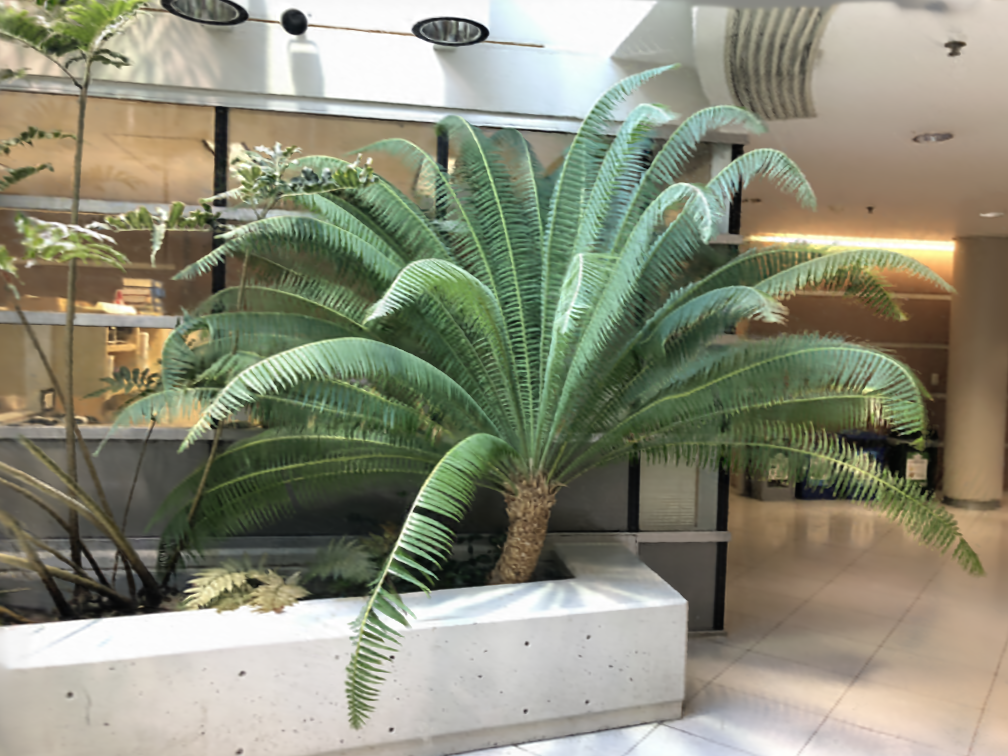}
        \caption{Full}
    \end{subfigure}
    \begin{subfigure}[b]{\tvwidth}
        \adjincludegraphics[trim={{\tvleft} {\tvbottom} {\tvright} {\tvtop}}, clip, width=\linewidth]{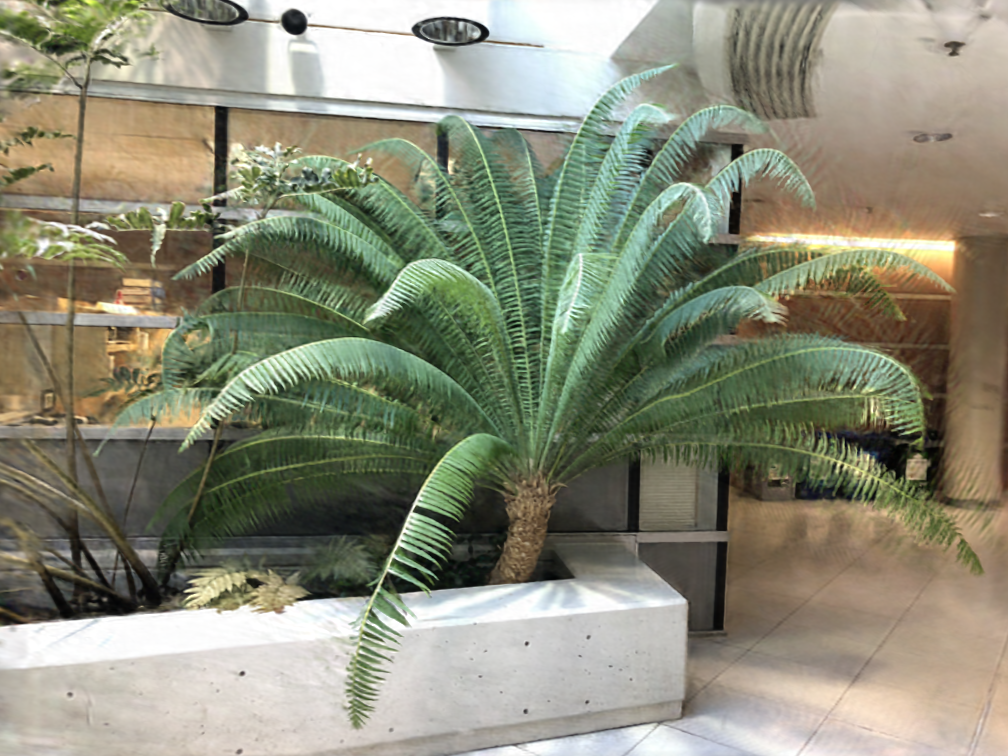}
        \caption{No SH TV}
    \end{subfigure}
    \begin{subfigure}[b]{\tvwidth}
        \adjincludegraphics[trim={{\tvleft} {\tvbottom} {\tvright} {\tvtop}}, clip, width=\linewidth]{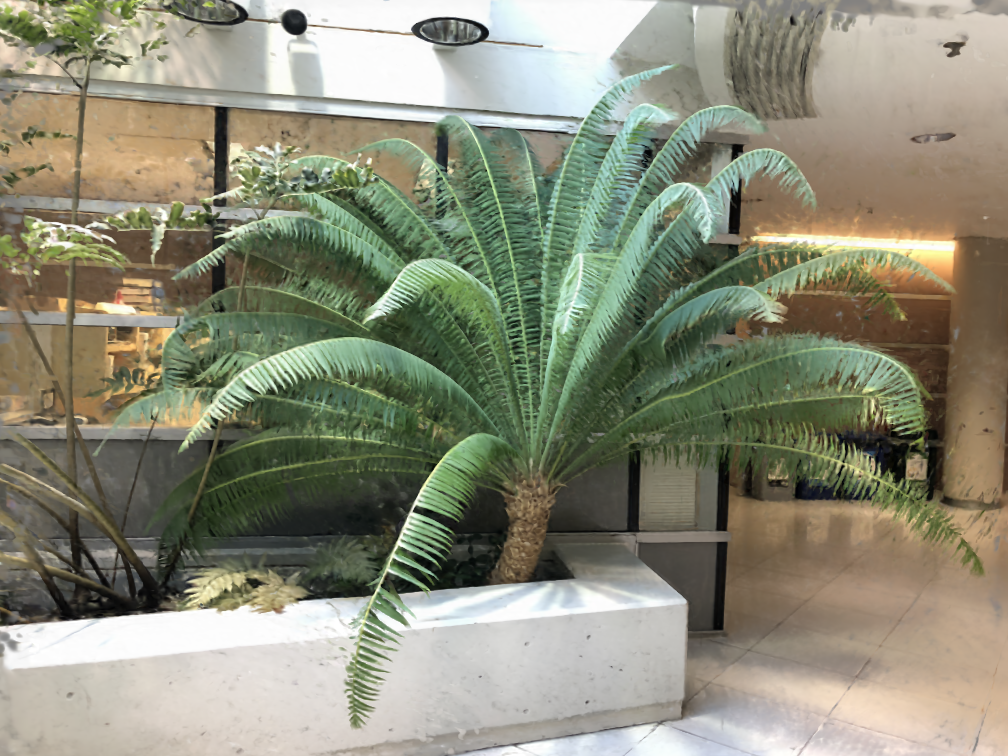}
        \caption{No $\sigma$ TV}
    \end{subfigure}
    \begin{subfigure}[b]{\tvwidth}
        \adjincludegraphics[trim={{\tvleft} {\tvbottom} {\tvright} {\tvtop}}, clip, width=\linewidth]{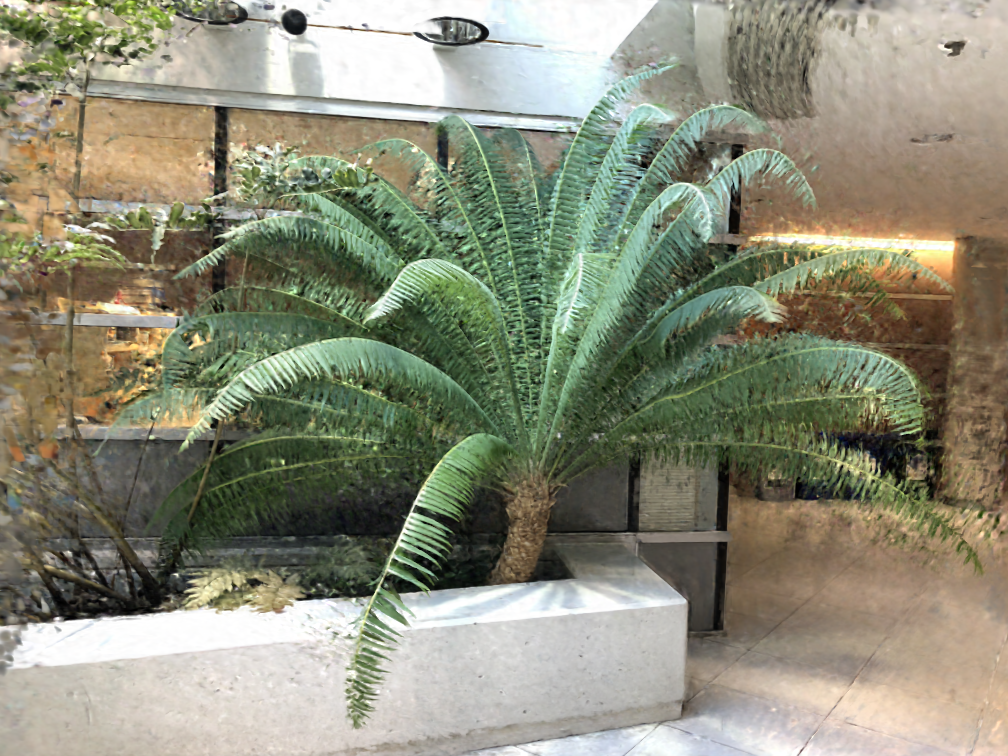}
        \caption{No TV}
    \end{subfigure}
  \caption{\textbf{Ablation over TV regularization.} Clear artifacts are visible in the forward-facing scenes without TV on both $\sigma$ and SH coefficients, although PSNR does not always reflect this.}
    \label{fig:tv}
\end{figure}

%% file: figures_tex/sparsegrad.tex
\begin{figure}[t]
  \centering
  \includegraphics[width=\linewidth]{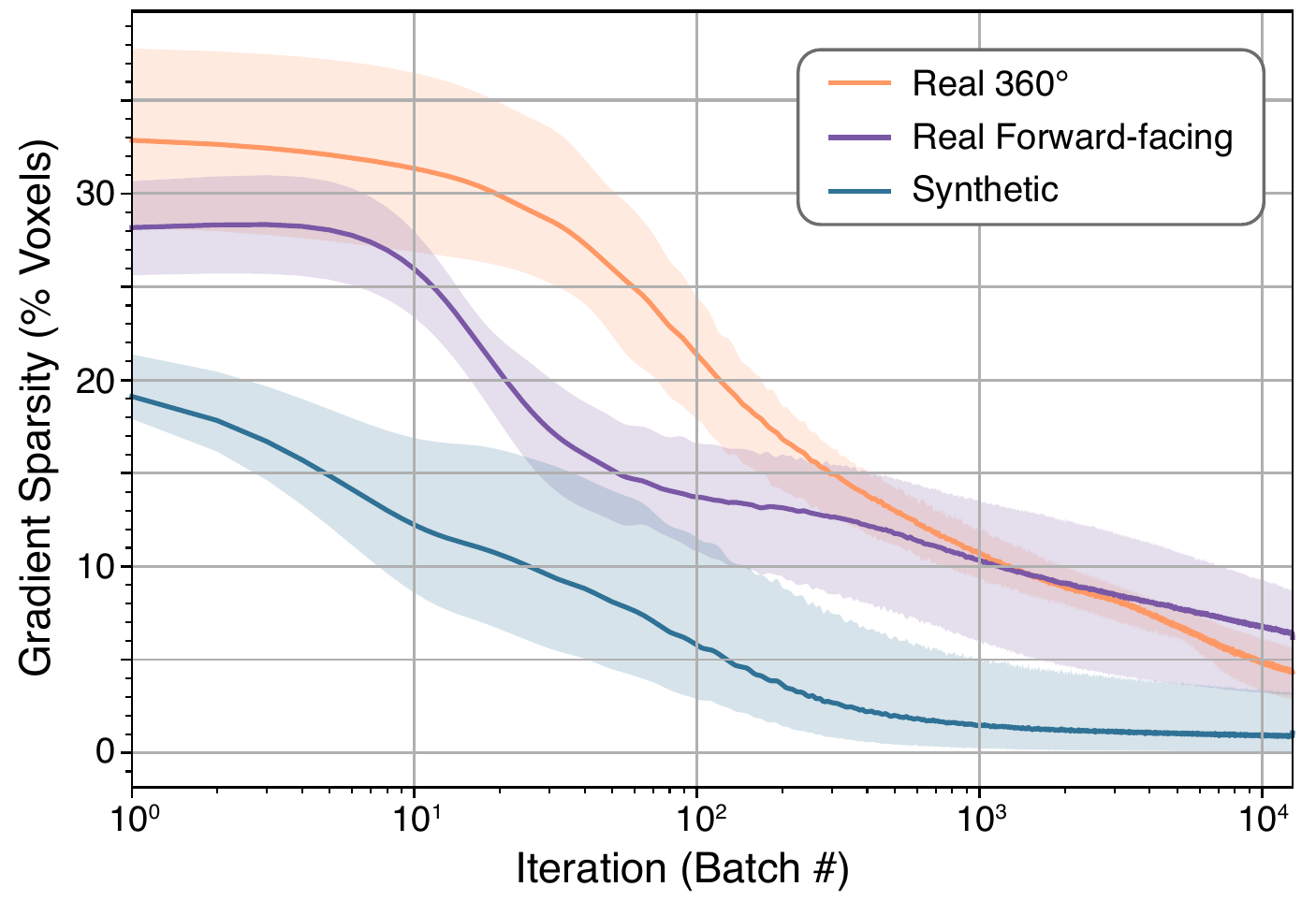}
  \caption{\textbf{Gradient sparsity.} The gradient becomes very sparse spatially within the first 12800 batches (one epoch for the synthetic scenes), with  as few as 1\% of the voxels updating per batch in the synthetic case.
  This enables efficient training via sparse parameter updates.
The solid lines show the mean and the shaded regions show the full range of values among all scenes of each type.}
  \label{fig:sparsegrad}
\end{figure}

%% file: figures_tex/1epoch.tex
\newcommand{\epochwidth}{0.12\linewidth}

\begin{figure*}[t]
  \centering
  \includegraphics[width=\epochwidth]{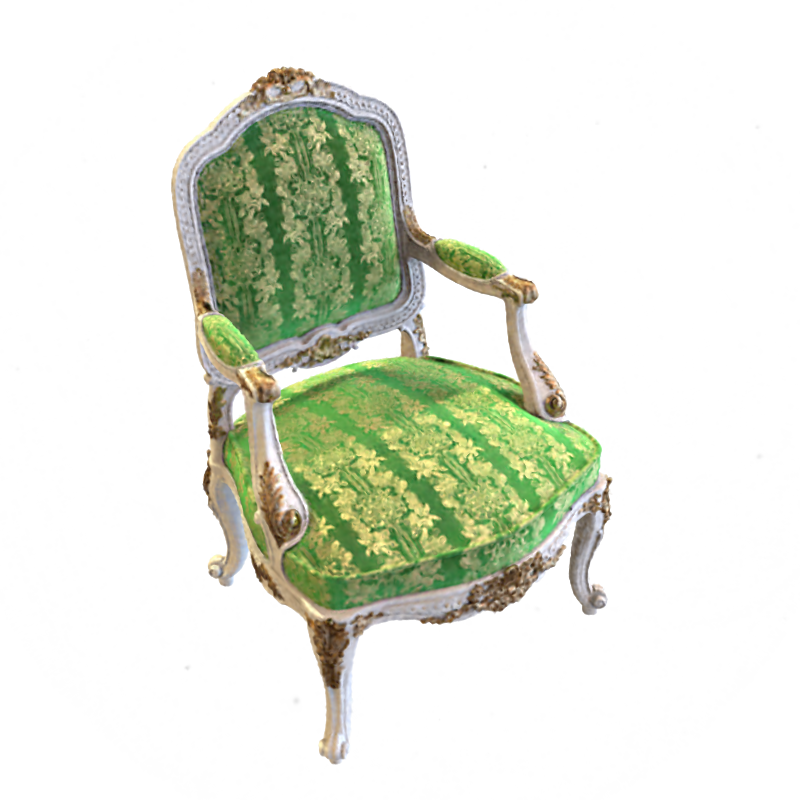}
  \includegraphics[width=\epochwidth]{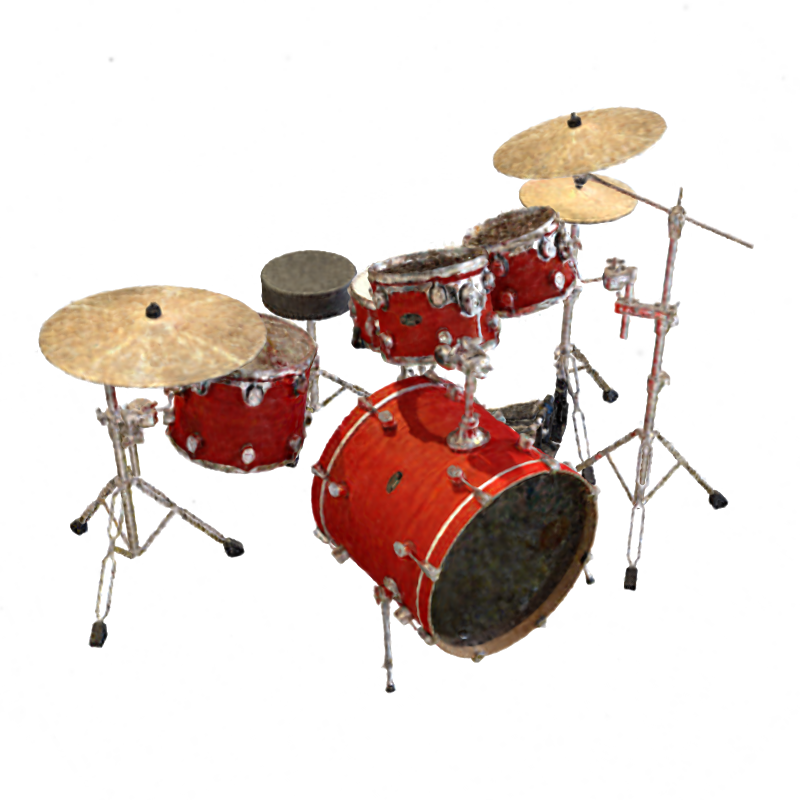}
  \includegraphics[width=\epochwidth]{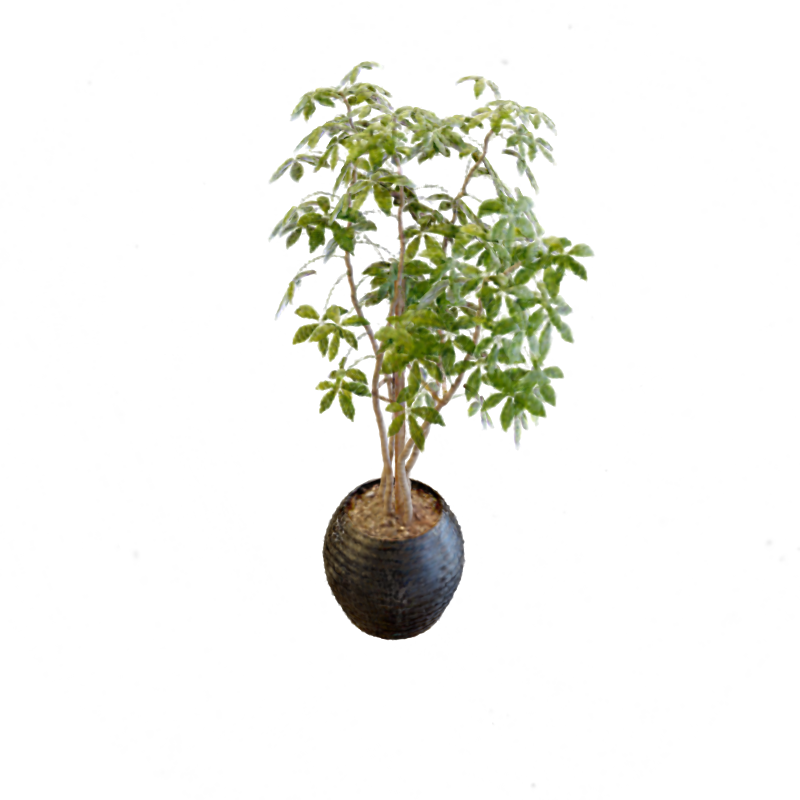}
  \includegraphics[width=\epochwidth]{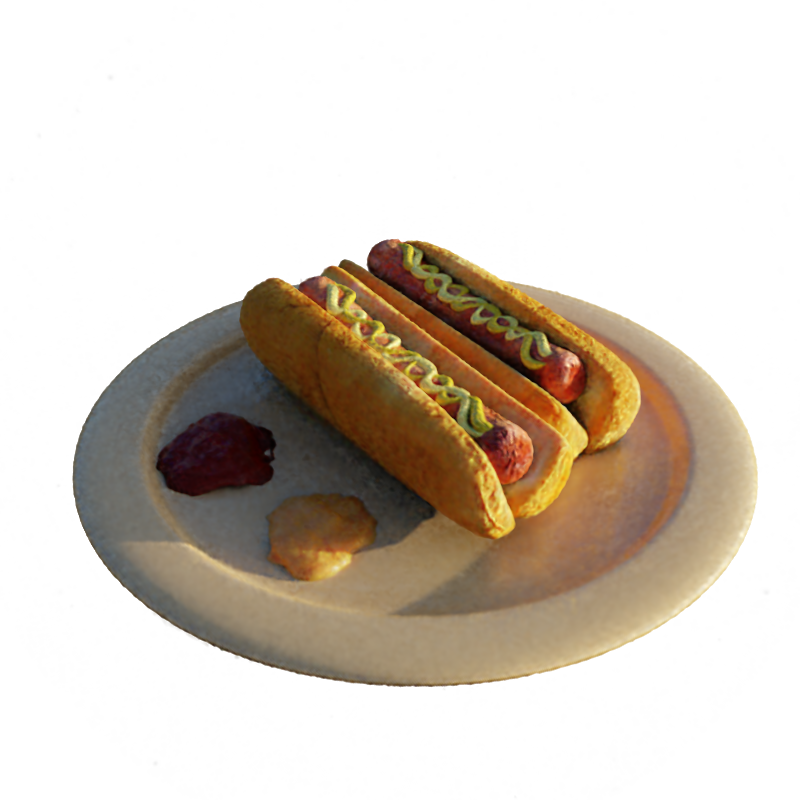}
  \includegraphics[width=\epochwidth]{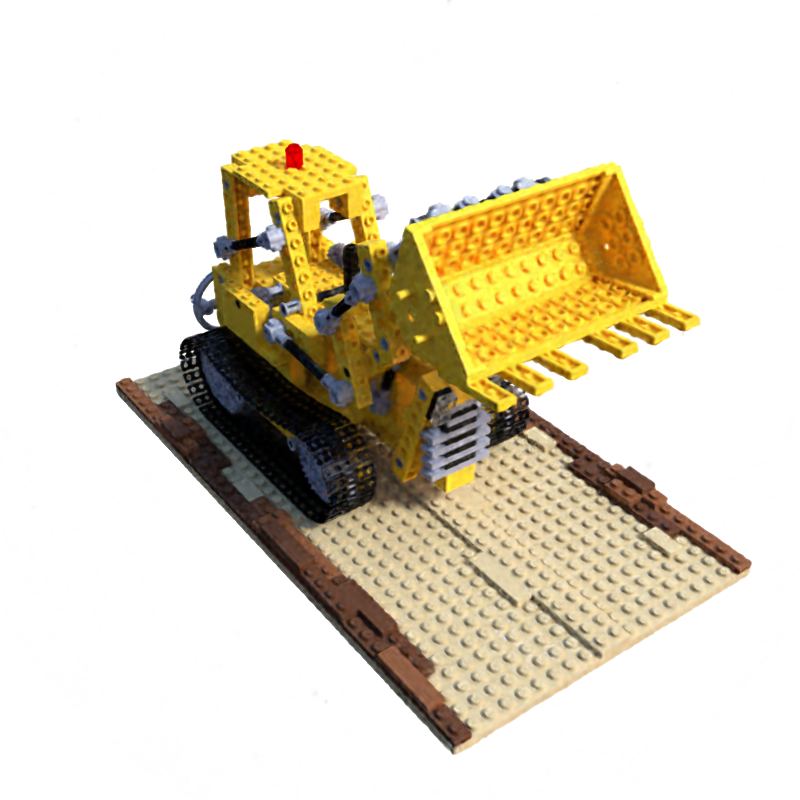}
  \includegraphics[width=\epochwidth]{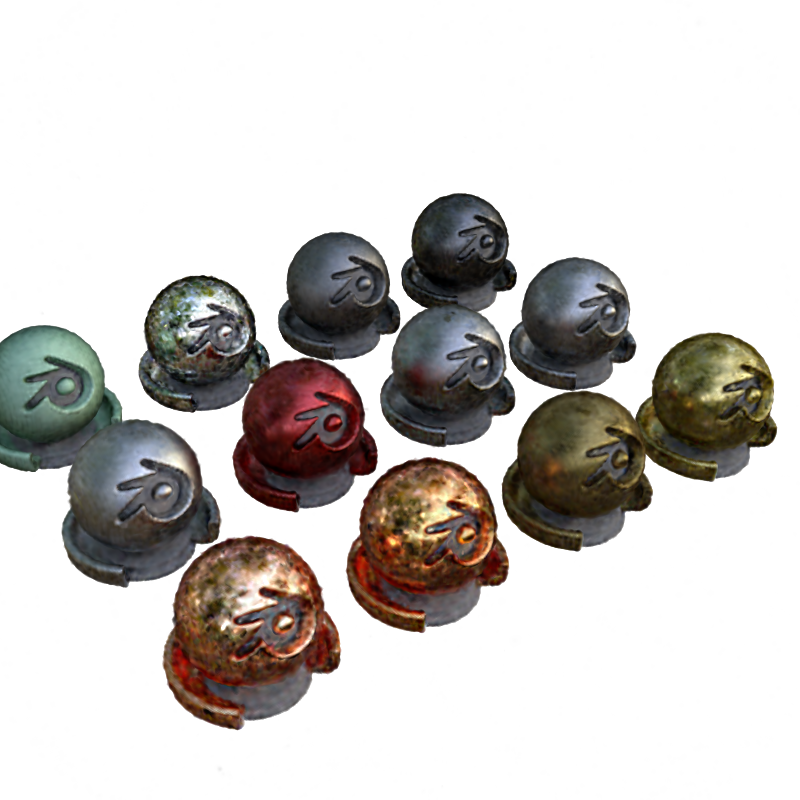}
  \includegraphics[width=\epochwidth]{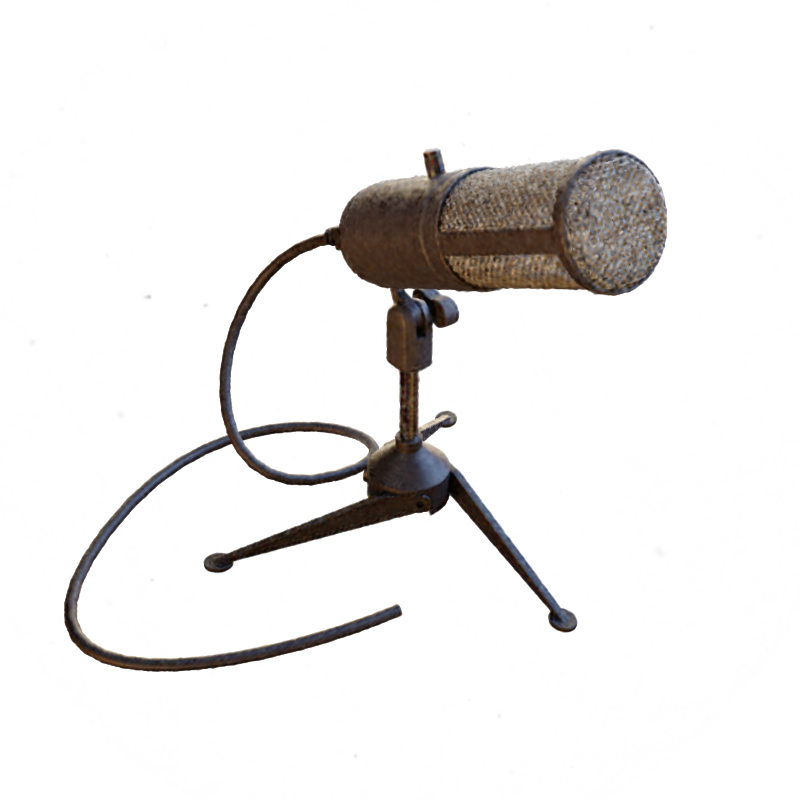}
  \includegraphics[width=\epochwidth]{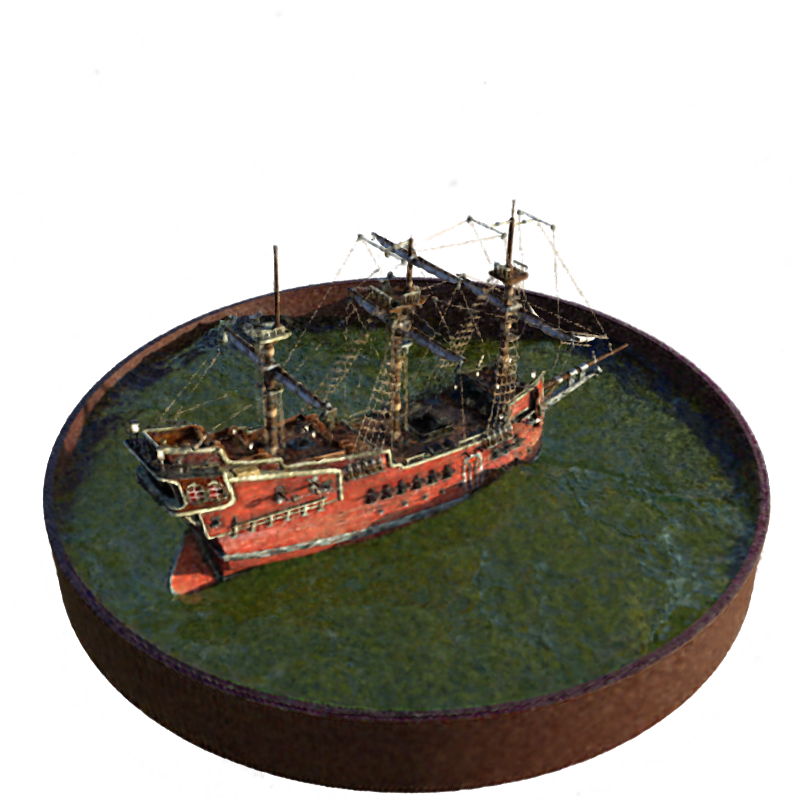}
  \caption{\textbf{1 minute, 20 seconds.} Results on the synthetic scenes after 1 epoch of optimization, an average of 1 minute and 20 seconds.}
  \label{fig:1epoch}
\end{figure*}

%% file: figures_tex/synthetic.tex
\newcommand{\synthwidth}{0.3\linewidth}

\newcommand{\legoleft}{.45\width}
\newcommand{\legoright}{.25\width}
\newcommand{\legobottom}{.35\height}
\newcommand{\legotop}{.35\height}

\newcommand{\shipleft}{.55\width}
\newcommand{\shipright}{.15\width}
\newcommand{\shipbottom}{.25\height}
\newcommand{\shiptop}{.45\height}

\begin{figure}[]
  \captionsetup[subfigure]{labelformat=empty}
  \centering
    \begin{subfigure}[b]{\synthwidth}
        \adjincludegraphics[trim={{\legoleft} {\legobottom} {\legoright} {\legotop}}, clip, width=\linewidth]{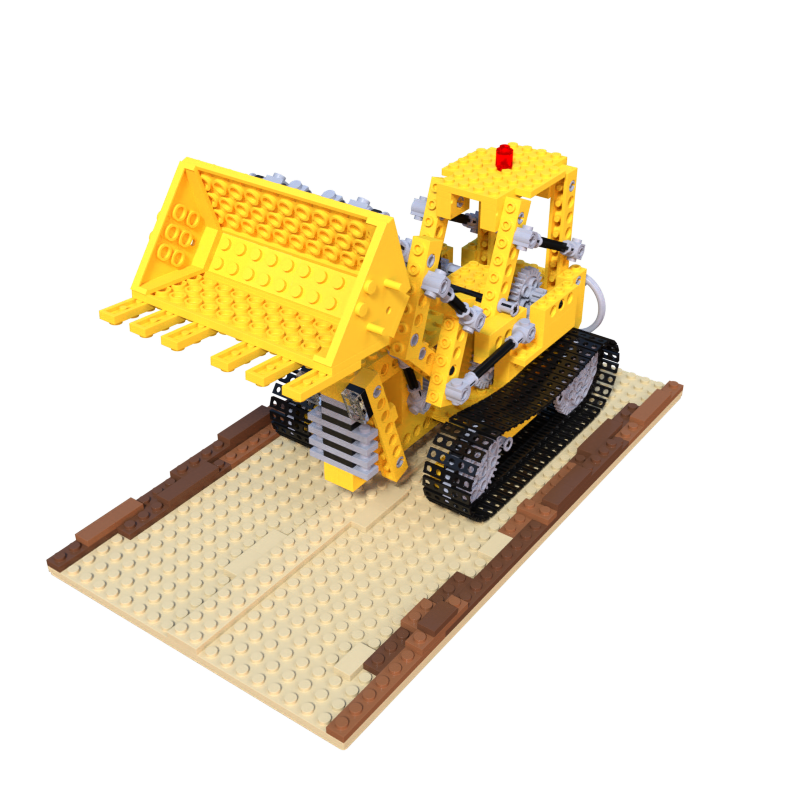}
    \end{subfigure}
    \begin{subfigure}[b]{\synthwidth}
        \adjincludegraphics[trim={{\legoleft} {\legobottom} {\legoright} {\legotop}}, clip, width=\linewidth]{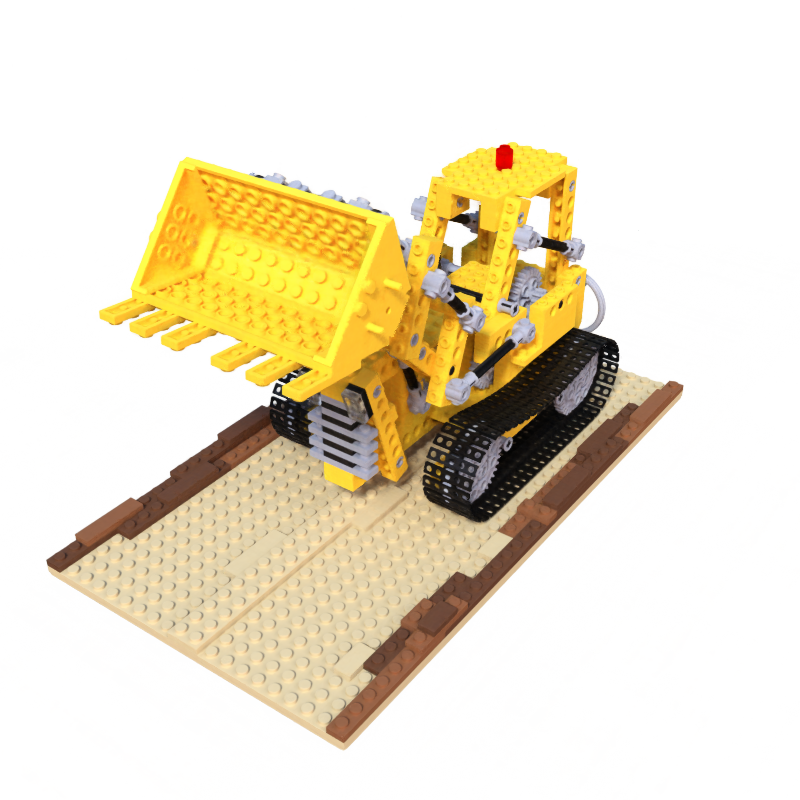}
    \end{subfigure}
    \begin{subfigure}[b]{\synthwidth}
         \adjincludegraphics[trim={{\legoleft} {\legobottom} {\legoright} {\legotop}}, clip, width=\linewidth]{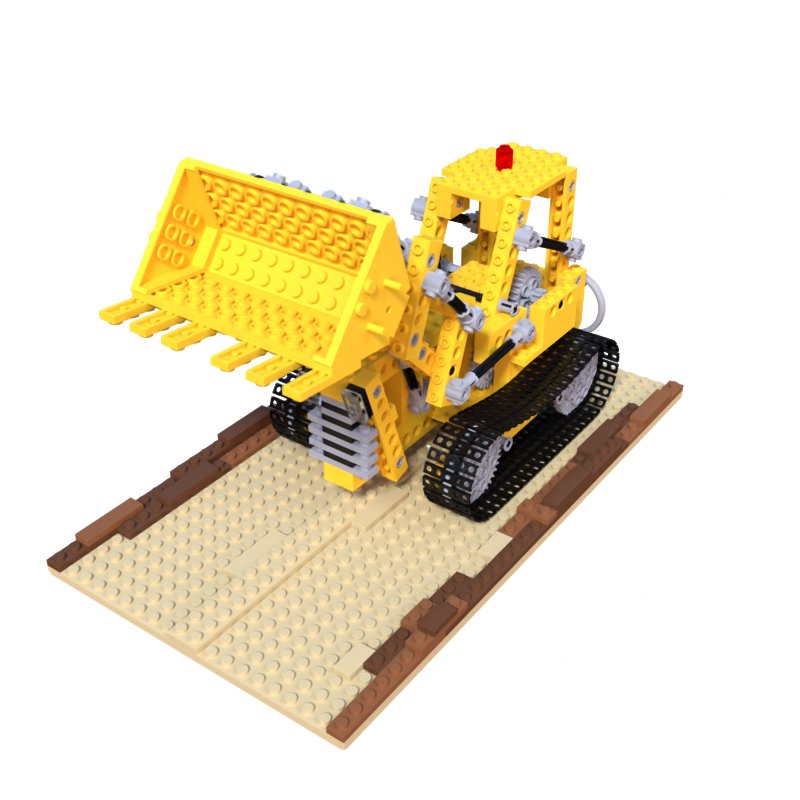}
    \end{subfigure}
    \begin{subfigure}[b]{\synthwidth}
        \adjincludegraphics[trim={{\shipleft} {\shipbottom} {\shipright} {\shiptop}}, clip, width=\linewidth]{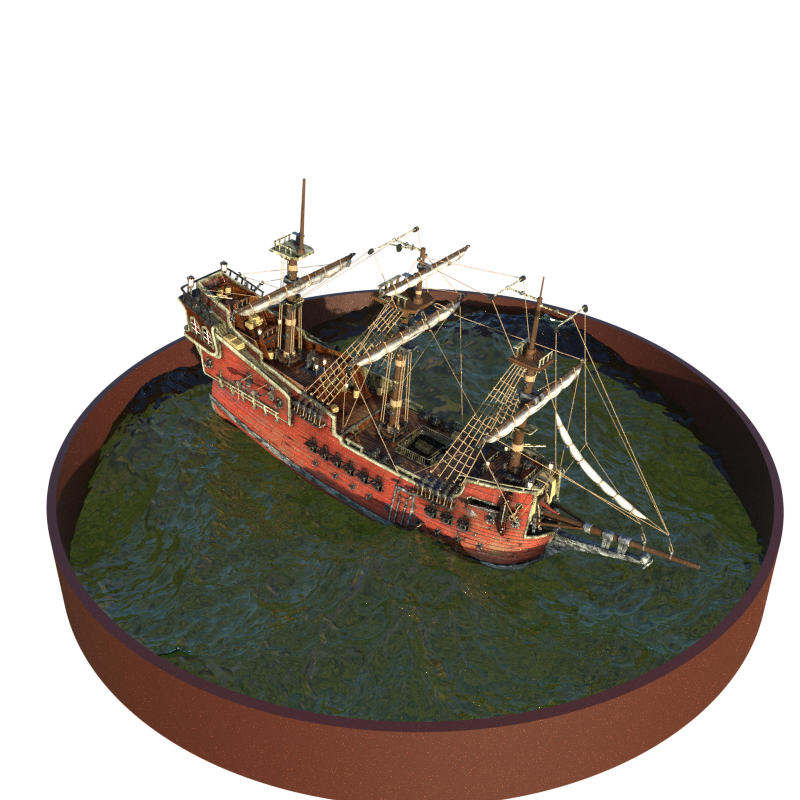}
        \caption{Ground Truth}
    \end{subfigure}
    \begin{subfigure}[b]{\synthwidth}
        \adjincludegraphics[trim={{\shipleft} {\shipbottom} {\shipright} {\shiptop}}, clip, width=\linewidth]{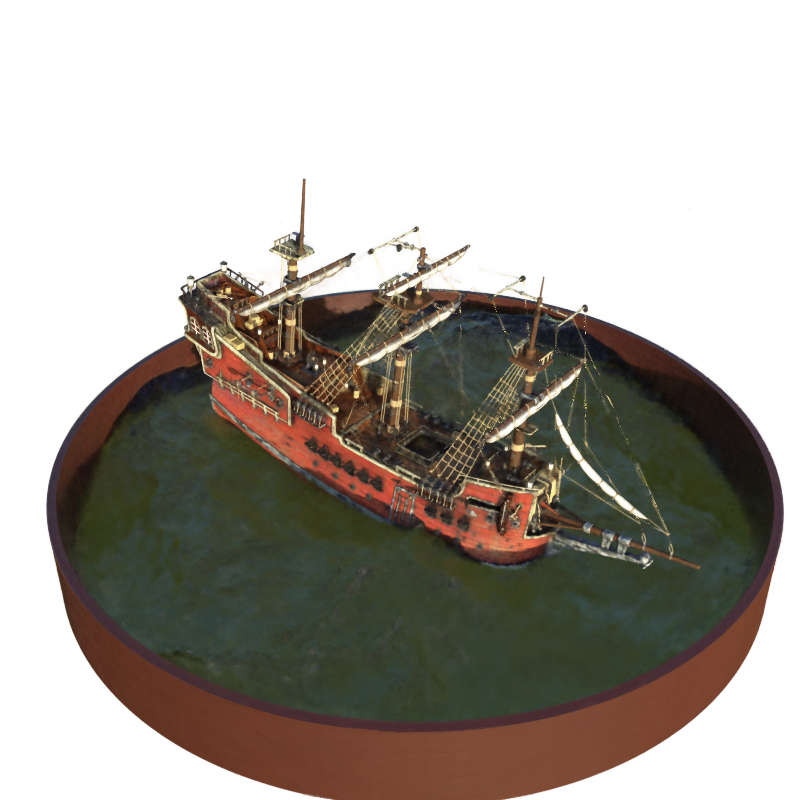}
        \caption{JAXNeRF~\cite{jaxnerf2020github, mildenhall2020nerf}}
    \end{subfigure}
    \begin{subfigure}[b]{\synthwidth}
        \adjincludegraphics[trim={{\shipleft} {\shipbottom} {\shipright} {\shiptop}}, clip, width=\linewidth]{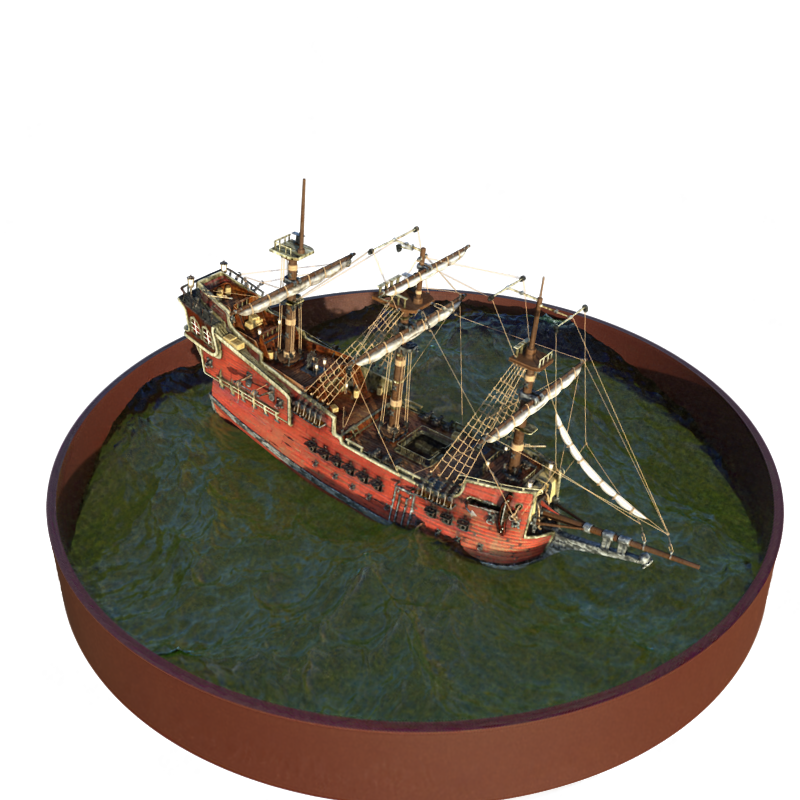}
        \caption{Plenoxels}
    \end{subfigure}
  \caption{\textbf{Synthetic, bounded scenes.} Example results on the lego and ship synthetic scenes from NeRF \cite{mildenhall2020nerf}. Please see the supplementary material for more images.}
  \label{fig:synthetic}
\end{figure}

%% file: figures_tex/forward_facing.tex
\begin{figure}[]
  \captionsetup[subfigure]{labelformat=empty}
  \centering
  \begin{subfigure}[b]{\synthwidth}
        \includegraphics[width=\linewidth]{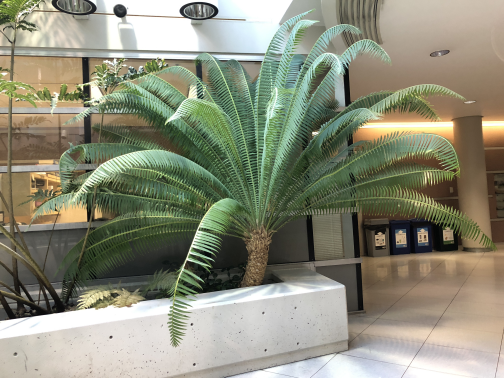}
    \end{subfigure}
    \begin{subfigure}[b]{\synthwidth}
        \includegraphics[width=\linewidth]{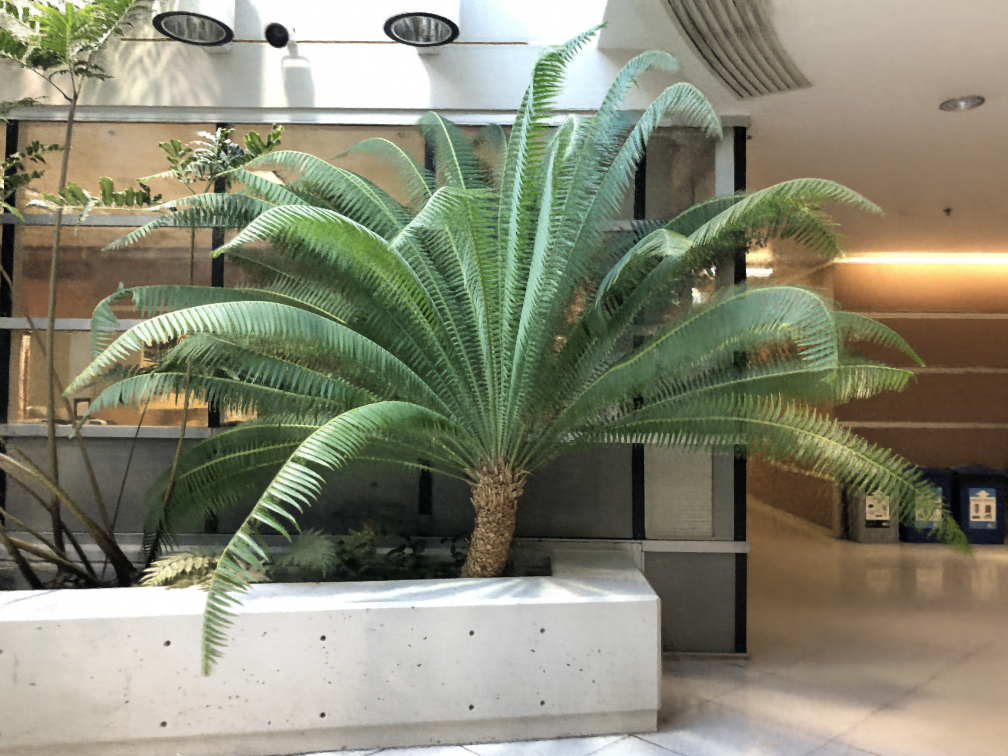}
    \end{subfigure}
    \begin{subfigure}[b]{\synthwidth}
        \includegraphics[width=\linewidth]{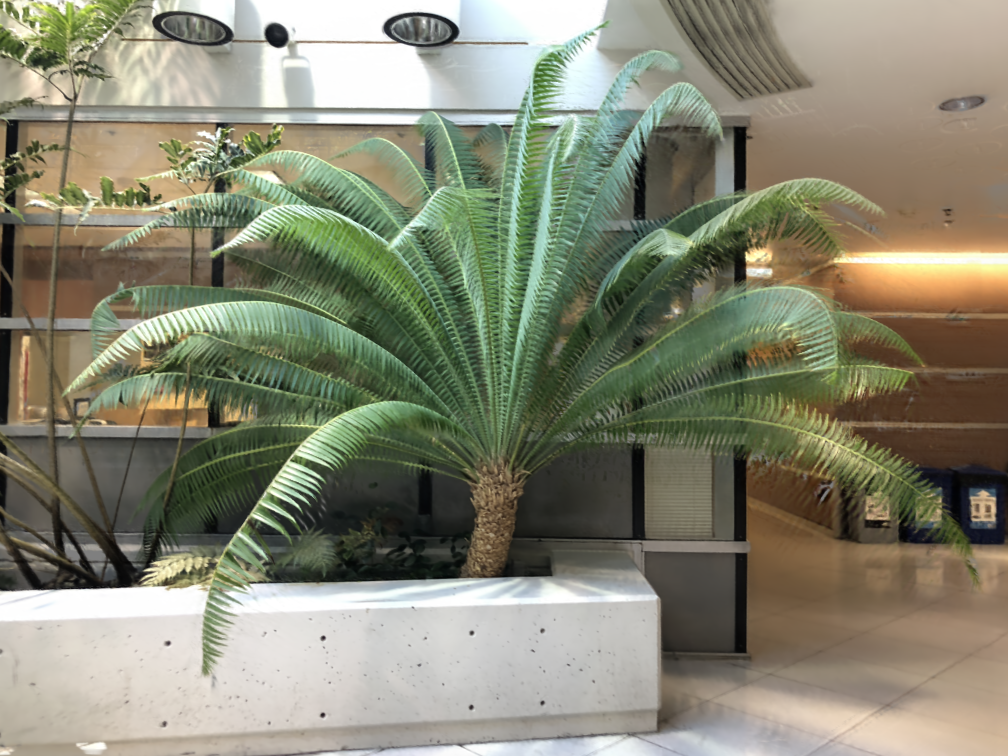}
    \end{subfigure}
    \begin{subfigure}[b]{\synthwidth}
        \includegraphics[width=\linewidth]{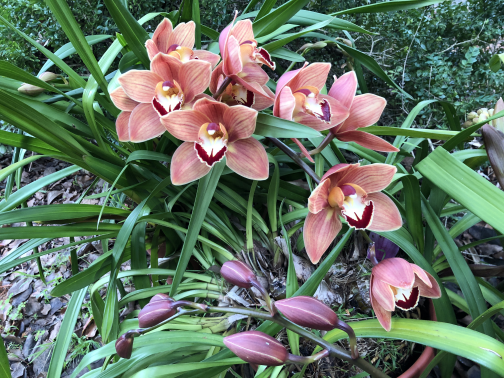}
        \caption{Ground Truth}
    \end{subfigure}
    \begin{subfigure}[b]{\synthwidth}
        \includegraphics[width=\linewidth]{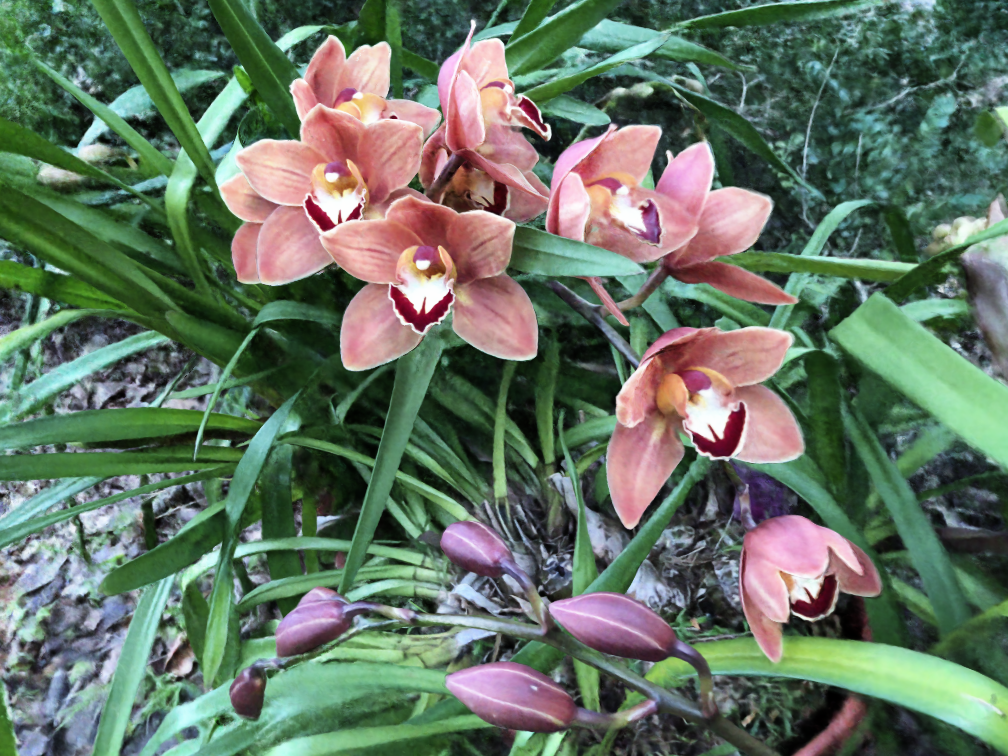}
        \caption{JAXNeRF~\cite{jaxnerf2020github, mildenhall2020nerf}}
    \end{subfigure}
    \begin{subfigure}[b]{\synthwidth}
        \includegraphics[width=\linewidth]{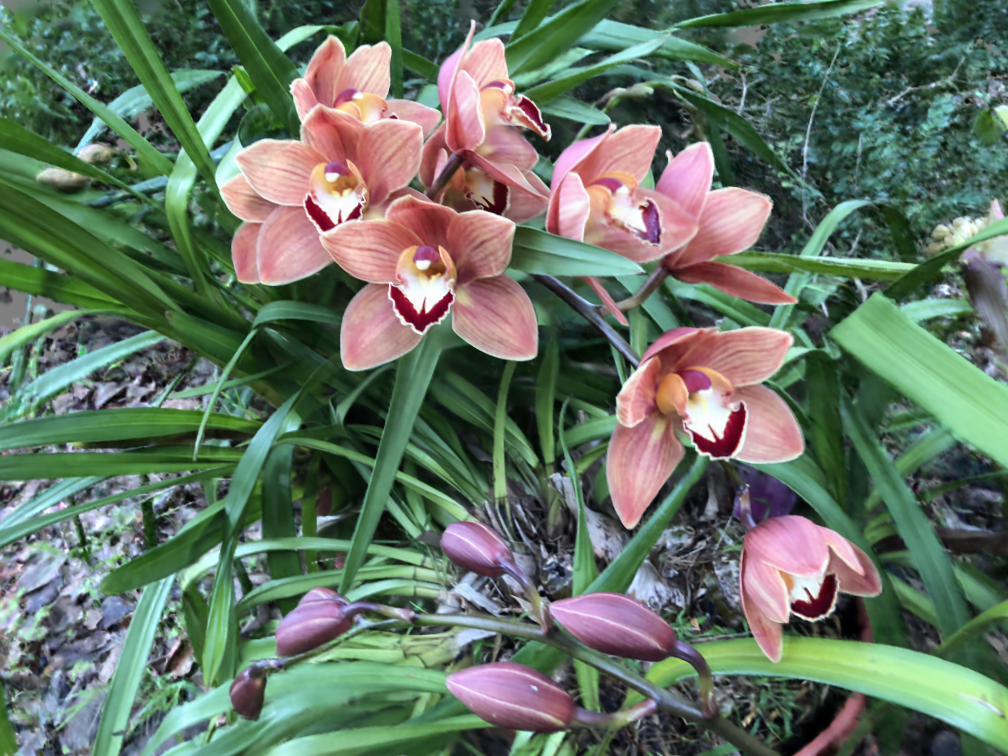}
        \caption{Plenoxels}
    \end{subfigure}
  \caption{\textbf{Real, forward-facing scenes.} Example results on the fern and orchid forward-facing scenes from NeRF.}
  \label{fig:forward-facing}
\end{figure}

%% file: figures_tex/360.tex
\newcommand{\playleft}{.2\width}
\newcommand{\playright}{\width-.35\width}
\newcommand{\playbottom}{.4\height}
\newcommand{\playtop}{\height-.7\height}

\newcommand{\playlefttwo}{.35\width}
\newcommand{\playrighttwo}{\width-.5\width}
\newcommand{\playbottomtwo}{.2\height}
\newcommand{\playtoptwo}{\height-.5\height}

\newcommand{\playleftthree}{.55\width}
\newcommand{\playrightthree}{\width-.7\width}
\newcommand{\playbottomthree}{.6\height}
\newcommand{\playtopthree}{\height-.9\height}

\newcommand{\truckleft}{.2\width}
\newcommand{\truckright}{\width-.35\width}
\newcommand{\truckbottom}{.2\height}
\newcommand{\trucktop}{\height-.5\height}

\newcommand{\trucklefttwo}{.6\width}
\newcommand{\truckrighttwo}{\width-.7\width}
\newcommand{\truckbottomtwo}{.8\height}
\newcommand{\trucktoptwo}{\height-1.0\height}

\newcommand{\truckleftthree}{.7\width}
\newcommand{\truckrightthree}{\width-.8\width}
\newcommand{\truckbottomthree}{.5\height}
\newcommand{\trucktopthree}{\height-.7\height}

\begin{figure*}
\centering
\captionsetup[subfigure]{labelformat=empty}
\begin{subfigure}[b]{.3\linewidth}
    \centering
    \begin{subfigure}[b]{\linewidth}
   \includegraphics[width=\linewidth]{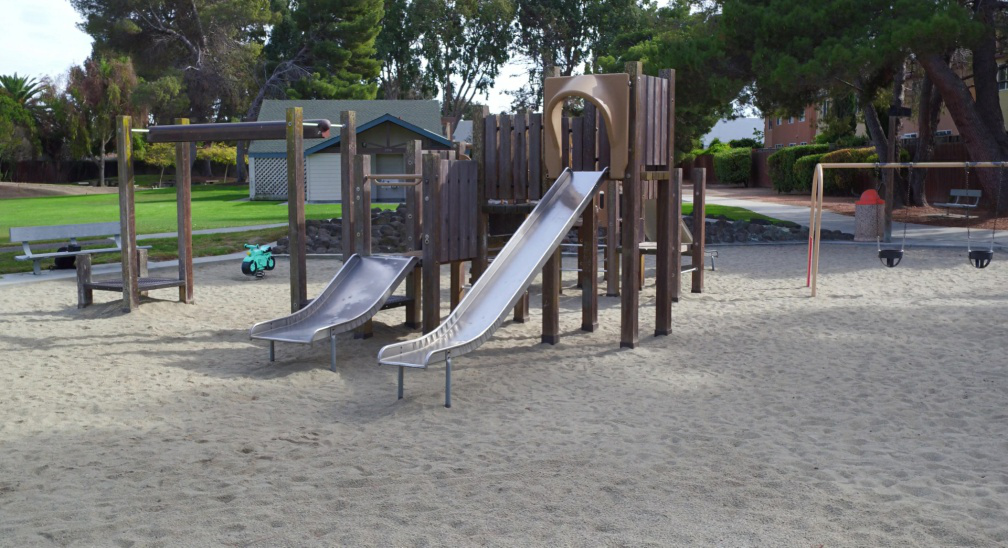}
    \end{subfigure}
    
    \begin{subfigure}[b]{.322\linewidth}
    \adjincludegraphics[trim={{\playleft} {\playbottom} {\playright} {\playtop}}, clip, width=\linewidth]{figures/360/gt/playground.png}
    \end{subfigure}
    \begin{subfigure}[b]{.322\linewidth}
    \adjincludegraphics[trim={{\playlefttwo} {\playbottomtwo} {\playrighttwo} {\playtoptwo}}, clip, width=\linewidth]{figures/360/gt/playground.png}
    \end{subfigure}
    \begin{subfigure}[b]{.322\linewidth}
    \adjincludegraphics[trim={{\playleftthree} {\playbottomthree} {\playrightthree} {\playtopthree}}, clip, width=\linewidth]{figures/360/gt/playground.png}
    \end{subfigure}
\end{subfigure}
\begin{subfigure}[b]{.3\linewidth}
    \centering
    \begin{subfigure}[b]{\linewidth}
    \includegraphics[width=\linewidth]{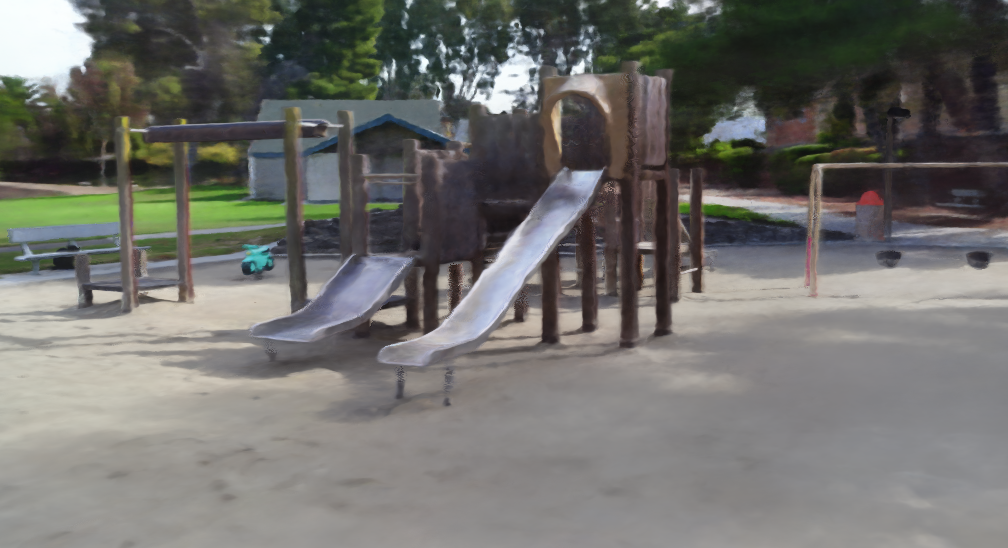}
    \end{subfigure}
    
    \begin{subfigure}[b]{.322\linewidth}
    \adjincludegraphics[trim={{\playleft} {\playbottom} {\playright} {\playtop}}, clip, width=\linewidth]{figures/360/nerfpp/playground.png}
    \end{subfigure}
    \begin{subfigure}[b]{.322\linewidth}
    \adjincludegraphics[trim={{\playlefttwo} {\playbottomtwo} {\playrighttwo} {\playtoptwo}}, clip, width=\linewidth]{figures/360/nerfpp/playground.png}
    \end{subfigure}
    \begin{subfigure}[b]{.322\linewidth}
    \adjincludegraphics[trim={{\playleftthree} {\playbottomthree} {\playrightthree} {\playtopthree}}, clip, width=\linewidth]{figures/360/nerfpp/playground.png}
    \end{subfigure}
\end{subfigure}
\begin{subfigure}[b]{.3\linewidth}
    \centering
    \begin{subfigure}[b]{\linewidth}
    \includegraphics[width=\linewidth]{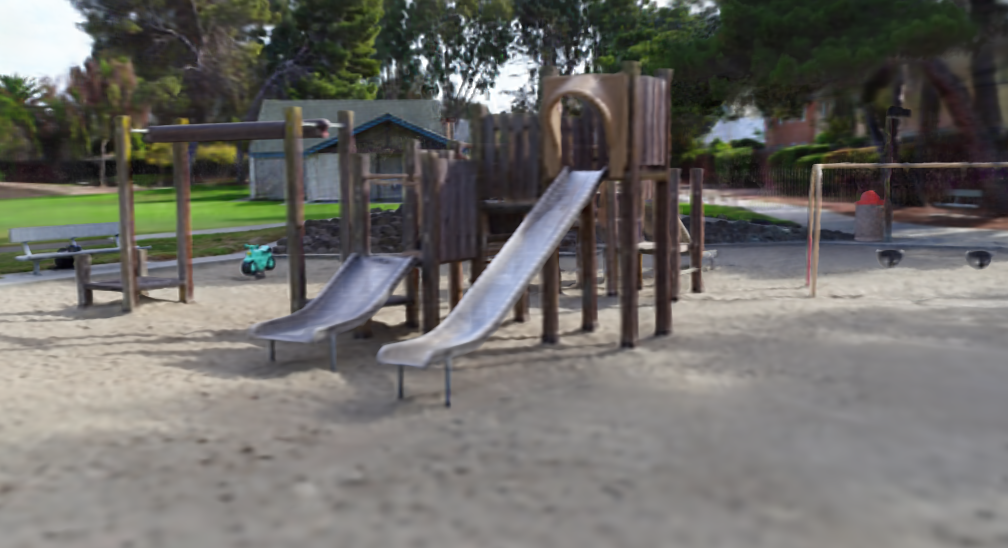}
    \end{subfigure}
    
    \begin{subfigure}[b]{.322\linewidth}
    \adjincludegraphics[trim={{\playleft} {\playbottom} {\playright} {\playtop}}, clip, width=\linewidth]{figures/360/ours/playground.png}
    \end{subfigure}
    \begin{subfigure}[b]{.322\linewidth}
    \adjincludegraphics[trim={{\playlefttwo} {\playbottomtwo} {\playrighttwo} {\playtoptwo}}, clip, width=\linewidth]{figures/360/ours/playground.png}
    \end{subfigure}
    \begin{subfigure}[b]{.322\linewidth}
    \adjincludegraphics[trim={{\playleftthree} {\playbottomthree} {\playrightthree} {\playtopthree}}, clip, width=\linewidth]{figures/360/ours/playground.png}
    \end{subfigure}
\end{subfigure}

\vspace{5pt}

\begin{subfigure}[b]{.3\linewidth}
    \centering
    \begin{subfigure}[b]{\linewidth}
    \includegraphics[width=\linewidth]{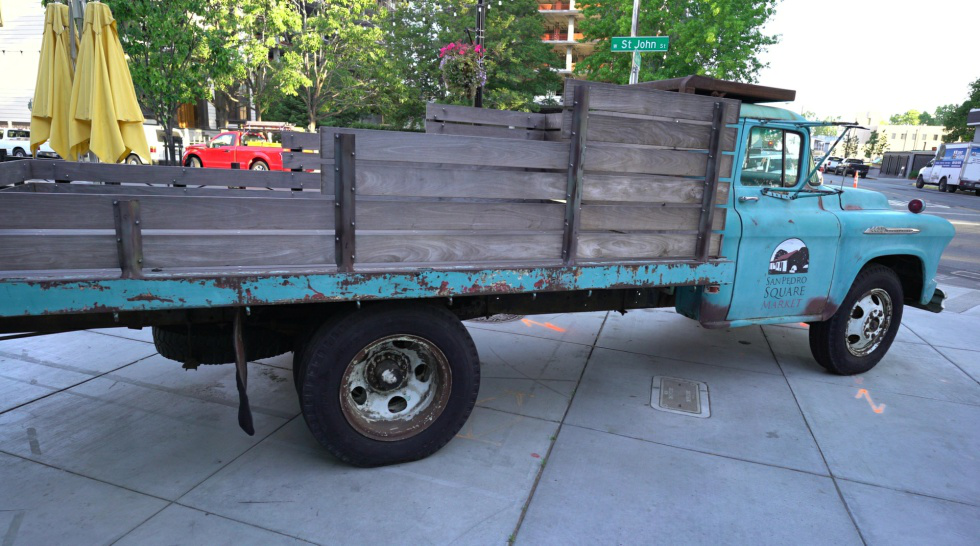}
    \end{subfigure}
    
    \begin{subfigure}[b]{.322\linewidth}
    \adjincludegraphics[trim={{\truckleft} {\truckbottom} {\truckright} {\trucktop}}, clip, width=\linewidth]{figures/360/gt/truck_2.png}
    \end{subfigure}
    \begin{subfigure}[b]{.322\linewidth}
    \adjincludegraphics[trim={{\trucklefttwo} {\truckbottomtwo} {\truckrighttwo} {\trucktoptwo}}, clip, width=\linewidth]{figures/360/gt/truck_2.png}
    \end{subfigure}
    \begin{subfigure}[b]{.322\linewidth}
    \adjincludegraphics[trim={{\truckleftthree} {\truckbottomthree} {\truckrightthree} {\trucktopthree}}, clip, width=\linewidth]{figures/360/gt/truck_2.png}
    \end{subfigure}
\caption{Ground Truth}
\end{subfigure}
\begin{subfigure}[b]{.3\linewidth}
    \centering
    \begin{subfigure}[b]{\linewidth}
    \includegraphics[width=\linewidth]{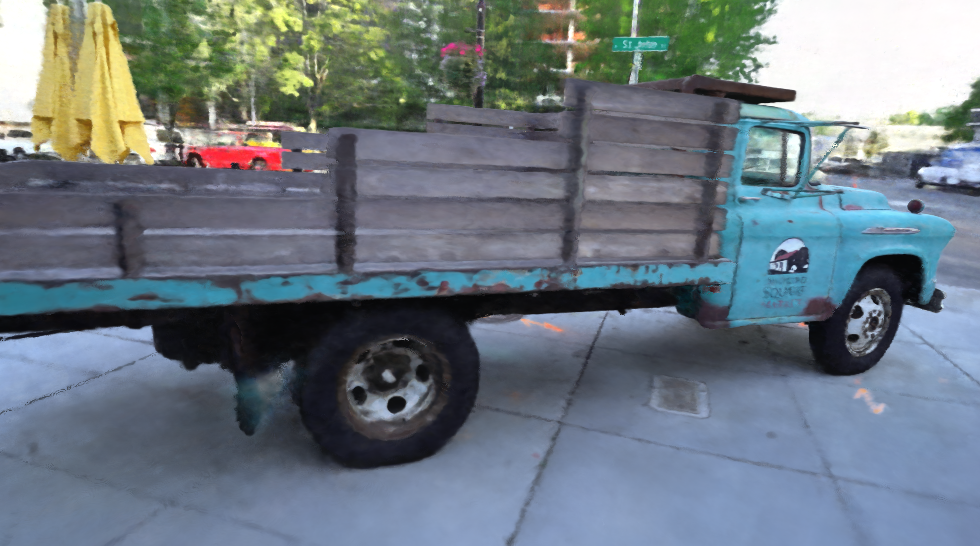}
    \end{subfigure}
    
    \begin{subfigure}[b]{.322\linewidth}
    \adjincludegraphics[trim={{\truckleft} {\truckbottom} {\truckright} {\trucktop}}, clip, width=\linewidth]{figures/360/nerfpp/truck_2.png}
    \end{subfigure}
    \begin{subfigure}[b]{.322\linewidth}
    \adjincludegraphics[trim={{\trucklefttwo} {\truckbottomtwo} {\truckrighttwo} {\trucktoptwo}}, clip, width=\linewidth]{figures/360/nerfpp/truck_2.png}
    \end{subfigure}
    \begin{subfigure}[b]{.322\linewidth}
    \adjincludegraphics[trim={{\truckleftthree} {\truckbottomthree} {\truckrightthree} {\trucktopthree}}, clip, width=\linewidth]{figures/360/nerfpp/truck_2.png}
    \end{subfigure}
\caption{NeRF++~\cite{zhang2020nerf}}
\end{subfigure}
\begin{subfigure}[b]{.3\linewidth}
    \centering
    \begin{subfigure}[b]{\linewidth}
    \includegraphics[width=\linewidth]{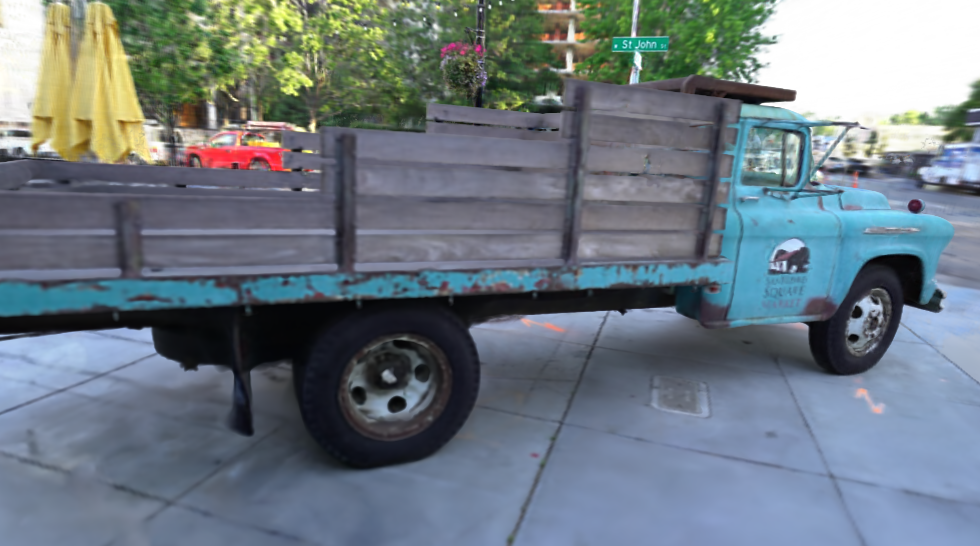}
    \end{subfigure}
    
    \begin{subfigure}[b]{.322\linewidth}
    \adjincludegraphics[trim={{\truckleft} {\truckbottom} {\truckright} {\trucktop}}, clip, width=\linewidth]{figures/360/ours/truck_2.png}
    \end{subfigure}
    \begin{subfigure}[b]{.322\linewidth}
    \adjincludegraphics[trim={{\trucklefttwo} {\truckbottomtwo} {\truckrighttwo} {\trucktoptwo}}, clip, width=\linewidth]{figures/360/ours/truck_2.png}
    \end{subfigure}
    \begin{subfigure}[b]{.322\linewidth}
    \adjincludegraphics[trim={{\truckleftthree} {\truckbottomthree} {\truckrightthree} {\trucktopthree}}, clip, width=\linewidth]{figures/360/ours/truck_2.png}
    \end{subfigure}
\caption{Plenoxels}
\end{subfigure}

\caption{\textbf{Real, $360^\circ$ scenes.} Example results on the playground and truck $360^{\circ}$ scenes from Tanks and Temples \cite{tanks}.}
\label{fig:360}
\end{figure*}

%% file: figures_tex/drumfail.tex

\newcommand{\drumsleft}{.55\width}
\newcommand{\drumsright}{.15\width}
\newcommand{\drumsbottom}{.35\height}
\newcommand{\drumstop}{.35\height}

\begin{figure}[]
  \captionsetup[subfigure]{labelformat=empty}
  \centering
  \begin{subfigure}[b]{\synthwidth}
    \adjincludegraphics[trim={{\drumsleft} {\drumsbottom} {\drumsright} {\drumstop}}, clip, width=\linewidth]{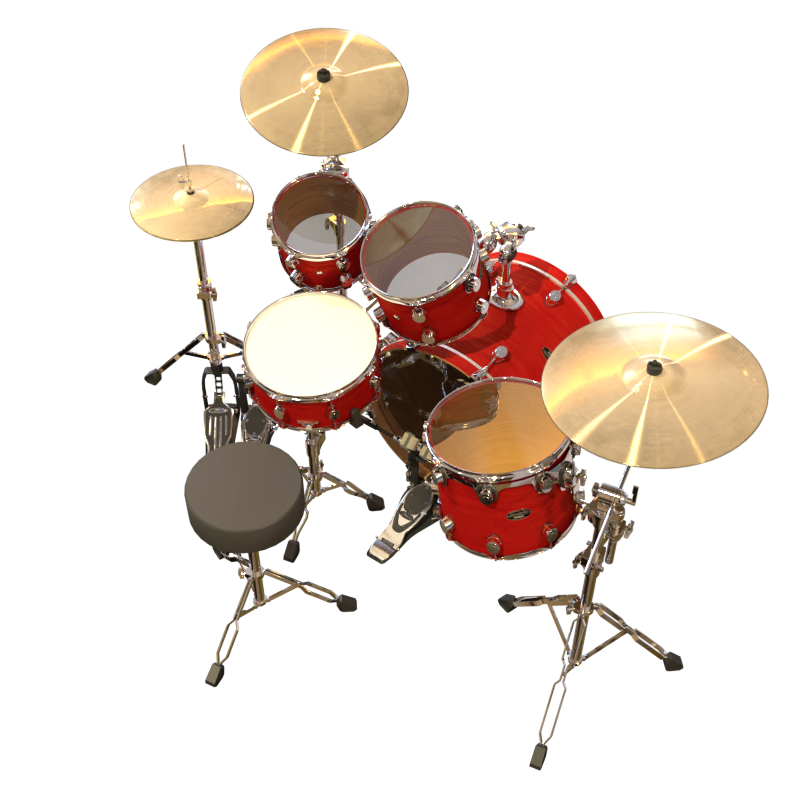}
    \caption{Ground Truth}
  \end{subfigure}
  \begin{subfigure}[b]{\synthwidth}
    \adjincludegraphics[trim={{\drumsleft} {\drumsbottom} {\drumsright} {\drumstop}}, clip, width=\linewidth]{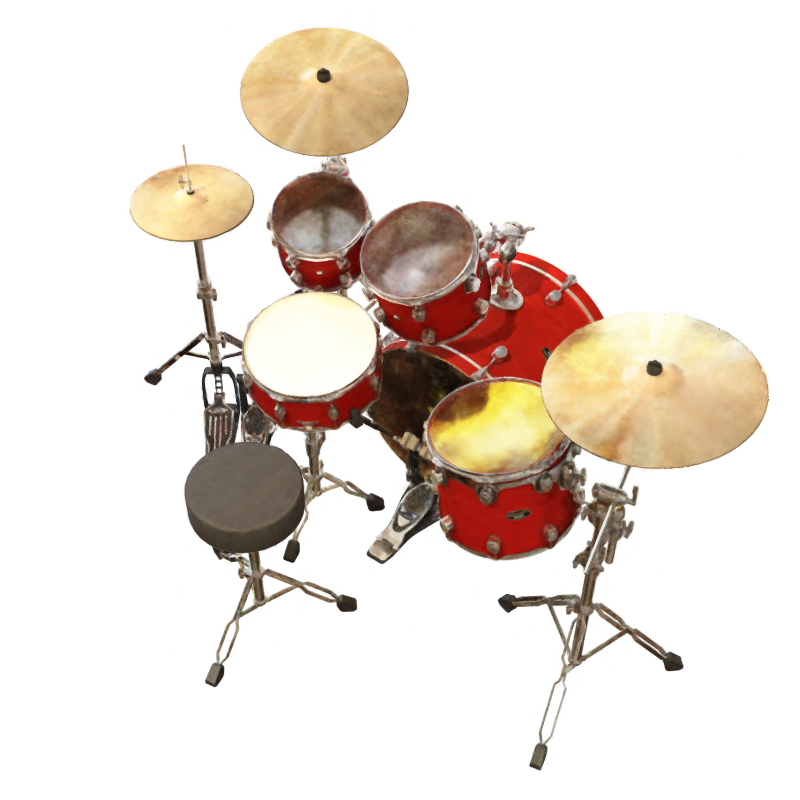}
    \caption{JAXNeRF~\cite{jaxnerf2020github, mildenhall2020nerf}}
  \end{subfigure}
  \begin{subfigure}[b]{\synthwidth}
    \adjincludegraphics[trim={{\drumsleft} {\drumsbottom} {\drumsright} {\drumstop}}, clip, width=\linewidth]{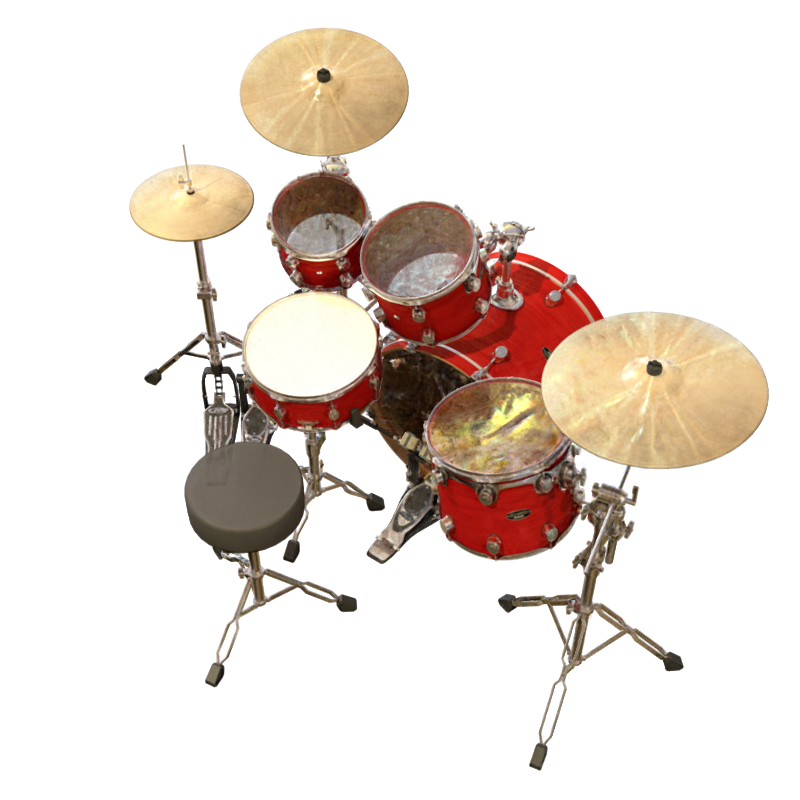}
    \caption{Plenoxels}
  \end{subfigure}
  \caption{\textbf{Artifacts.} JAXNeRF and Plenoxel models both exhibit artifacts, but the artifacts are different, as shown here in the specularities in the synthetic drums scene. Note that some artifacts are unavoidable for any underdetermined inverse problem, but the specific artifacts vary depending on the priors induced by the model and regularizer.}
    \label{fig:artifacts}
\end{figure}

%% file: supp.tex
\appendix

\setcounter{page}{1}

\twocolumn[
\centering
\Large
\textbf{Plenoxels: Radiance Fields without Neural Networks} \\
\vspace{0.5em}Supplementary Material \\
\vspace{1.0em}
] 
\appendix

\section{Overview}
In the supplementary material, we include additional experimental details and present results and visualizations of further ablation studies. We also present full, per-scene quantitative and visual comparisons between our method and prior work. We encourage the reader to see the video for results of our method on a wide range of scenes.

\section{Experimental Details}

\subsection{Implementation Details}
As briefly discussed in~\cref{sec:grid}, we use a simple data structure
which consists of a data table in addition to a dense grid, where each cell is either
\textsf{NULL} or a pointer into the data table.
Each entry in the data table consists of the density value and the SH coefficients for each of the RGB color channels.
\textsf{NULL} cells are considered to have all 0 values.
This data structure allows for reasonably efficient trilinear interpolation both in the forward and backward passes while maintaining sparsity; due
to the relatively large memory requirements to store the SH coefficients, gradients, and RMSProp running averages, the dense pointer grid is usually not dominant in size.
Nevertheless, reading the pointers currently appears to
take a significant amount of rendering time, and optimizations are likely possible.

Our main CUDA rendering and gradient kernels simultaneously parallelize across rays, colors, and SH coefficients. Each CUDA warp (32 threads) handles one ray, with threads processing one SH coefficient each; since coefficients are stored contiguously, this means access to global memory is highly coalesced. The SH coefficients are combined into colors using warp-level operations from NVIDIA CUB~\cite{nvidiacub}. These features are particularly significant in the case of trilinear interpolation.

Note that in order to correctly perform trilinear color interpolation, instead of using the sigmoid function to ensure that predicted sample colors are always between 0 and 1 as in NeRF \cite{mildenhall2020nerf},
we simply clip negative color values to 0 with a ReLU to preserve linearity as much as possible. 

We use weight-based thresholding (as in PlenOctrees \cite{yu2021plenoctrees}) for the synthetic and real, $360^\circ$ scenes, and opacity-based thresholding for the forward-facing scenes. 
The reason for this is that some content (especially at the edges) in the forward-facing scenes is not visible in most of the training views, so weight-based thresholding tends to prune these sparsely-supervised features.

We use a batch size of 5000 rays and optimize with RMSProp \cite{hinton}. For $\sigma$ we use the same delayed exponential learning rate schedule as Mip-NeRF \cite{barron2021mipnerf}, where the exponential is scaled by a learning rate of 30 (this is where the exponential would start, if not for the delay) and decays to 0.05 at step 250000, with an initial delay period of 15000 steps. For SH we use a pure exponential decay learning rate schedule, with an initial learning rate of 0.01 that decays to \num{5e-6} at step 250000. 

The TV losses are evaluated stochastically; they are applied only to 1\% of all voxels in the grid in each step. Note that empty voxels can be selected, as their neighbors may not be empty. 
In practice, for performance reasons, we always apply the TV regularization on random contiguous segments of voxels (in the order that the pointer grid is stored). This is much faster to evaluate on the GPU due to locality.   In all cases, the voxel differences in the TV loss defined below \cref{eq:loss} is in practice normalized by the voxel resolution in each dimension, relative to 256 (for historical reasons):
\begin{equation}
\Delta_x((i, j, k), d) = \frac{|V_d (i + 1, j, k)  - V_d (i, j, k)|}{256 / D_x}
\end{equation}
Where $D_x$ is the grid resolution in the $x$ dimension, and $V_d (i, j, k)$ is the $d$th value of voxel $(i, j, k)$ (either density or a SH coefficient).
We scale $\Delta_y, \Delta_z$ analogously. Note that the same loss is applied in NDC and to the background model, except in the background model, the TV also wraps around the edges of the equirectangular image.
For SH, empty grid cells and edges are considered to have the same value as the current cell (instead of 0) for purposes of TV.

\subsection{Synthetic experiments}
On the synthetic scenes, we found that our method performs nearly identically when TV regularization is present only in the first stage of optimization; turning off the regularization after pruning voxels and increasing resolution reduces our training time modestly. We suspect (see \cref{tab:ntrainablate}) this is due to the large number of training views (100) available for these scenes as well as the low level of noise; for the other datasets we retain TV regularization throughout optimization.

We start at resolution $256^3$, prune and upsample to resolution $512^3$ after 38400 steps (the equivalent of 3 epochs), and optimize for a total of 128000 steps (the equivalent of 10 epochs). We prune using a weight threshold of 0.256, and use $\lambda_{TV}$ of \num{1e-5} for $\sigma$ and \num{1e-3} for SH, only during the initial 38400 steps (and then turn off regularization after pruning and upsampling, for faster optimization).

\subsection{Forward-facing experiments}
For the forward-facing scenes we start at resolution $256\times256\times128$, prune and upsample to resolution $512\times512\times128$ at step 38400, prune and upsample to resolution $1408\times1156\times128$ at step 76800, and optimize for a total of 128000 steps. 
The final grid resolution is derived from the image resolution of the dataset, with some padding added on each side.
We prune using a $\sigma$ threshold of 5, use $\lambda_{TV}$ of \num{5e-4} for $\sigma$ and \num{5e-3} for SH, and use a sparsity penalty $\lambda_s$ of \num{1e-12} to encourage empty voxels.

While these TV parameters work well for the forward-facing NeRF scenes,
more generally, we find that sometimes it is preferrable to use $\lambda_{TV}$ \num{5e-3} for density and \num{5e-2} for SH, which reduces artifacts while blurring the scene more. This is used for some of the examples in the video, for example the piano. In general, since scenes differ significantly in content, camera noise, and actual scale, a hyperparameter sweep of the TV weights can be helpful, and using different TV values across the scenes would improve the metrics for the NeRF scenes as well.

\subsection{$360^\circ$ experiments}
For the $360^\circ$ scenes our foreground Plenoxel grid starts at resolution $128^3$; we prune and upsample to $256^3$, $512^3$, and $640^3$ with 25600 steps in between each upsampling. We optimize for a total of 102400 steps. We prune using a weight threshold of 1.28, and use $\lambda_{TV}$ of \num{5e-5} for $\sigma$ and \num{5e-3} for SH for the inner grid and $\lambda_{TV}$ of \num{1e-3} for both $\sigma$ and SH for the 64 background grid layers of resolution $2048\times1024$. We use $\lambda_s$ of \num{1e-11} and $\lambda_\beta$ of \num{1e-5}.
For simplicity of implementation, we did not use coarse-to-fine
for the background and only use $\sigma$ thresholding.
We also do not use the delayed learning rate function for the background, opting instead to use an exponential decay to allow the background to optimize faster than the foreground at the beginning.

While the TV weights were fixed for these scenes, in general, a hyperparameter sweep of the TV weights can be helpful. For more general scenes, it is sometimes useful to use a near-bound on the camera rays (as in NeRF) to prevent floaters very close to the camera, or to only begin optimizing the foreground after, say, 1000 iterations. Further sparsity losses to encourage the weight distribution to be a delta function may also help.




\section{Ablation Studies}

We visualize ablations on the synthetic lego scene in \cref{fig:bigablate}.
In addition to comparing nearest neighbor and trilinear interpolation, we also experimented with tricubic interpolation, which produces a function approximation that is both continuous (like trilinear interpolation) and smooth. However, we found tricubic interpolation offered negligible improvements compared to trilinear, in exchange for a substantial increase in computation (this increase in computation is why we do not include a full ablation table for tricubic interpolation).

\input{figures_tex/ablations}

\begin{table}[]
\renewcommand{\tabcolsep}{2pt}
\begin{tabular}{@{}llccc@{}}
\toprule
LR Schedule &  & PSNR $\uparrow$ & SSIM $\uparrow$ & LPIPS $\downarrow$ \\ \cmidrule(r){1-1} \cmidrule(l){3-5} 
Exp for SH, Delayed for $\sigma$~\cite{barron2021mipnerf} &  & 30.57 & 0.950 & 0.065 \\
Exp for SH and $\sigma$ &  & 30.58 & 0.950 & 0.066 \\
Exp for SH, Constant for $\sigma$ &  & 30.37 & 0.948 & 0.068 \\
Constant for SH and $\sigma$ &  & 30.13 & 0.945 & 0.075 \\
\bottomrule
\end{tabular}
\caption{\textbf{Comparison of different learning rate schedules} for $\sigma$ (voxel opacity) and spherical harmonics (SH), with fixed resolution $256^3$ and RMSProp \cite{hinton}. Results are averaged over the 8 synthetic scenes from NeRF~\cite{mildenhall2020nerf}. Our method is robust to variations in learning rate schedule.}
\label{tab:lrablate}
\end{table}

\begin{table}[t]
\renewcommand{\tabcolsep}{2.5pt}
\begin{tabular}{@{}llccc@{}}
\toprule
Optimizer &  & PSNR $\uparrow$ & SSIM $\uparrow$ & LPIPS $\downarrow$ \\ \cmidrule(r){1-1} \cmidrule(l){3-5} 
RMSProp \cite{hinton} for SH and $\sigma$ &  & 30.57 & 0.950 & 0.065 \\
RMSProp for SH, SGD for $\sigma$ &  & 30.20 & 0.946 & 0.072 \\
SGD for SH, RMSProp for $\sigma$ &  & 29.82 & 0.940 & 0.076 \\
SGD for SH and $\sigma$ &  & 29.35 & 0.932 & 0.087 \\
\bottomrule
\end{tabular}
\caption{\textbf{Comparison of different optimizers} for $\sigma$ and SH, with fixed resolution $256^3$. Results are averaged over the 8 synthetic scenes from NeRF~\cite{mildenhall2020nerf}. Our method is robust to variations in optimizer, although there is a benefit to RMSProp particularly for optimizing the spherical harmonic coefficients.}
\label{tab:optimizerablate}
\end{table}

\begin{table}[]
\renewcommand{\tabcolsep}{5.5pt}
\begin{tabular}{@{}llccc@{}}
\toprule
Regularizer &  & PSNR $\uparrow$ & SSIM $\uparrow$ & LPIPS $\downarrow$ \\ \cmidrule(r){1-1} \cmidrule(l){3-5} 
TV SH, TV $\sigma$, Sparsity &  & 26.29 & 0.839 & 0.210 \\
- Sparsity &  & 26.31 & 0.839 & 0.210 \\
- TV $\sigma$ &  & 25.25 & 0.807 & 0.226 \\
- TV SH &  & 25.80 & 0.814 & 0.234 \\
\bottomrule
\end{tabular}
\caption{\textbf{Ablation over regularization.} Results are averaged over the 8 forward-facing scenes from NeRF, which are particularly sensitive to regularization due to the low number of training views. We find that the sparsity regularizer is not necessary for quality, but we retain it to reduce memory footprint. TV regularization is essential for $\sigma$ but also important for spherical harmonics, as visualized in \cref{fig:tv}, even though this effect is not as pronounced in the PSNR metric. Without any TV regularization (on SH or $\sigma$), three of the eight scenes run out of memory on our GPU.}
\label{tab:tvablate}
\end{table}

\cref{tab:lrablate} and \cref{tab:optimizerablate} show ablations over learning rate schedule and optimizer, respectively. We find that Plenoxel optimization is reasonably robust to both of these hyperparameters, although there is a noticeable improvement from using RMSProp compared to SGD, particularly for the spherical harmonic coefficients. Note that when comparing different learning rate schedules and optimizers, we tune the initial learning rate separately for each row to provide the best results possible for each configuration.

\cref{tab:tvablate} shows ablation over regularization, for the forward-facing scenes. We find that TV regularization is important for these scenes, likely due to their low number of training images. Regularization on opacity has a quantitatively larger effect than regularization on spherical harmonics, but both are important for avoiding visual artifacts (see \cref{fig:tv}).

\cref{tab:renderingcompare} compares the performance of Plenoxels when trained with the rendering formula used in NeRF (originally from Max \cite{max1995}) and when trained with the rendering formula used in Neural Volumes \cite{lombardi2019neural}. The Max formula is defined in \cref{eq:max} and rewritten here in a slightly more convenient format:

\begin{align}
T_i &= \exp \left(-\sum_{j=1}^{i-1} \sigma_j \delta_j \right) \\
\hat C(\textbf{r}) &= \sum_{i=1}^N (T_i - T_{i+1})\textbf{c}_i
\end{align}

The Neural Volumes formula is (up to optimizing in log space):
\begin{align}
    T_i &= \min \left\{1,~ \sum_{j=1}^{i-1} \exp(-\delta_i \sigma_i) \right\} \\
    \hat C(\textbf{r}) &= \sum_{i=1}^N (T_i - T_{i+1})\textbf{c}_i
\end{align}

These formulas only differ in their definition of the transmittance $T_i$. In particular, the Neural Volumes formula treats $\alpha_i$, the fraction of the ray contributed by sample $i$, as a function of the opacity and sampling distance of sample $i$ only. In contrast, the contribution of sample $i$ in the Max formula depends on the opacity of sample $i$ as well as the opacities of all preceding samples along the ray. In essence, opacity in the Neural Volumes formula is absolute and ray-independent (except for clipping the total contribution to 1), whereas opacity in the Max formula denotes the fraction of incoming light that each sample absorbs, a ray-dependent quantity. As we show in \cref{tab:renderingcompare}, the Max formula results in substantially better performance; we suspect this difference is due to its more physically-accurate modeling of transmittance.

\section{Per-Scene Results}

\subsection{Synthetic, Bounded Scenes}

Full, per-scene results for the 8 synthetic scenes from NeRF are presented in \cref{tab:fullsynthetic} and \cref{fig:fullsynthetic}. Note that the values for JAXNeRF are from our own rerunning with centered pixels (we ran JAXNeRF in parallel across 4 GPUs and multiplied the times by 4 to account for this parallelization).

\subsection{Real, Forward-Facing Scenes}

Full, per-scene results for the 8 forward-facing scenes from NeRF are presented in \cref{tab:fullforward}. Note that the values for JAXNeRF are from our own rerunning with centered pixels (we ran JAXNeRF in parallel across 4 GPUs and multiplied the times by 4 to account for this parallelization).

\subsection{Real, $360^{\circ}$ Scenes}

Full, per-scene results for the four $360^\circ$ scenes from Tanks and Temples \cite{tanks} are presented in \cref{tab:full360}. Note that the values for NeRF++ appear slightly different from the paper; we
re-evaluated the metrics independently using VGG LPIPS and standard SSIM, from rendered images shared by the original authors. 

\renewcommand{\tabcolsep}{2pt}
\begin{table}[h]
  \centering
  \begin{tabular}{llcccclc}
    \multicolumn{8}{c}{PSNR $\uparrow$} \\
    \toprule
     & & M60 & Playground & Train & Truck & & Mean \\ \cmidrule(){1-1} \cmidrule(){3-7} \cmidrule(){8-8}
Ours & & 17.93 & 23.03 & 17.97 & 22.67 & & 20.40 \\
NeRF++~\cite{zhang2020nerf} & & 18.49 & 22.93 & 17.77 & 22.77 & & 20.49 \\
    \bottomrule
 & & & & & & & \\ 
    \multicolumn{8}{c}{SSIM $\uparrow$} \\
    \toprule
     & & M60 & Playground & Train & Truck & & Mean \\ \cmidrule(){1-1} \cmidrule(){3-7} \cmidrule(){8-8}
Ours & & 0.687 & 0.712 & 0.629 & 0.758 & & 0.696 \\
NeRF++ & & 0.650 & 0.672 & 0.558 & 0.712 & & 0.648 \\
    \bottomrule
 & & & & & & & \\ 
    \multicolumn{8}{c}{LPIPS $\downarrow$} \\
    \toprule
     & & M60 & Playground & Train & Truck & & Mean \\ \cmidrule(){1-1} \cmidrule(){3-7} \cmidrule(){8-8}
Ours & & 0.439 & 0.435 & 0.443 & 0.364 & & 0.420 \\
NeRF++ & & 0.481 & 0.477 & 0.531 & 0.424 & & 0.478 \\
    \bottomrule
 & & & & & & & \\ 
    \multicolumn{8}{c}{Optimization Time $\downarrow$} \\
    \toprule
     & & M60 & Playground & Train & Truck & & Mean \\ \cmidrule(){1-1} \cmidrule(){3-7} \cmidrule(){8-8}
Ours & & 25.5m & 26.3m & 29.5m & 28.0m & & 27.3m \\
    \bottomrule
  \end{tabular}
\newline
  \caption{\textbf{Full results on $360^\circ$ scenes.}}
  \label{tab:full360}
\end{table}

\renewcommand{\tabcolsep}{6pt}
\begin{table*}[]
  \centering
  \begin{tabular}{llcccccccclc}
    \multicolumn{12}{c}{PSNR $\uparrow$} \\
    \toprule
     & & Chair & Drums & Ficus & Hotdog & Lego & Materials & Mic & Ship & & Mean \\ \cmidrule(){1-1} \cmidrule(){3-10} \cmidrule(){12-12}
Ours & & 33.98 & 25.35 & 31.83 & 36.43 & 34.10 & 29.14 & 33.26 & 29.62 & & 31.71 \\
NV~\cite{lombardi2019neural} & & 28.33 & 22.58 & 24.79 & 30.71 & 26.08 & 24.22 & 27.78 & 23.93 & & 26.05 \\
JAXNeRF~\cite{jaxnerf2020github, mildenhall2020nerf} & & 34.20 & 25.27 & 31.15 & 36.81 & 34.02 & 30.30 & 33.72 & 29.33 & & 31.85 \\
    \bottomrule
 & & & & & & & & & & & \\ 
    \multicolumn{12}{c}{SSIM $\uparrow$} \\
    \toprule
     & & Chair & Drums & Ficus & Hotdog & Lego & Materials & Mic & Ship & & Mean \\ \cmidrule(){1-1} \cmidrule(){3-10} \cmidrule(){12-12}
Ours & & 0.977 & 0.933 & 0.976 & 0.980 & 0.975 & 0.949 & 0.985 & 0.890 & & 0.958 \\
NV & & 0.916 & 0.873 & 0.910 & 0.944 & 0.880 & 0.888 & 0.946 & 0.784 &  &0.893 \\
JAXNeRF & & 0.975 & 0.929 & 0.970 & 0.978 & 0.970 & 0.955 & 0.983 & 0.868 & & 0.954 \\
    \bottomrule
 & & & & & & & & & & & \\ 
    \multicolumn{12}{c}{LPIPS $\downarrow$} \\
    \toprule
     & & Chair & Drums & Ficus & Hotdog & Lego & Materials & Mic & Ship & & Mean \\ \cmidrule(){1-1} \cmidrule(){3-10} \cmidrule(){12-12}
Ours & & 0.031 & 0.067 & 0.026 & 0.037 & 0.028 & 0.057 & 0.015 & 0.134 & & 0.049 \\
NV & & 0.109 & 0.214 & 0.162 & 0.109 & 0.175 & 0.130 & 0.107 & 0.276 & & 0.160 \\
JAXNeRF & & 0.036 & 0.085 & 0.037 & 0.074 & 0.068 & 0.057 & 0.023 & 0.192 & & 0.072 \\
    \bottomrule
 & & & & & & & & & & & \\ 
    \multicolumn{12}{c}{Optimization Time $\downarrow$} \\
    \toprule
     & & Chair & Drums & Ficus & Hotdog & Lego & Materials & Mic & Ship & & Mean \\ \cmidrule(){1-1} \cmidrule(){3-10} \cmidrule(){12-12}
Ours & & 9.6m & 9.8m & 8.8m & 12.5m & 10.8m & 11.0m & 8.2m & 18.0m & & 11.1m \\
JAXNeRF & & 37.8h & 37.8h & 37.7h & 38.0h & 26.0h & 38.1h & 37.8h & 26.0h & & 34.9h \\
    \bottomrule
  \end{tabular}
\newline
  \caption{\textbf{Full results on synthetic scenes.}}
  \label{tab:fullsynthetic}
\end{table*}

\input{figures_tex/fullsynthetic}

\renewcommand{\tabcolsep}{6pt}
\begin{table*}[]
  \centering
  \begin{tabular}{llcccccccclc}
    \multicolumn{12}{c}{PSNR $\uparrow$} \\
    \toprule
     & & Fern & Flower & Fortress & Horns & Leaves & Orchids & Room & T-Rex & & Mean \\ \cmidrule(){1-1} \cmidrule(){3-10} \cmidrule(){12-12}
Ours & & 25.46 & 27.83 & 31.09 & 27.58 & 21.41 & 20.24 & 30.22 & 26.48 & & 26.29 \\
LLFF~\cite{mildenhall2019local} & & 28.42 & 22.85 & 19.52 & 29.40 & 18.52 & 25.46 & 24.15 & 24.70 & & 24.13 \\
JAXNeRF~\cite{jaxnerf2020github, mildenhall2020nerf} & & 25.20 & 27.80 & 31.57 & 27.70 & 21.10 & 20.37 & 32.81 & 27.12 & & 26.71 \\
    \bottomrule
 & & & & & & & & & & & \\ 
    \multicolumn{12}{c}{SSIM $\uparrow$} \\
    \toprule
     & & Fern & Flower & Fortress & Horns & Leaves & Orchids & Room & T-Rex & & Mean \\ \cmidrule(){1-1} \cmidrule(){3-10} \cmidrule(){12-12}
Ours & & 0.832 & 0.862 & 0.885 & 0.857 & 0.760 & 0.687 & 0.937 & 0.890 & & 0.839 \\
LLFF & & 0.932 & 0.753 & 0.697 & 0.872 & 0.588 & 0.844 & 0.857 & 0.840 & & 0.798 \\
JAXNeRF & & 0.798 & 0.840 & 0.890 & 0.840 & 0.703 & 0.649 & 0.952 & 0.890 & & 0.820 \\
    \bottomrule
 & & & & & & & & & & & \\ 
    \multicolumn{12}{c}{LPIPS $\downarrow$} \\
    \toprule
     & & Fern & Flower & Fortress & Horns & Leaves & Orchids & Room & T-Rex & & Mean \\ \cmidrule(){1-1} \cmidrule(){3-10} \cmidrule(){12-12}
Ours & & 0.224 & 0.179 & 0.180 & 0.231 & 0.198 & 0.242 & 0.192 & 0.238 & & 0.210 \\
LLFF & & 0.155 & 0.247 & 0.216 & 0.173 & 0.313 & 0.174 & 0.222 & 0.193 & & 0.212 \\
JAXNeRF & & 0.272 & 0.198 & 0.151 & 0.249 & 0.305 & 0.307 & 0.164 & 0.235 & & 0.235 \\
    \bottomrule
 & & & & & & & & & & & \\ 
    \multicolumn{12}{c}{Optimization Time $\downarrow$} \\
    \toprule
     & & Fern & Flower & Fortress & Horns & Leaves & Orchids & Room & T-Rex & & Mean \\ \cmidrule(){1-1} \cmidrule(){3-10} \cmidrule(){12-12}
Ours & & 23.7m & 22.0m & 31.2m & 26.3m & 13.3m & 23.4m & 28.8m & 24.8m & & 24.2m \\
JAXNeRF & & 38.9h & 38.8h & 38.6h & 38.7h & 38.8h & 38.7h & 39.1h & 38.6h & & 38.8h \\
    \bottomrule
  \end{tabular}
\newline
  \caption{\textbf{Full results on forward-facing scenes.}}
  \label{tab:fullforward}
\end{table*}

\input{figures_tex/fullforward}

\input{figures_tex/full360}

%% file: figures_tex/ablations.tex
\newcommand{\awidth}{0.16\linewidth}

\newcommand{\plotbig}[1]{\adjincludegraphics[trim={{0.2\width} {0} {0.2\width} {0}}, clip, width=\linewidth]{#1}}

\newcommand{\plotcrop}[1]{\adjincludegraphics[trim={{0.45\width} {0.45\width} {0.45\width} {0.45\width}}, clip, width=\linewidth]{#1}}

\begin{figure*}[t]
  \centering
  \begin{subfigure}[b]{\awidth}
  \plotbig{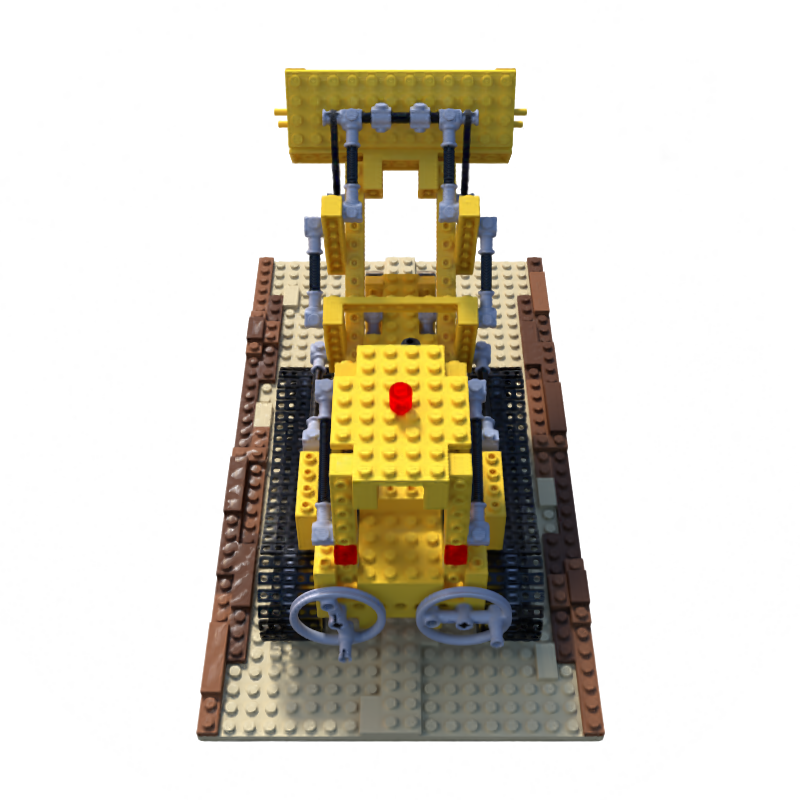}
    \plotcrop{figures/synthetic/ablations/lego256.png}
  \caption{Trilinear, $256^3$}
\end{subfigure}
  \begin{subfigure}[b]{\awidth}
  \plotbig{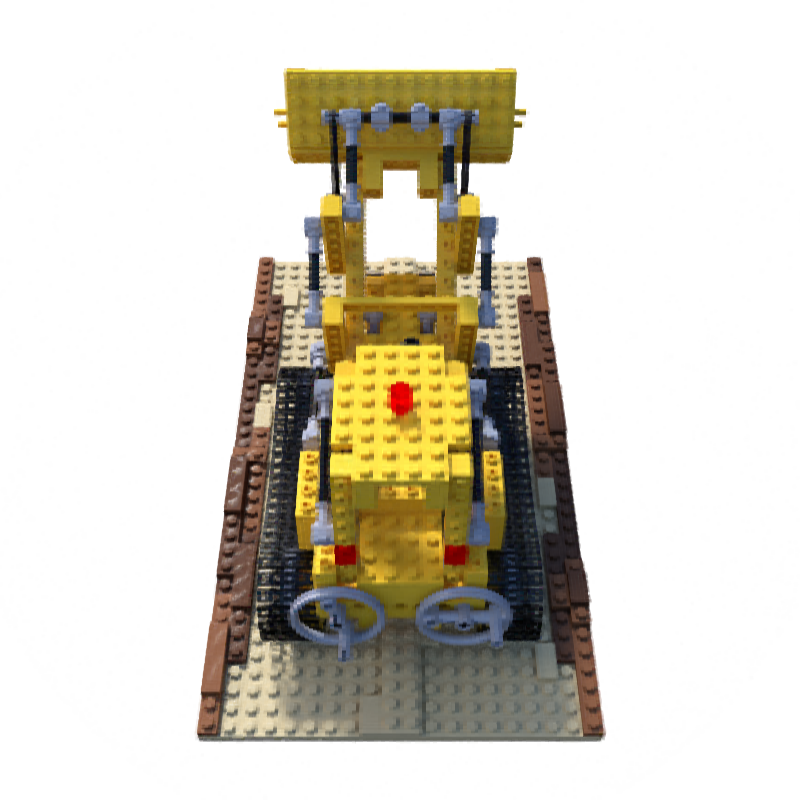}
    \plotcrop{figures/synthetic/ablations/legonn.png}
  \caption{Nearest, $256^3$}
\end{subfigure}
  \begin{subfigure}[b]{\awidth}
  \plotbig{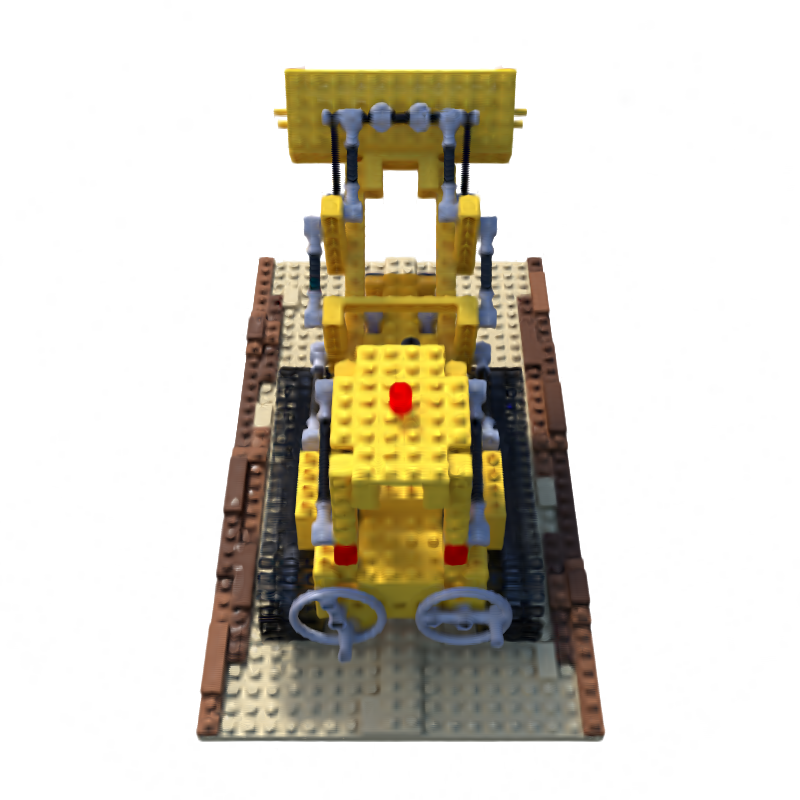}
     \plotcrop{figures/synthetic/ablations/lego128.png}
  \caption{Trilinear, $128^3$}
\end{subfigure}
\begin{subfigure}[b]{\awidth}
\plotbig{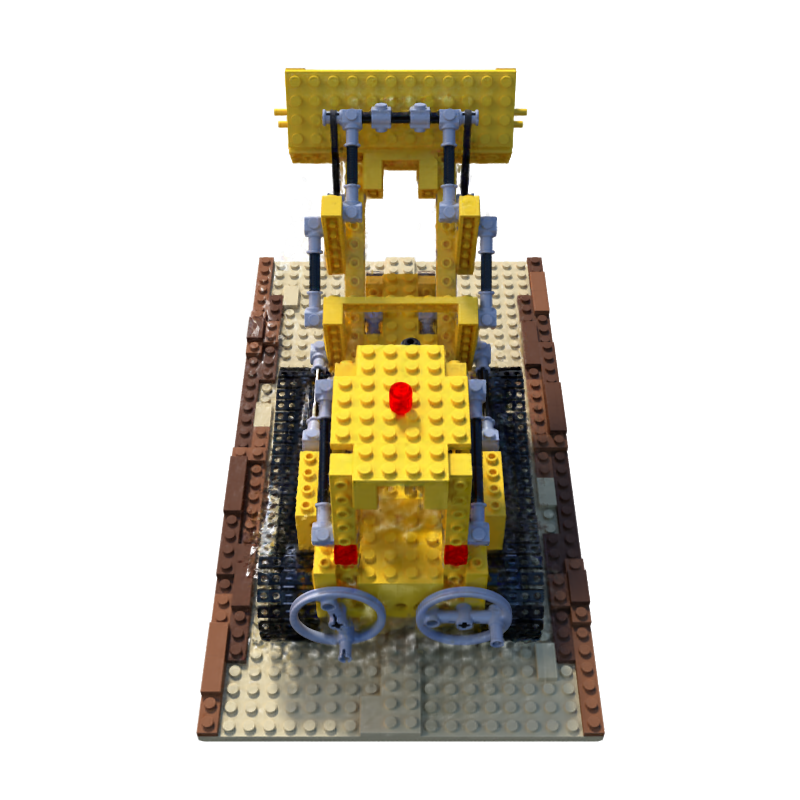}
  \plotcrop{figures/synthetic/ablations/lego25lotv.png}
  \caption{25 Images, Low TV}
\end{subfigure}
\begin{subfigure}[b]{\awidth}
\plotbig{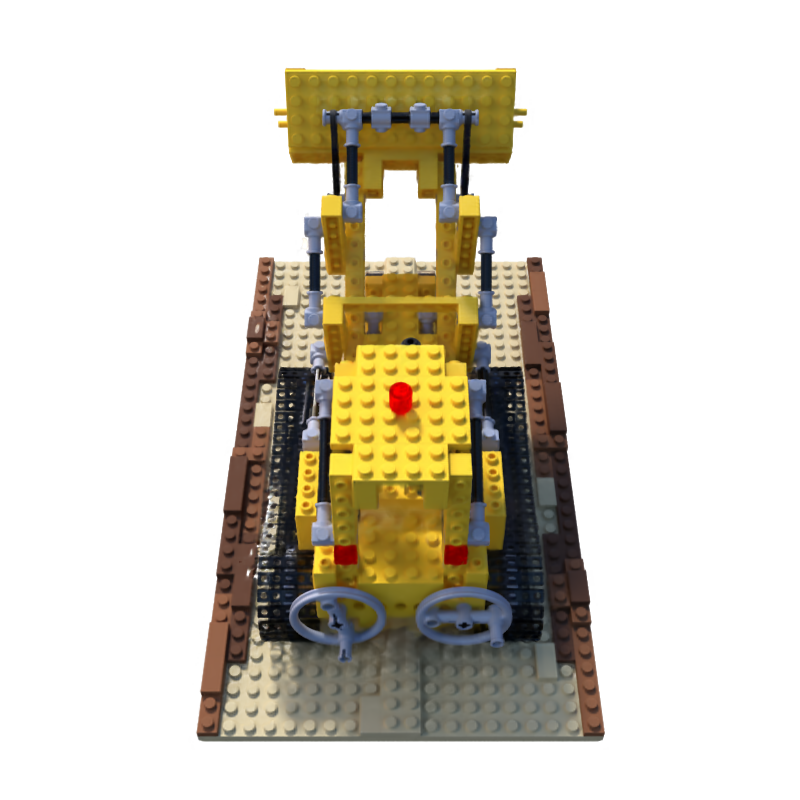}
  \plotcrop{figures/synthetic/ablations/lego25hitv.png}
  \caption{25 Images, High TV}
\end{subfigure}
\begin{subfigure}[b]{\awidth}
\plotbig{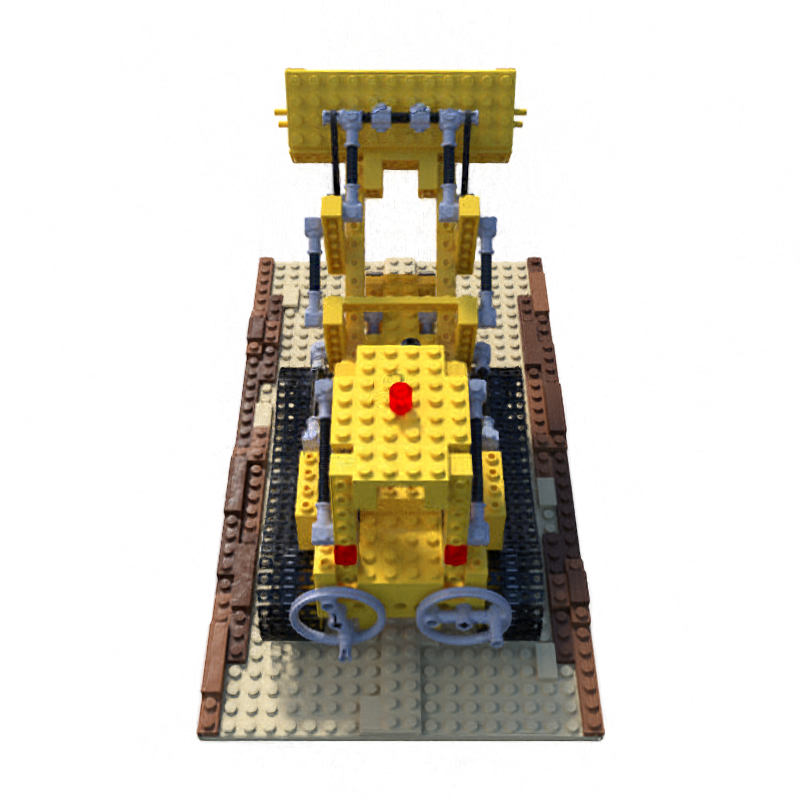}
  \plotcrop{figures/synthetic/ablations/legonv.png}
  \caption{NV Formula}
\end{subfigure}
  \caption{\textbf{Visual results of ablation studies} on the synthetic lego scene. Trilinear interpolation at resolution $256^3$ is quite similar to our full model at resolution $512^3$. Nearest neighbor interpolation shows clear voxel artifacts. Trilinear interpolation at lower resolution appears less detailed. Reducing the number of training views produces visual artifacts that are mostly resolved by increasing the TV regularization. Optimizing and rendering with the Neural Volumes \cite{lombardi2019neural} formula produces different visual artifacts.}
  \label{fig:bigablate}
\end{figure*}

%% file: figures_tex/fullsynthetic.tex
\newcommand{\fullwidth}{0.24\linewidth}

\newcommand{\plotzoom}[1]{\adjincludegraphics[trim={{0} {0} {0} {0}}, clip, width=\linewidth]{#1}}

\begin{figure*}[t]
  \centering
  \begin{subfigure}[b]{\fullwidth}
    \plotzoom{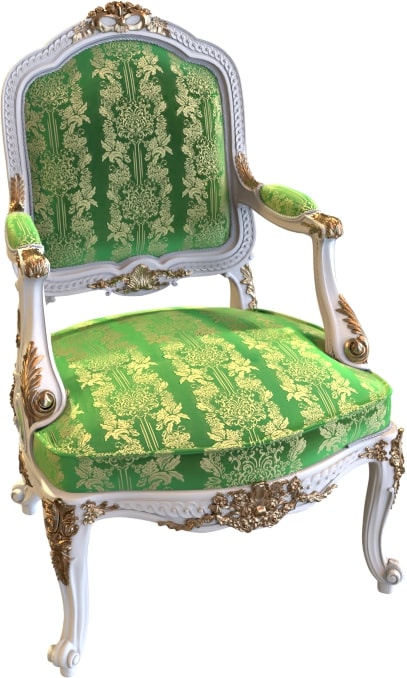}
  \plotzoom{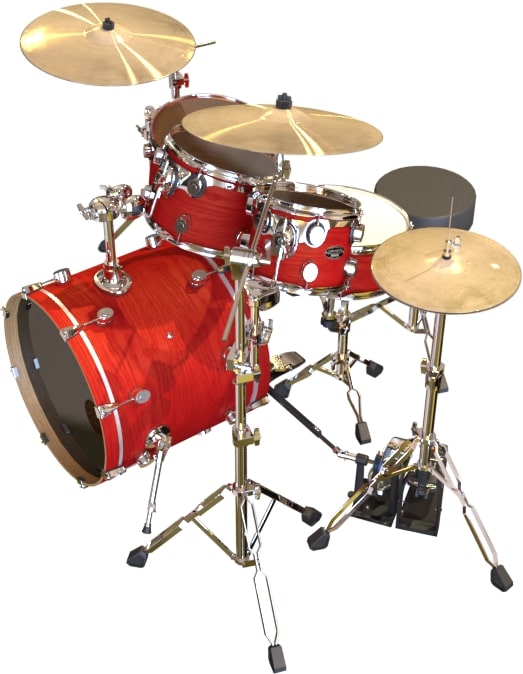}
  \plotzoom{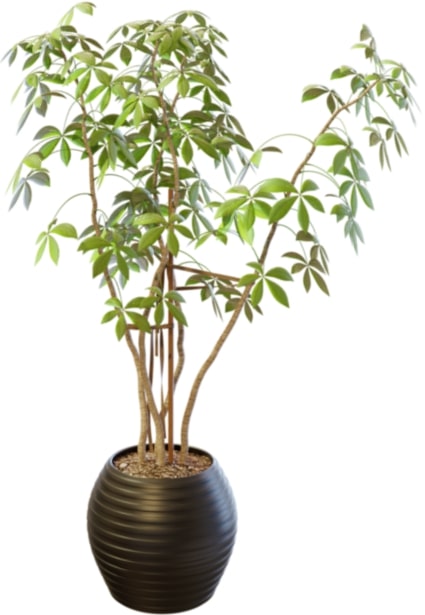}
  \plotzoom{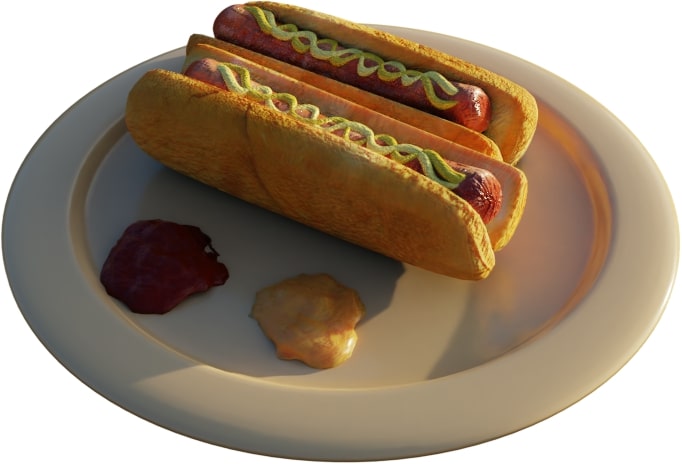}
  \caption{Ground Truth}
\end{subfigure}
  \begin{subfigure}[b]{\fullwidth}
    \plotzoom{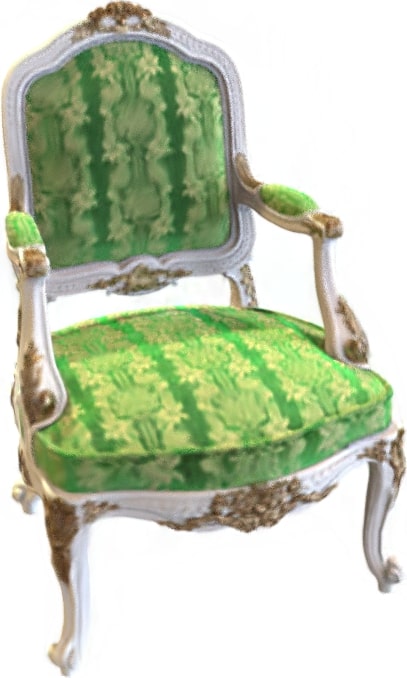}
   \plotzoom{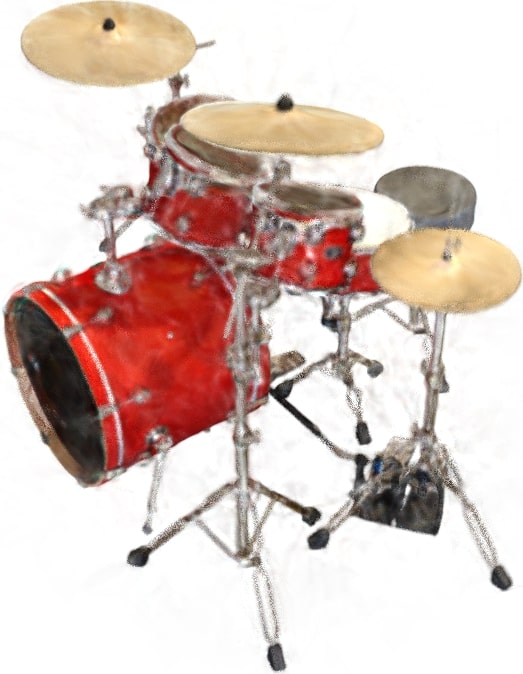}
   \plotzoom{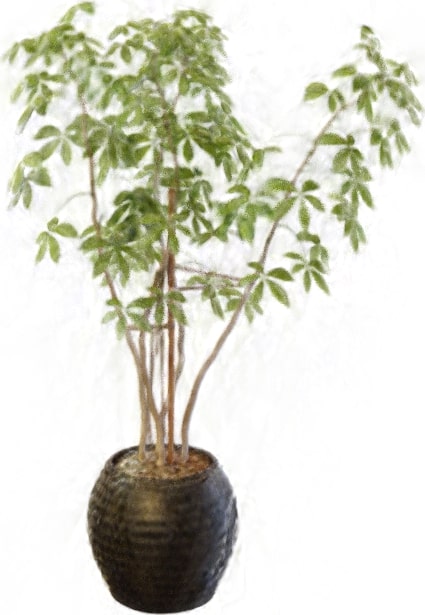}
   \plotzoom{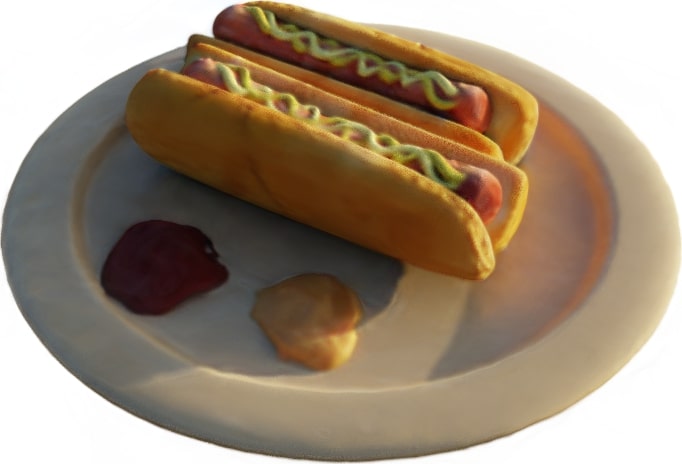}
  \caption{Neural Volumes}
\end{subfigure}
  \begin{subfigure}[b]{\fullwidth}
     \plotzoom{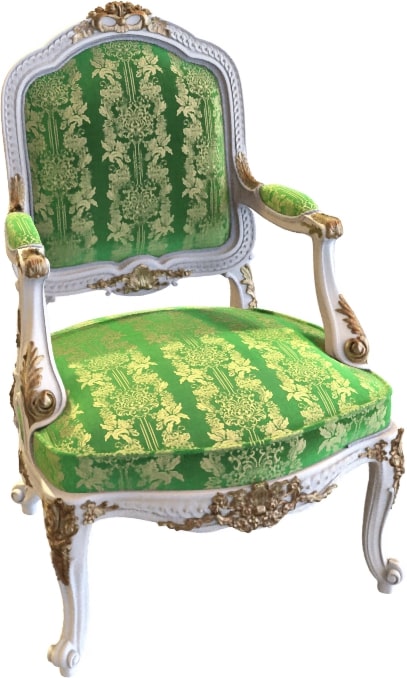}
   \plotzoom{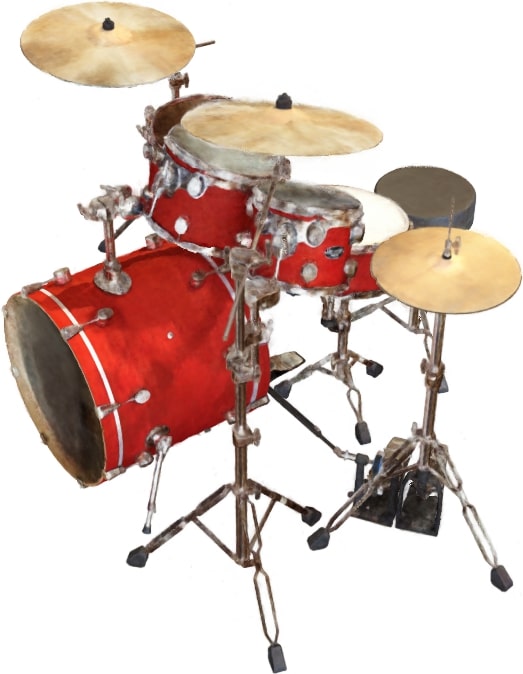}
   \plotzoom{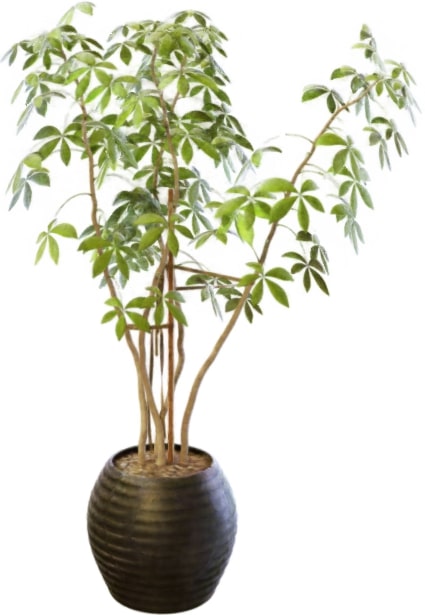}
   \plotzoom{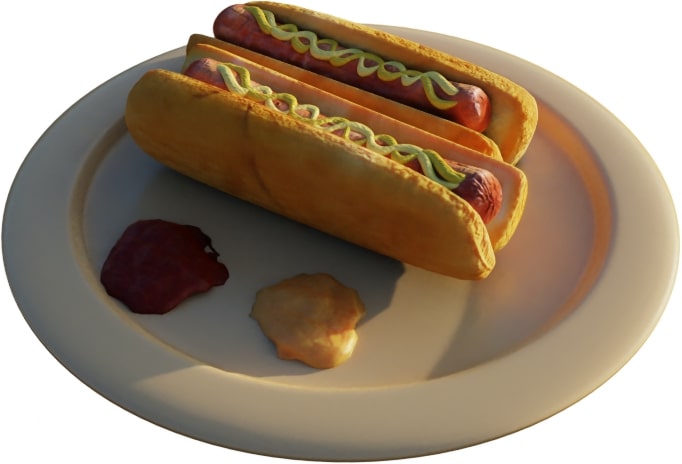}
  \caption{JAXNeRF}
\end{subfigure}
  \begin{subfigure}[b]{\fullwidth}
    \plotzoom{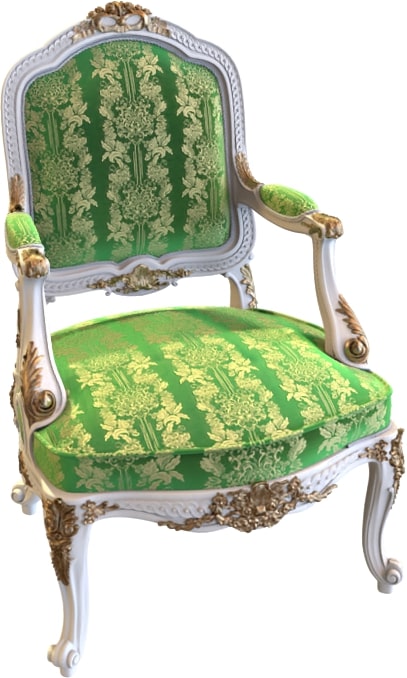}
  \plotzoom{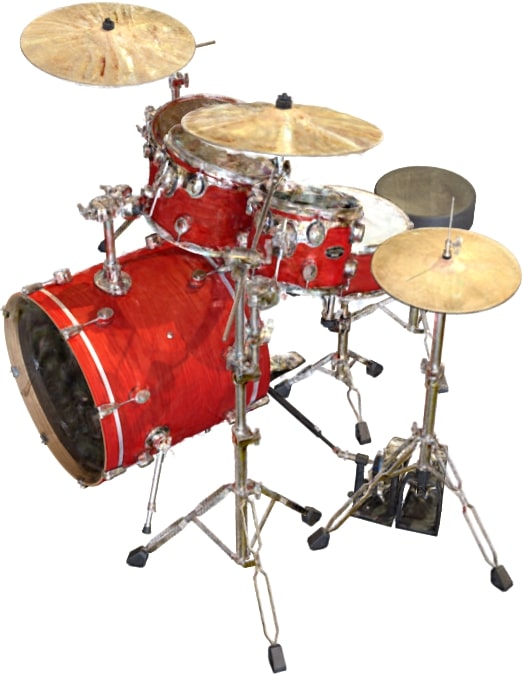}
  \plotzoom{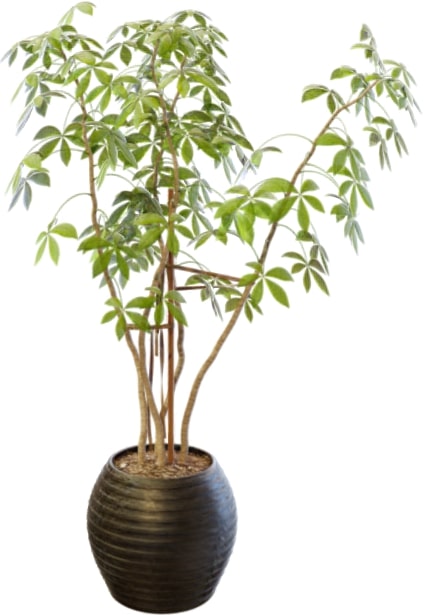}
  \plotzoom{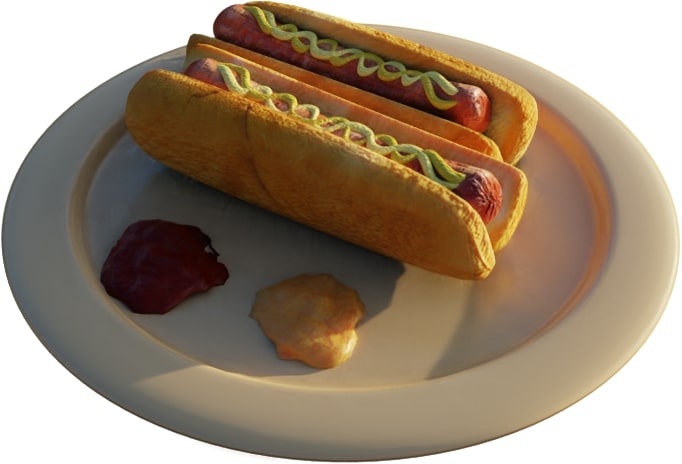}
  \caption{Plenoxels}
\end{subfigure}
\caption{\textbf{Synthetic scenes.} We show a random view from each of the synthetic scenes, comparing the ground truth, Neural Volumes~\cite{lombardi2019neural}, JAXNeRF~\cite{mildenhall2020nerf, jaxnerf2020github}, and our Plenoxels.}
\end{figure*}
\begin{figure*}
\ContinuedFloat
\centering
  \begin{subfigure}[b]{\fullwidth}
    \plotzoom{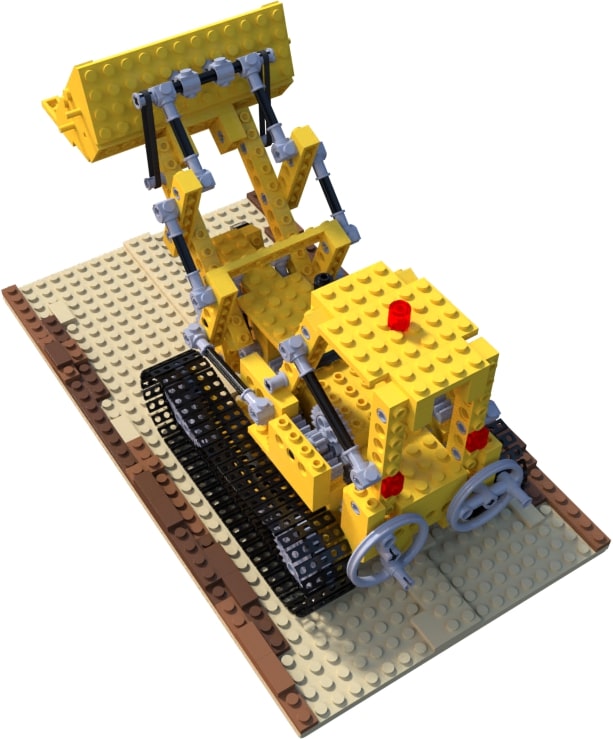}
  \plotzoom{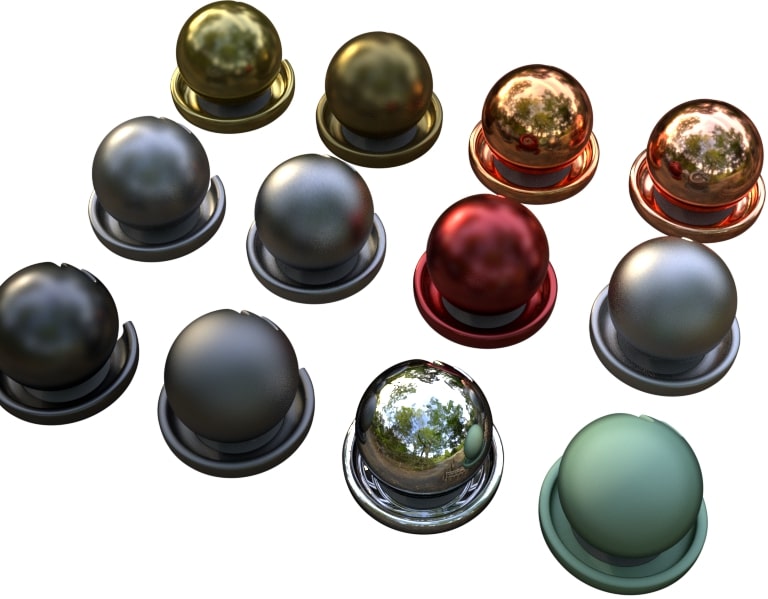}
  \plotzoom{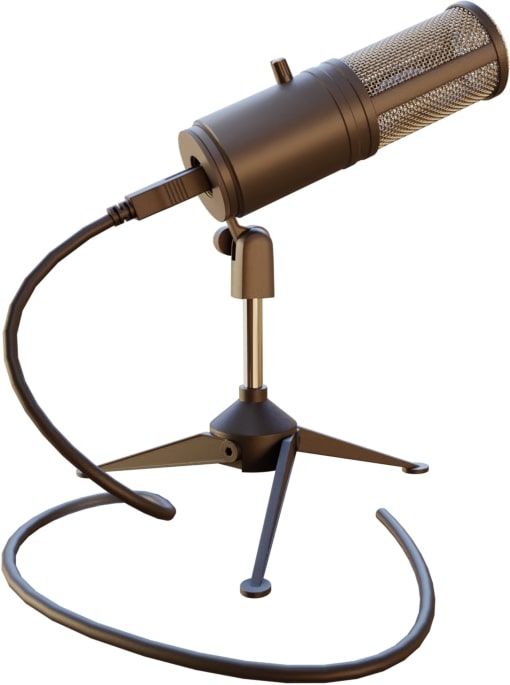}
  \plotzoom{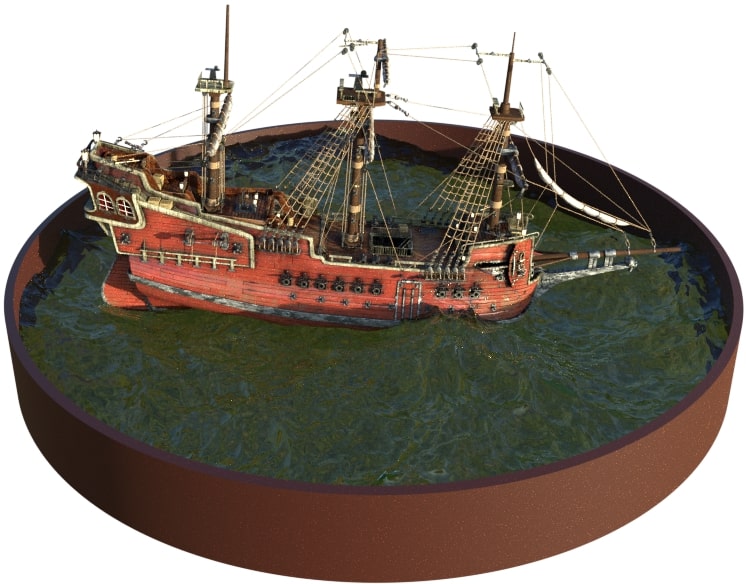}
  \caption{Ground Truth}
\end{subfigure}
  \begin{subfigure}[b]{\fullwidth}
    \plotzoom{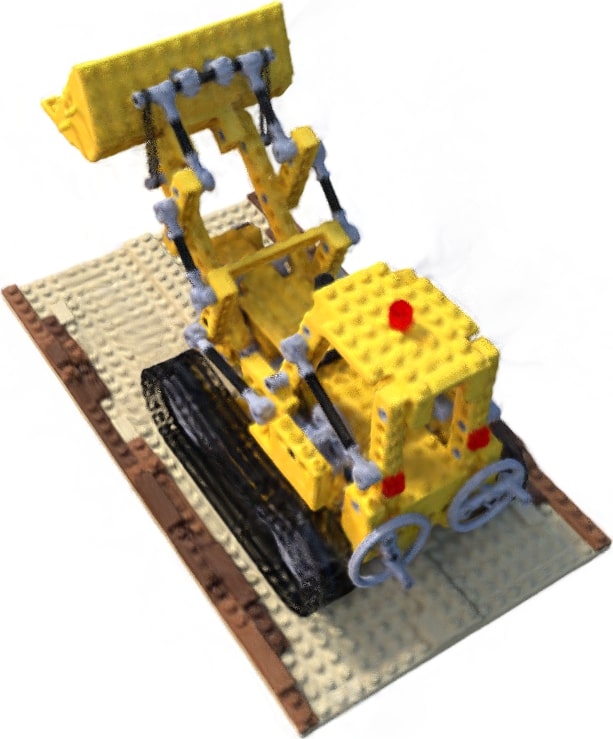}
  \plotzoom{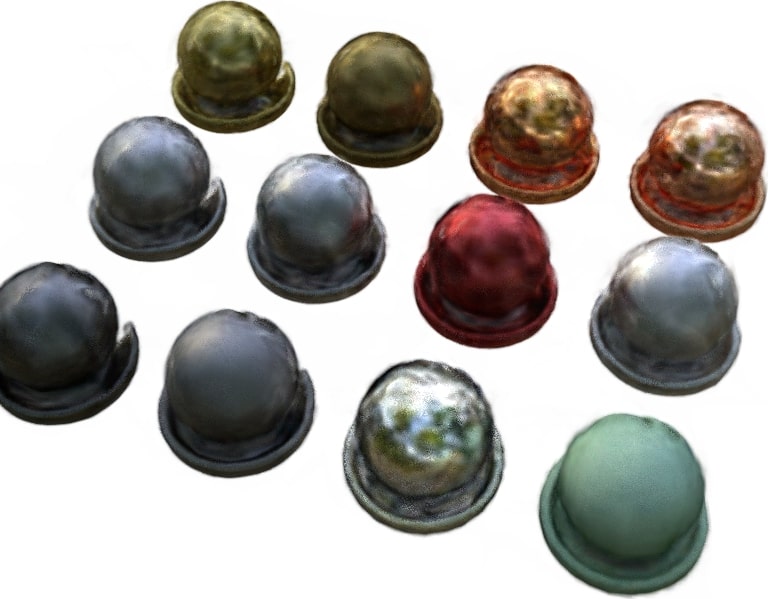}
  \plotzoom{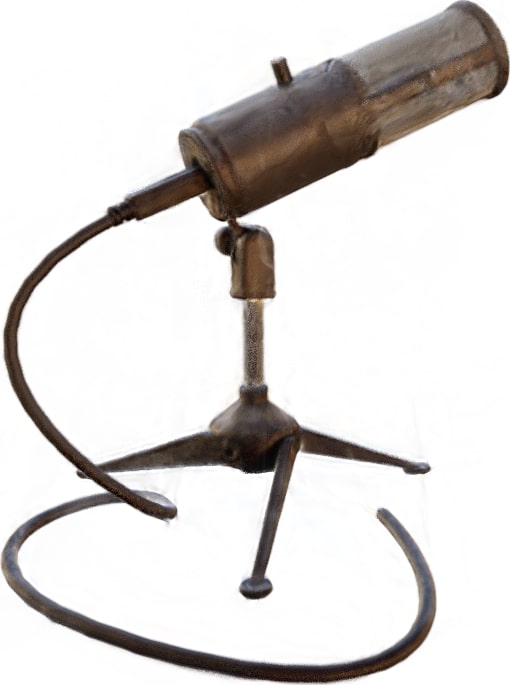}
  \plotzoom{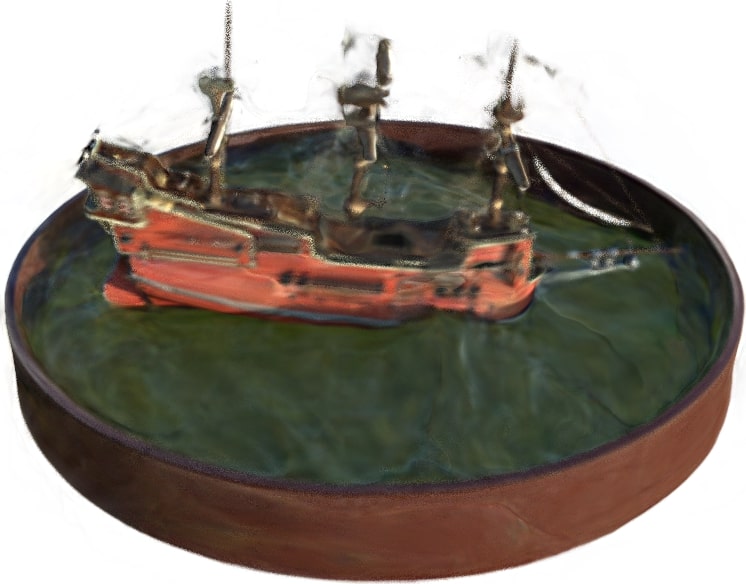}
  \caption{Neural Volumes}
\end{subfigure}
  \begin{subfigure}[b]{\fullwidth}
    \plotzoom{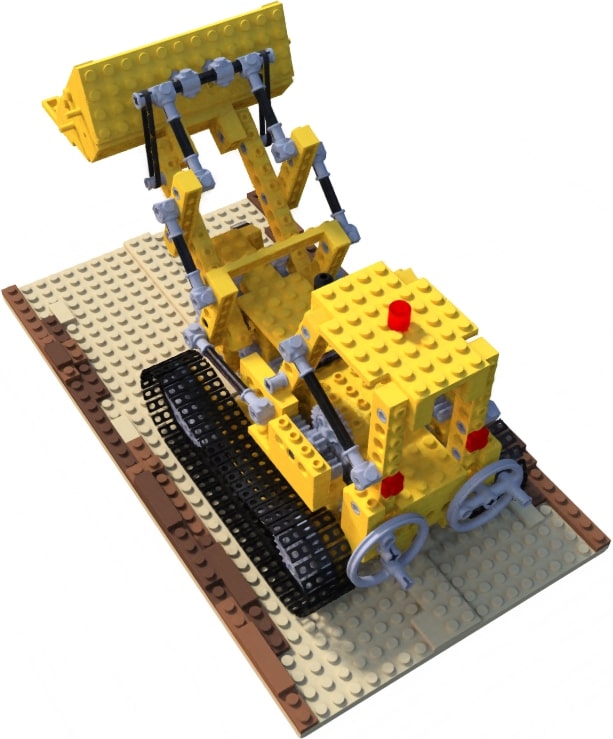}
  \plotzoom{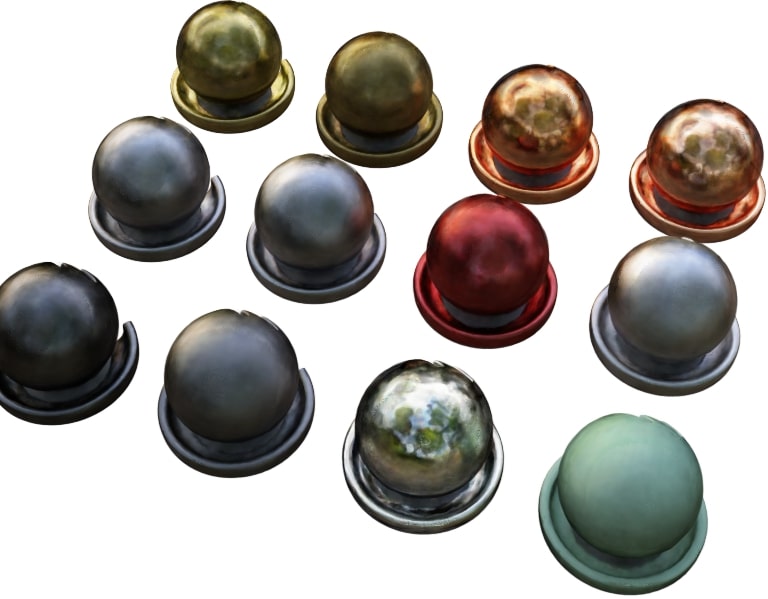}
  \plotzoom{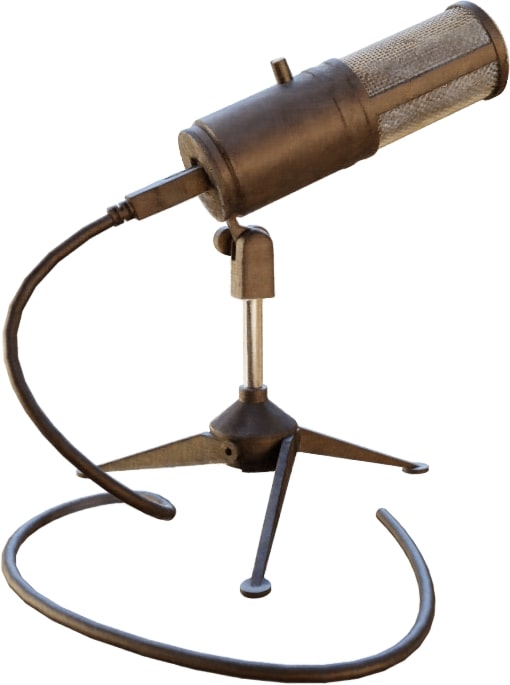}
  \plotzoom{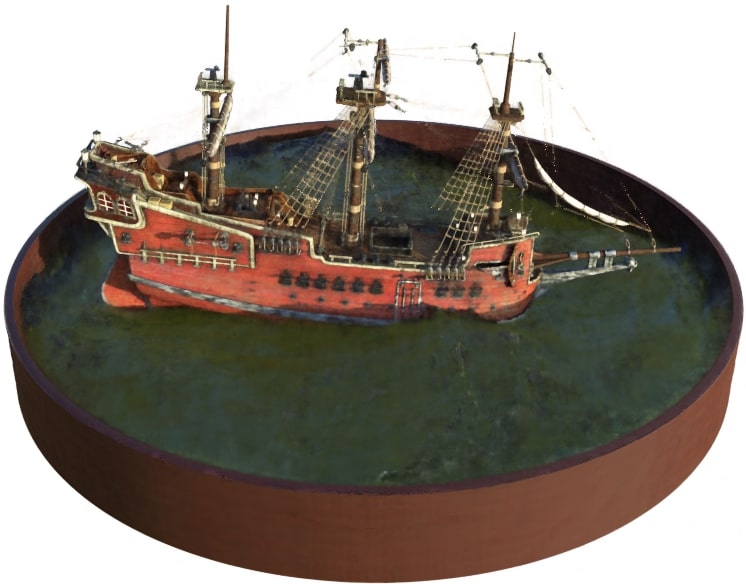}
  \caption{JAXNeRF}
\end{subfigure}
  \begin{subfigure}[b]{\fullwidth}
    \plotzoom{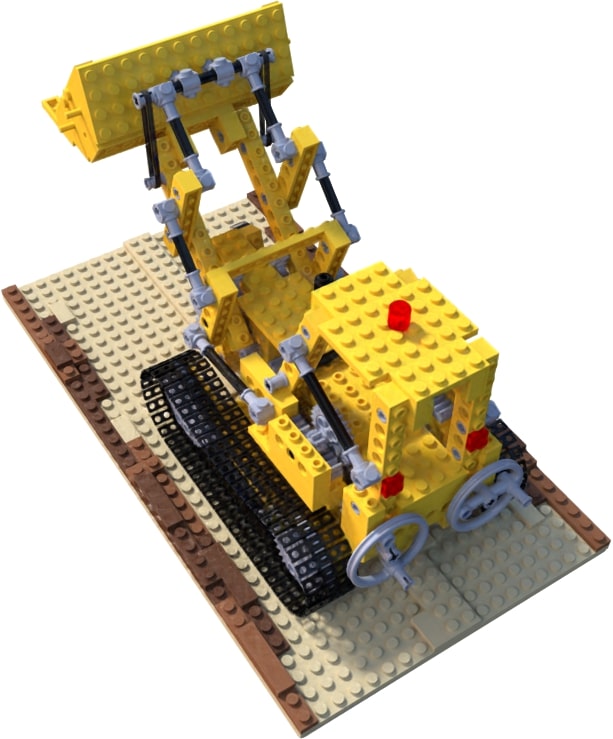}
  \plotzoom{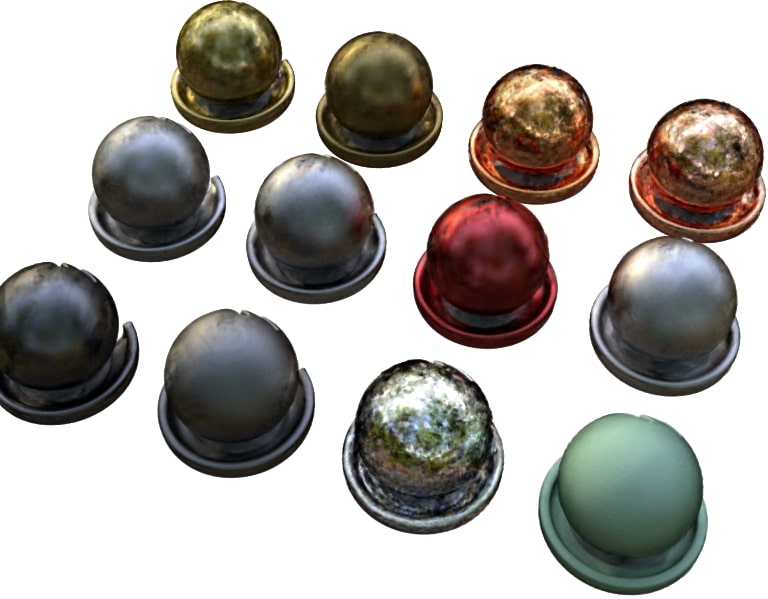}
  \plotzoom{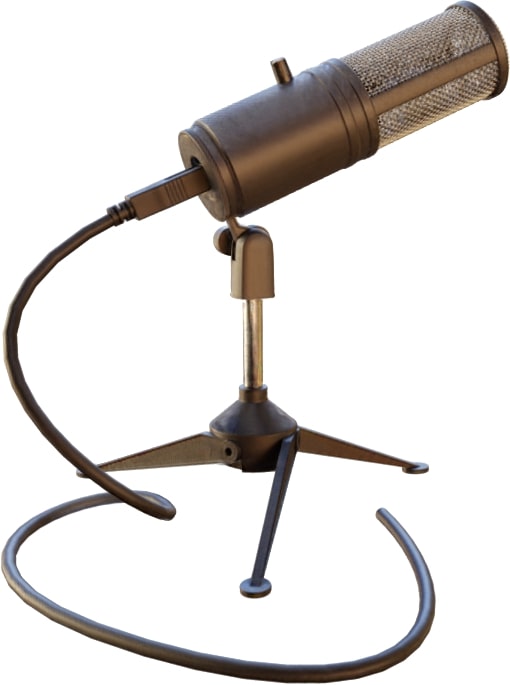}
  \plotzoom{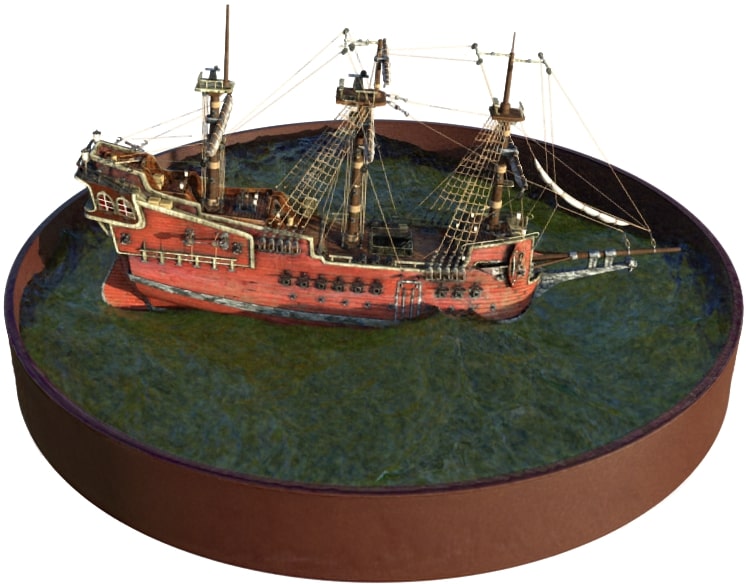}
  \caption{Plenoxels}
\end{subfigure}
  \caption{\textbf{Synthetic scenes.} We show a random view from each of the synthetic scenes, comparing the ground truth, Neural Volumes~\cite{lombardi2019neural}, JAXNeRF~\cite{mildenhall2020nerf, jaxnerf2020github}, and our Plenoxels.}
  \label{fig:fullsynthetic}
\end{figure*}

%% file: figures_tex/fullforward.tex
\newcommand{\thirdwidth}{0.32\linewidth}

\begin{figure*}[t]
  \centering
  \begin{subfigure}[b]{\thirdwidth}
    \plotzoom{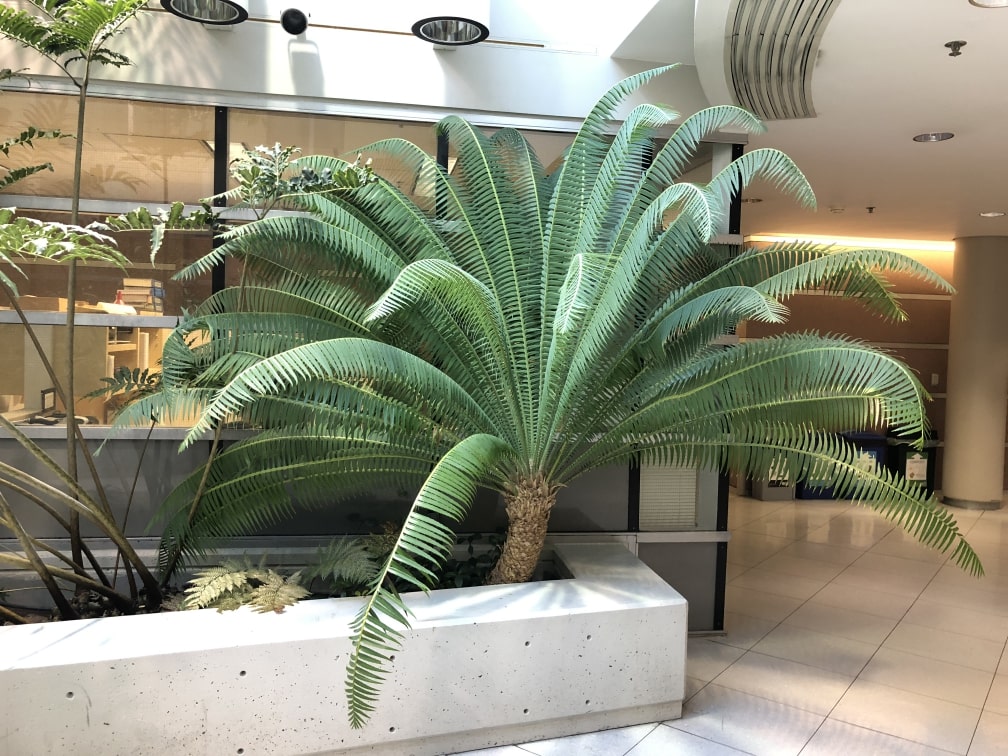}
  \plotzoom{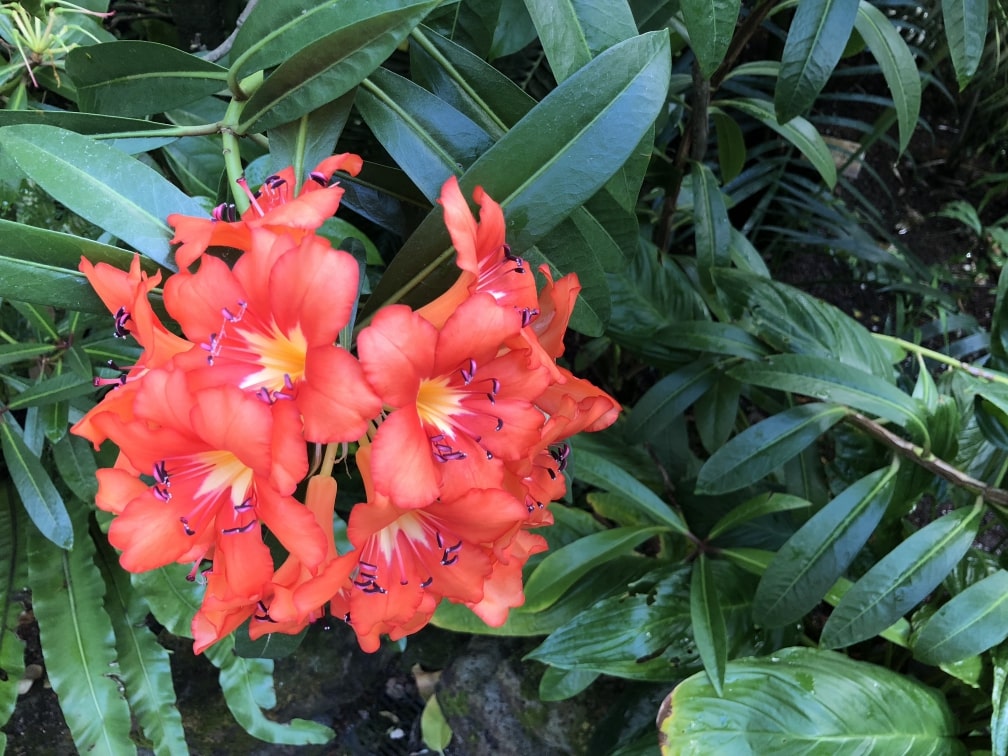}
  \plotzoom{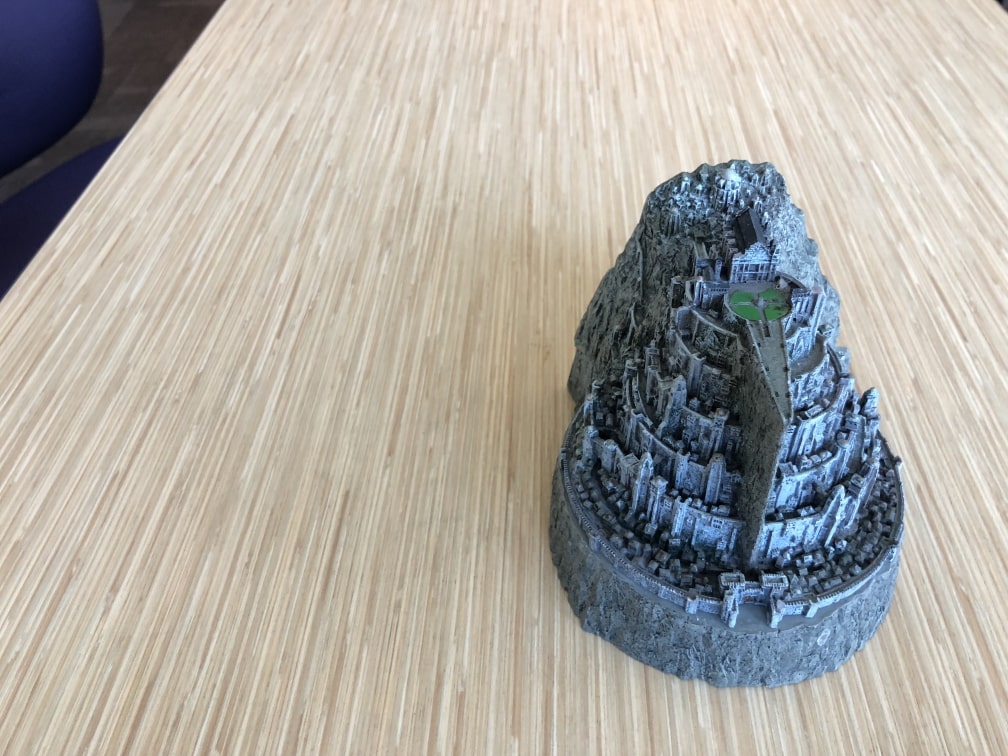}
  \plotzoom{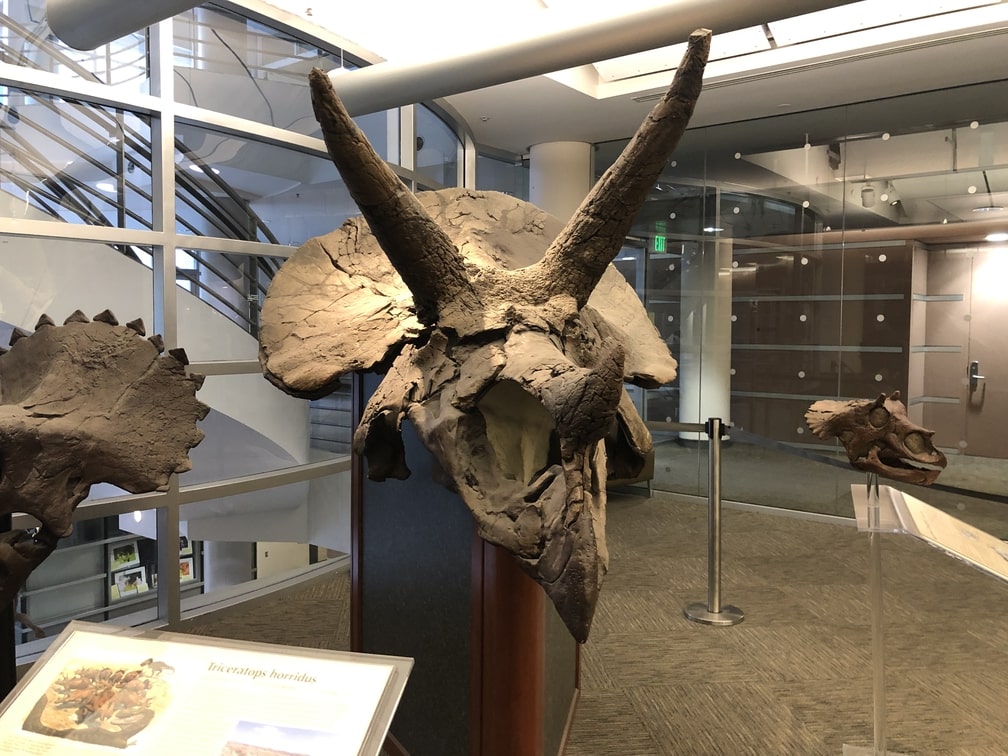}
  \caption{Ground Truth}
\end{subfigure}
  \begin{subfigure}[b]{\thirdwidth}
     \plotzoom{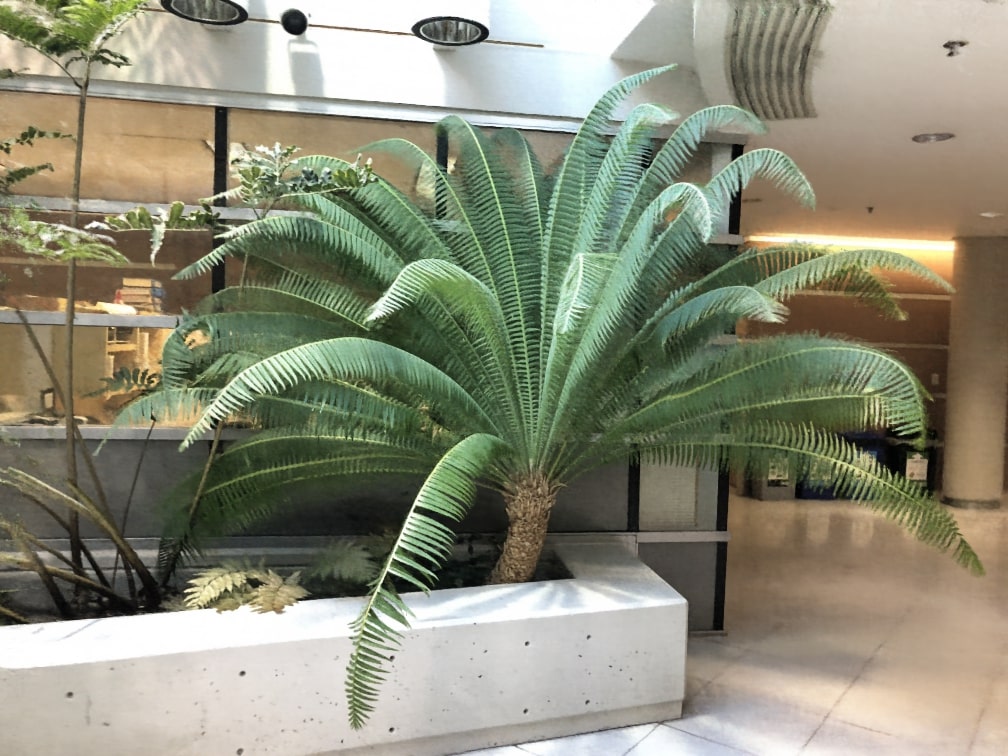}
   \plotzoom{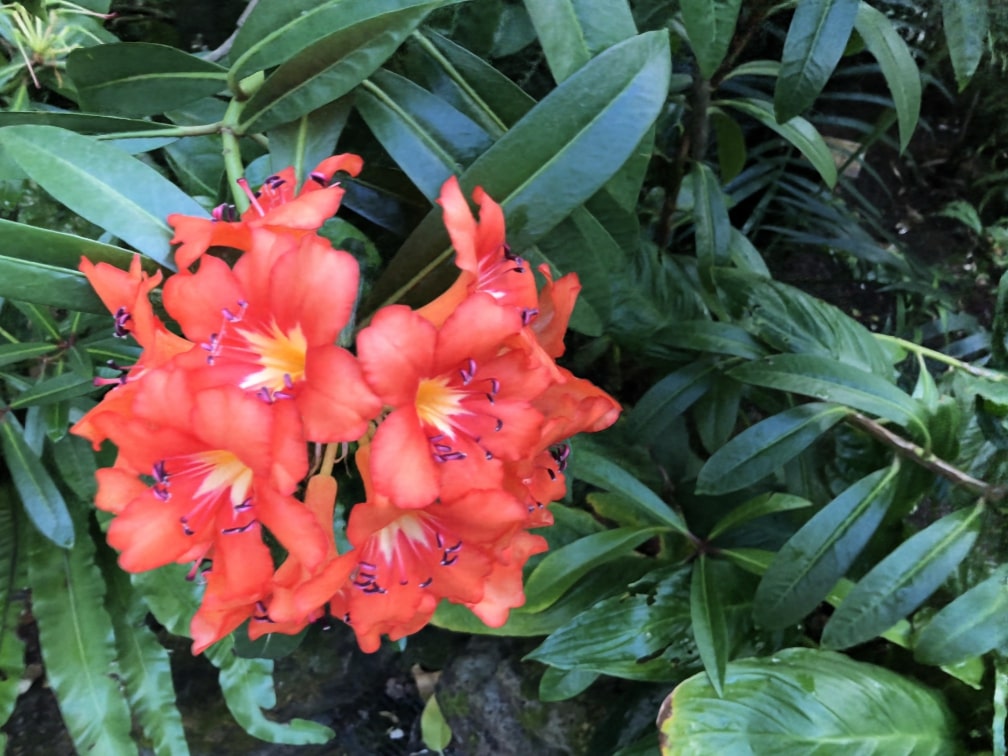}
   \plotzoom{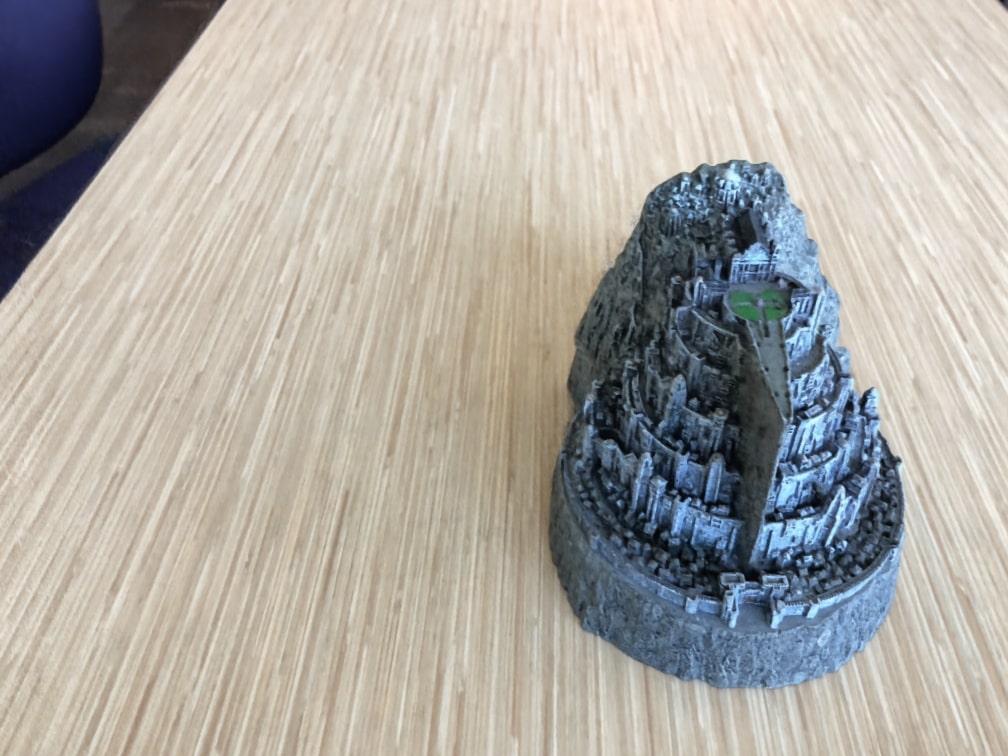}
   \plotzoom{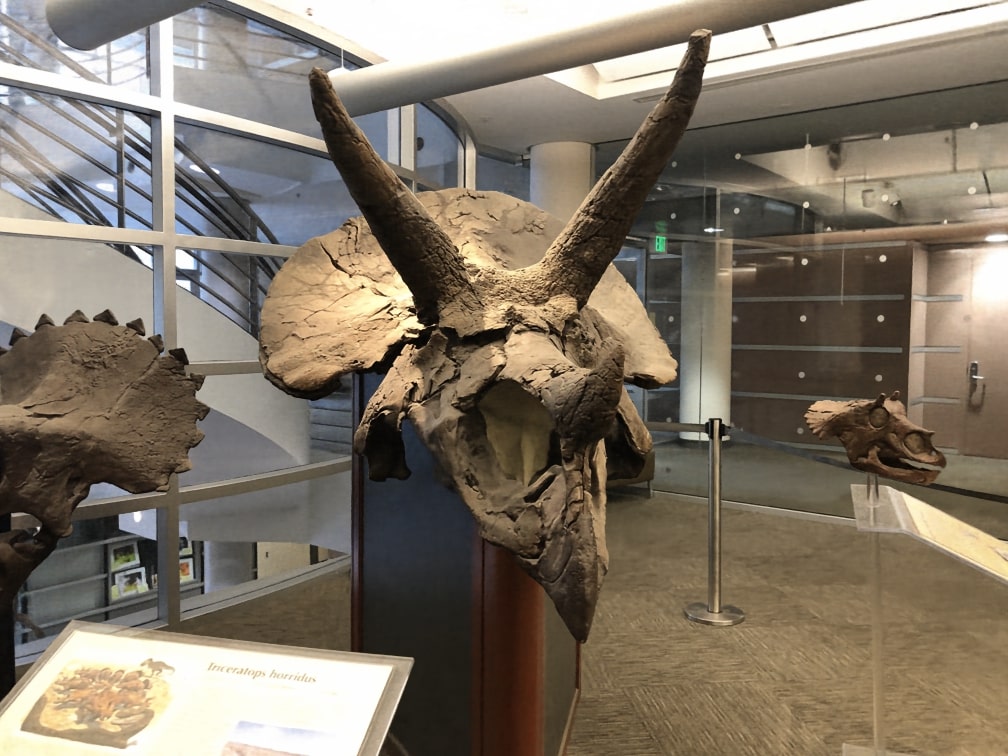}
  \caption{JAXNeRF}
\end{subfigure}
  \begin{subfigure}[b]{\thirdwidth}
    \plotzoom{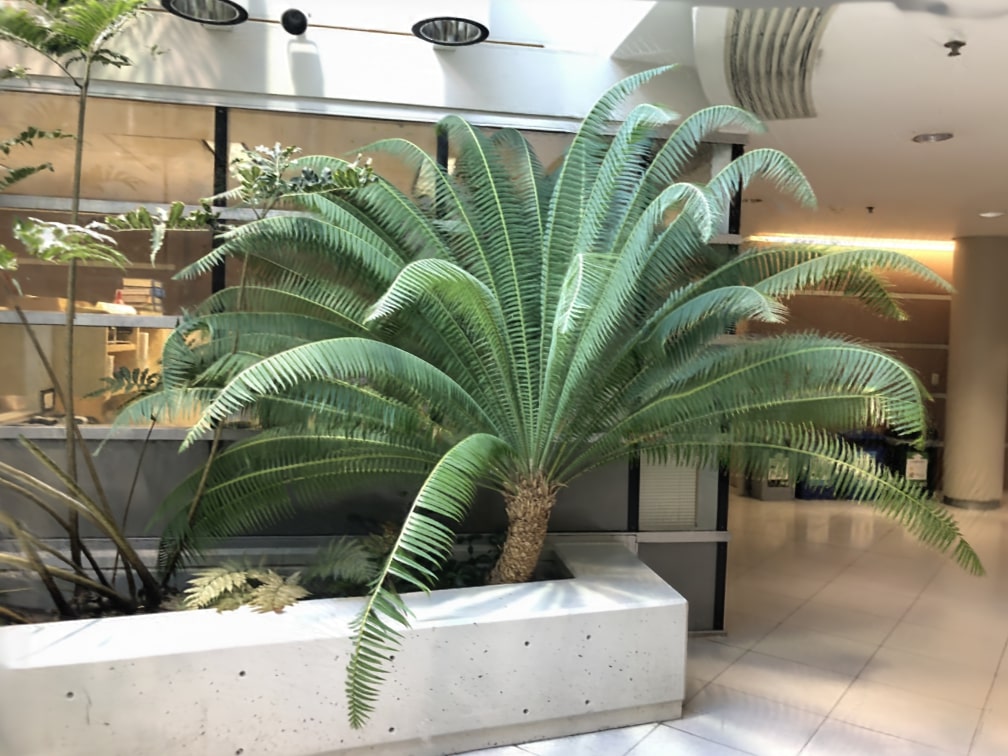}
  \plotzoom{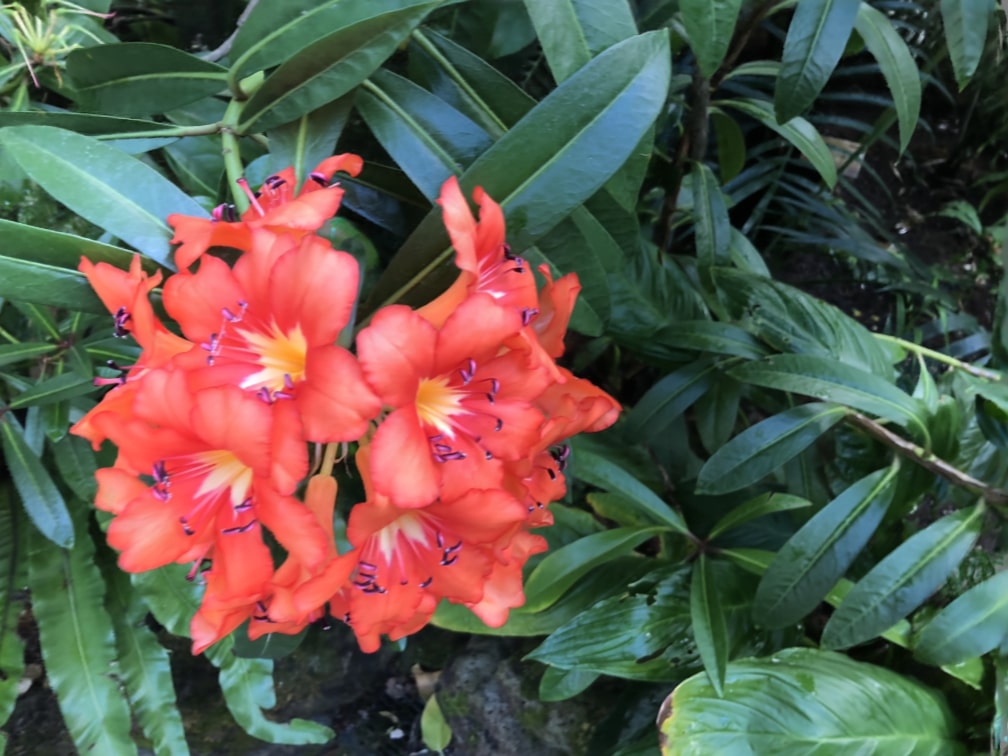}
  \plotzoom{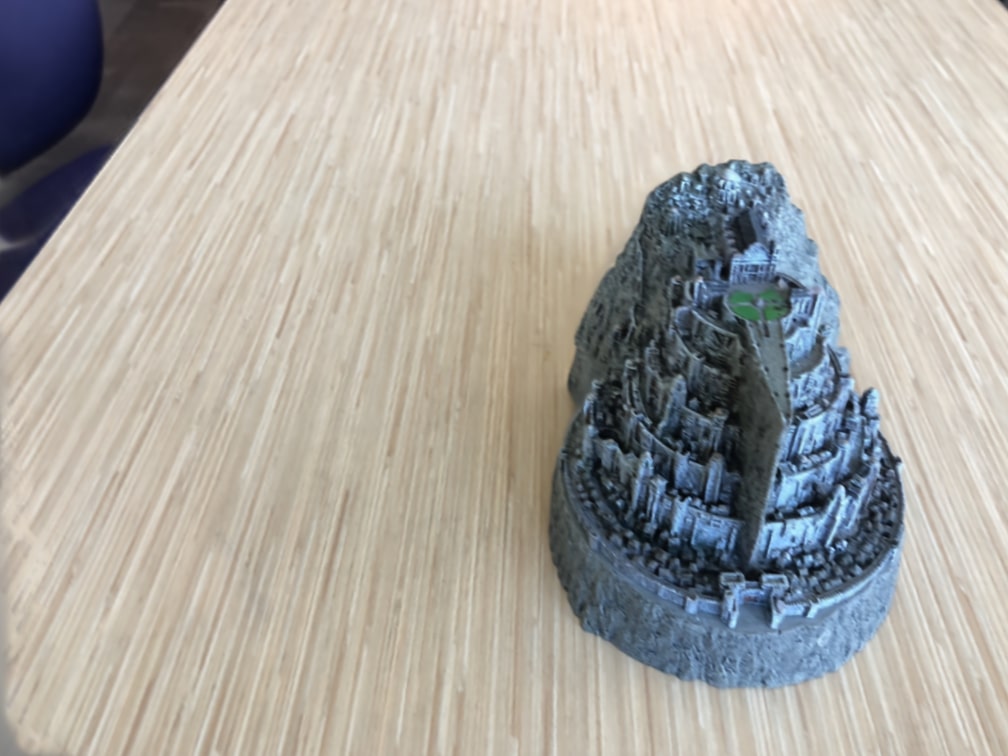}
  \plotzoom{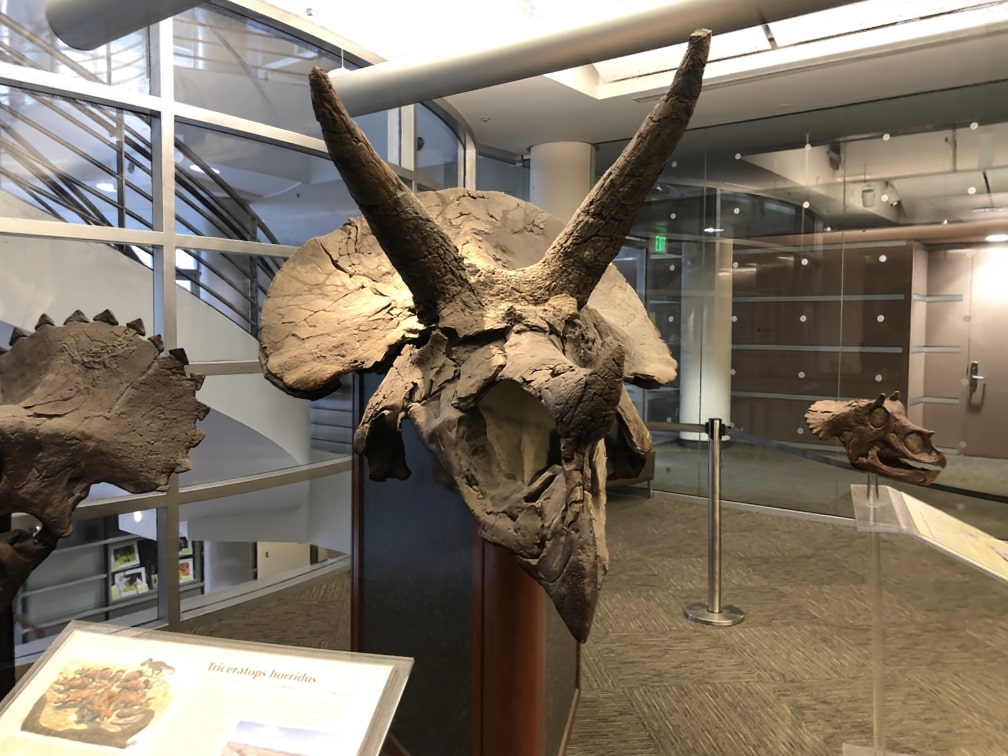}
  \caption{Plenoxels}
\end{subfigure}
\caption{\textbf{Forward-facing scenes.} We show a random view from each of the forward-facing scenes, comparing the ground truth, JAXNeRF~\cite{mildenhall2020nerf, jaxnerf2020github}, and our Plenoxels.}
\end{figure*}
\begin{figure*}
\ContinuedFloat
\centering
  \begin{subfigure}[b]{\thirdwidth}
    \plotzoom{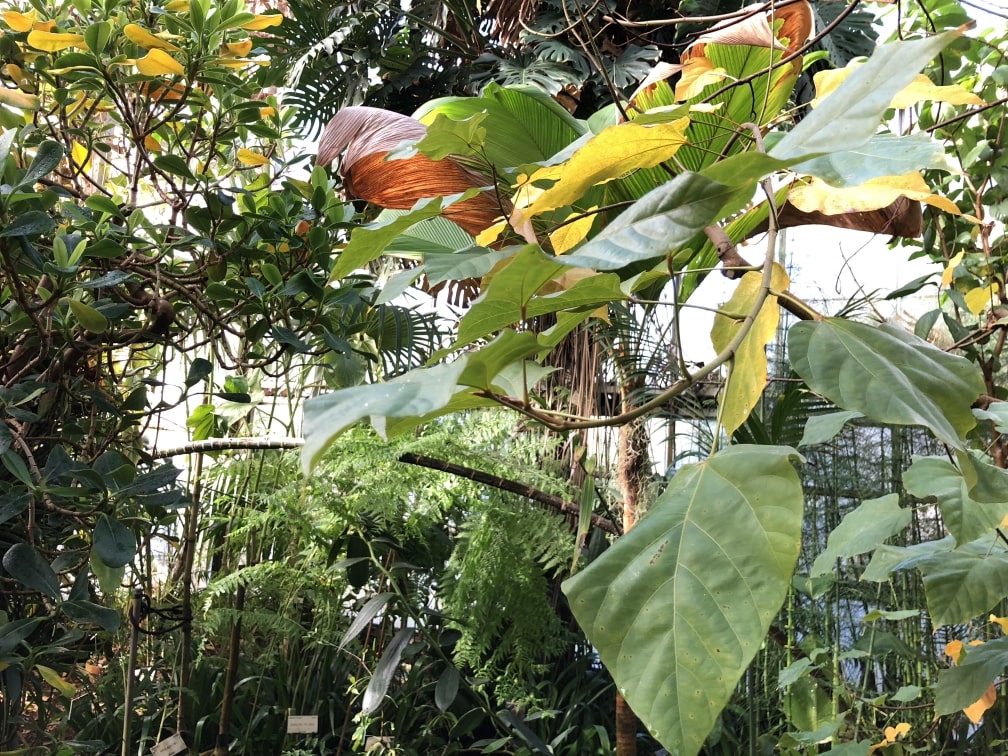}
  \plotzoom{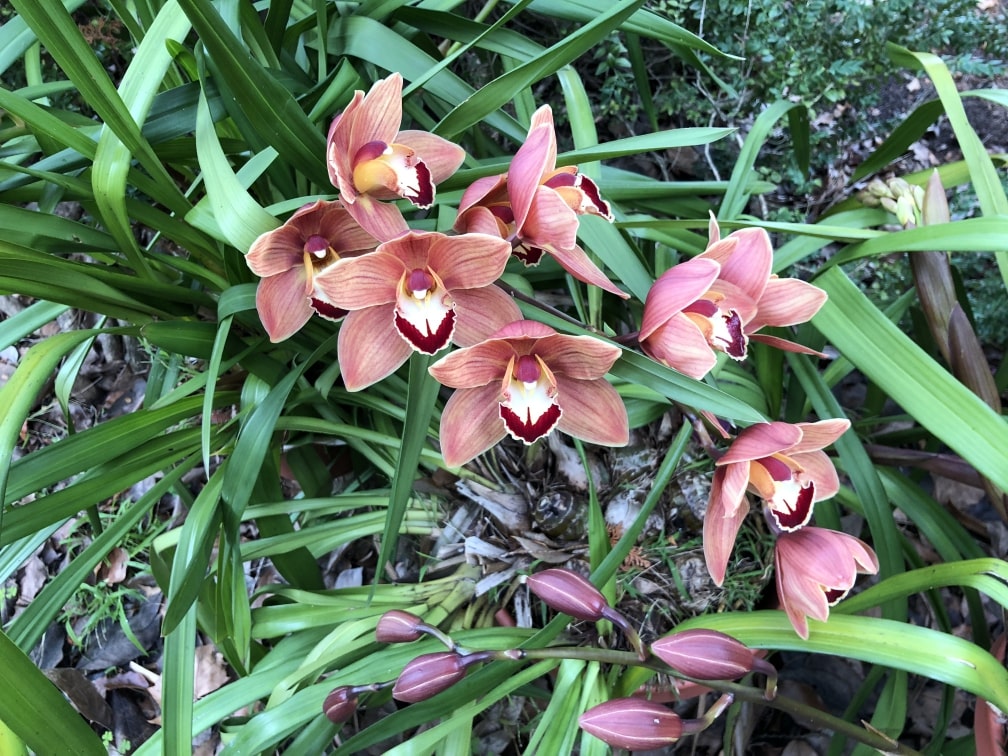}
  \plotzoom{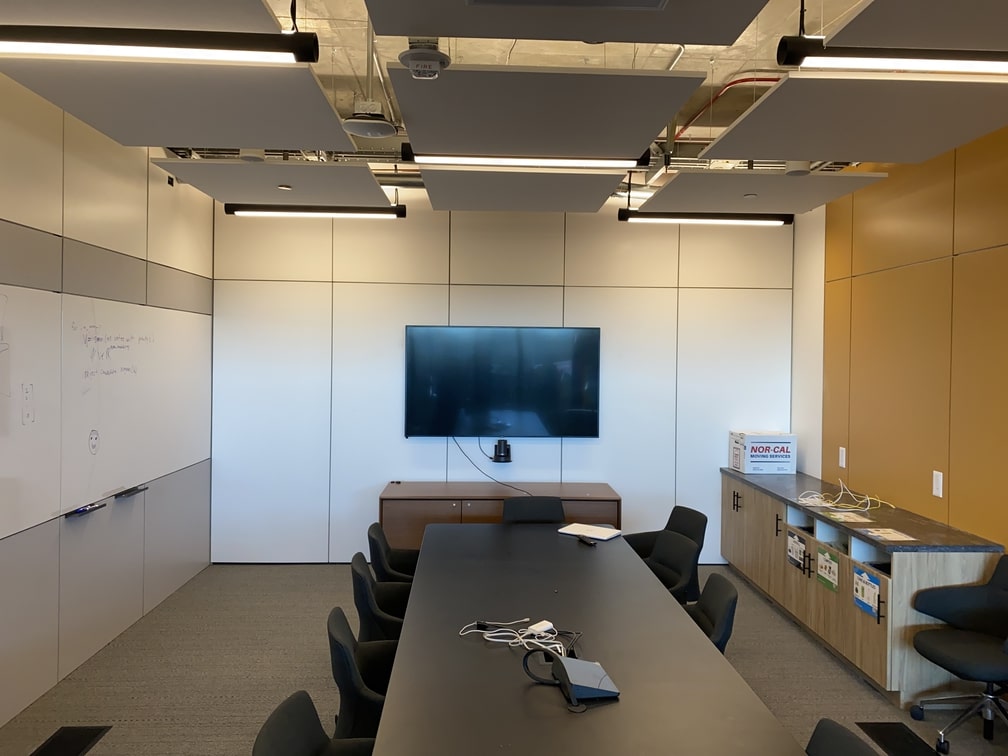}
  \plotzoom{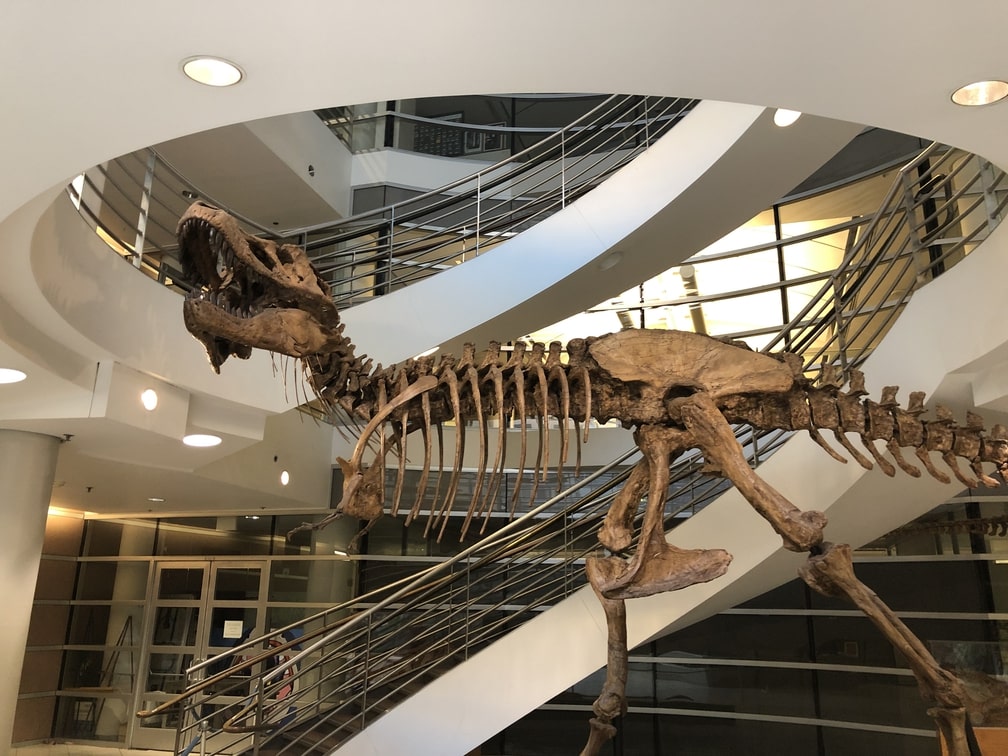}
  \caption{Ground Truth}
\end{subfigure}
  \begin{subfigure}[b]{\thirdwidth}
    \plotzoom{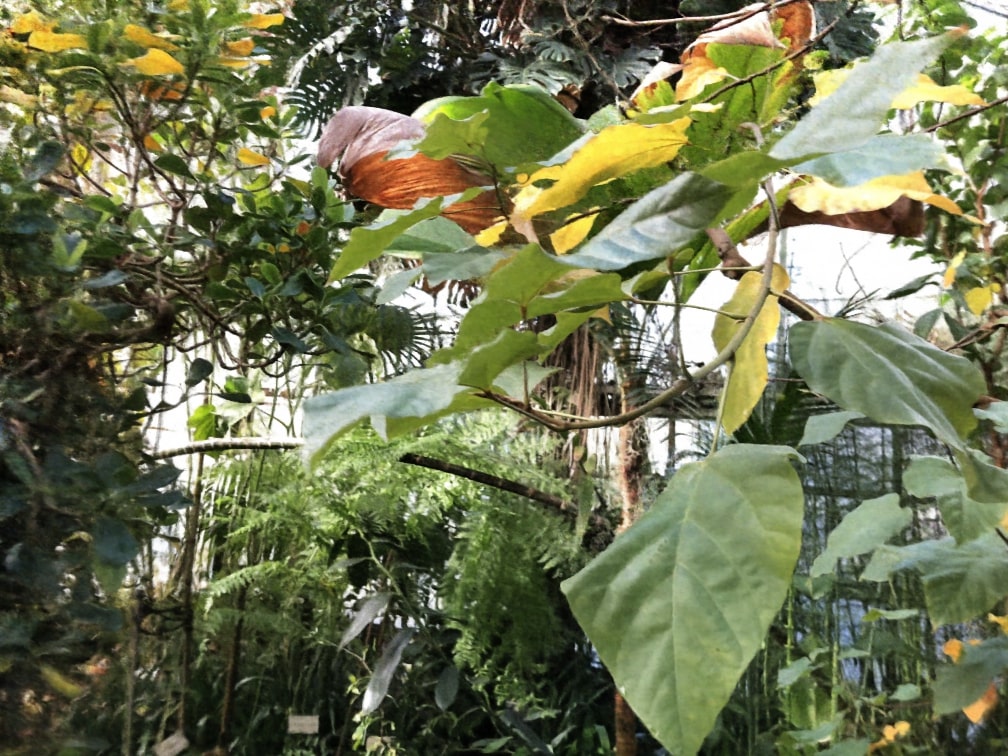}
  \plotzoom{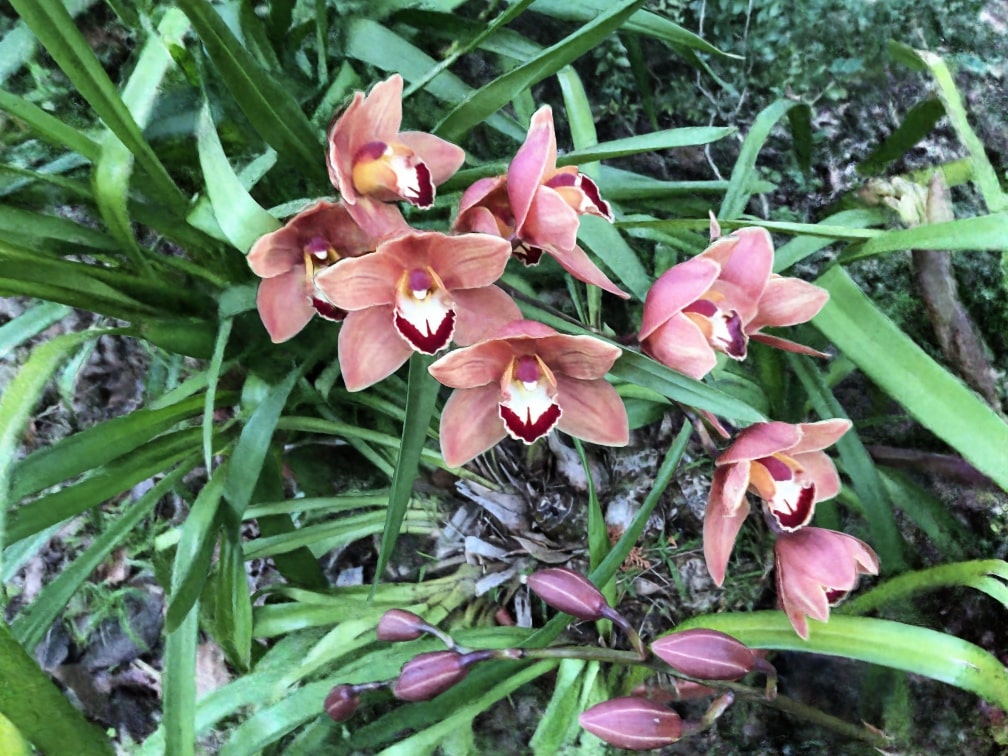}
  \plotzoom{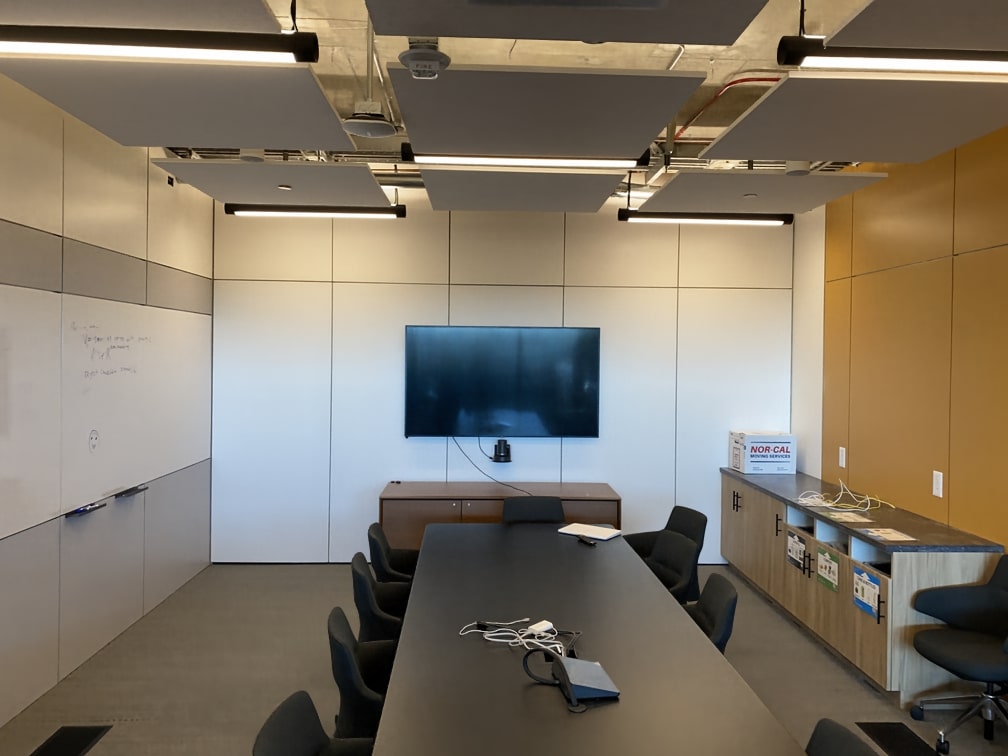}
  \plotzoom{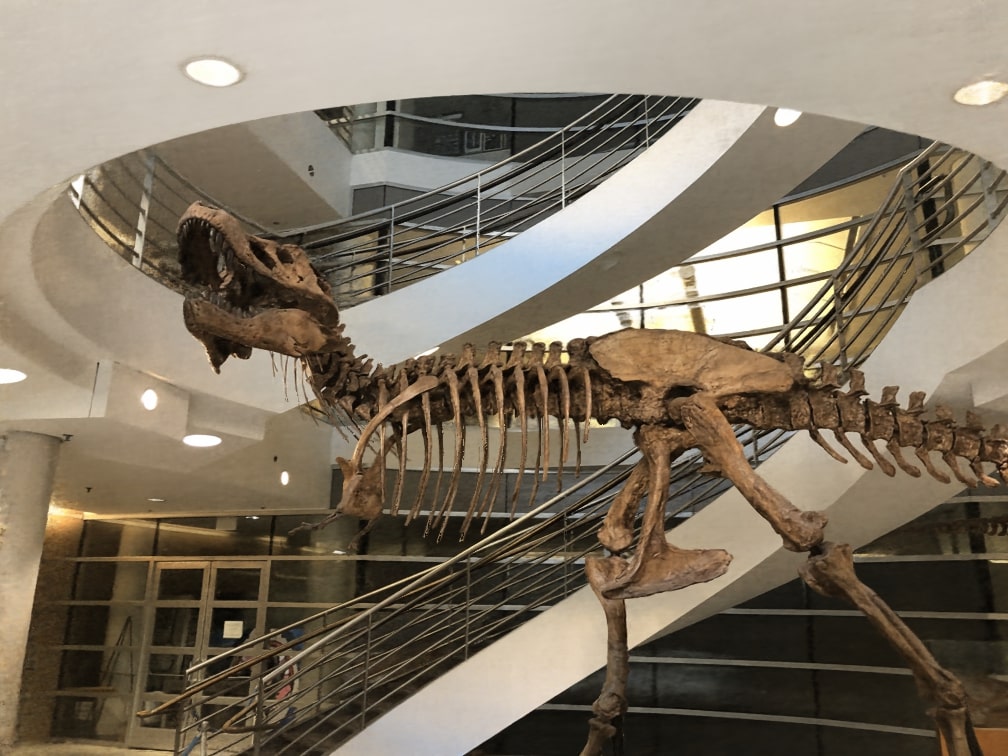}
  \caption{JAXNeRF}
\end{subfigure}
  \begin{subfigure}[b]{\thirdwidth}
    \plotzoom{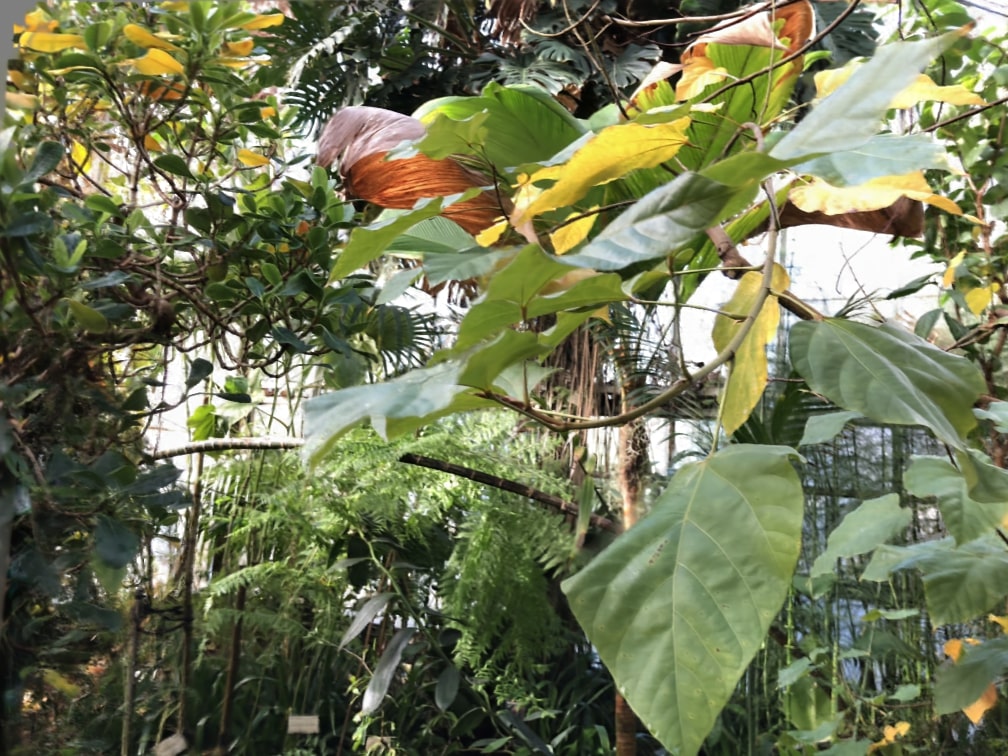}
  \plotzoom{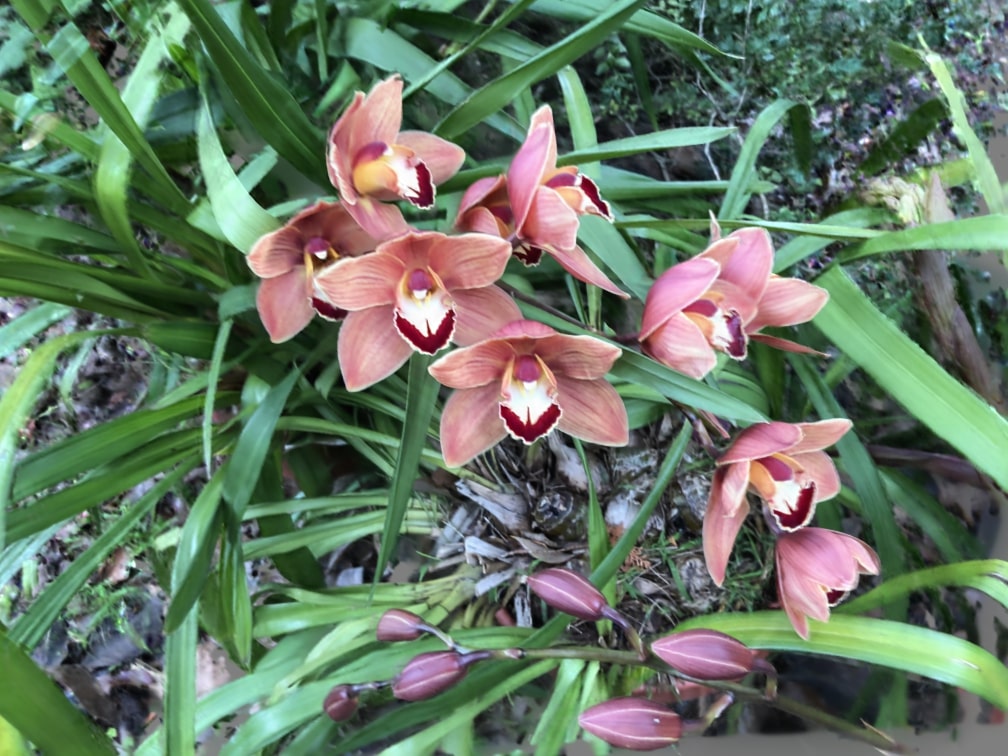}
  \plotzoom{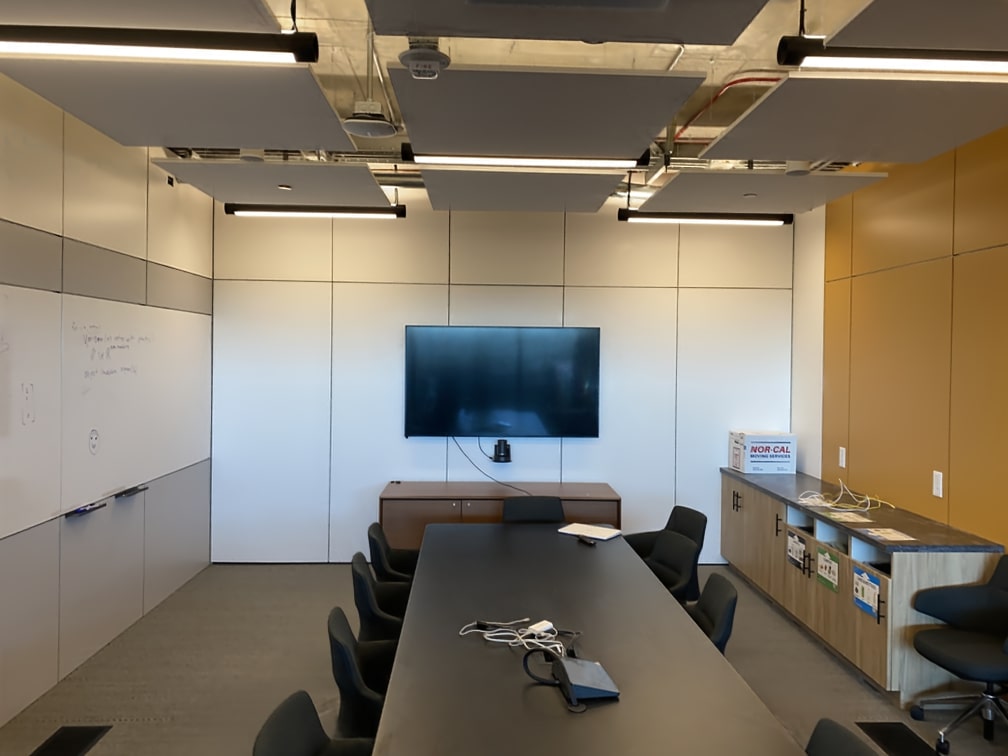}
  \plotzoom{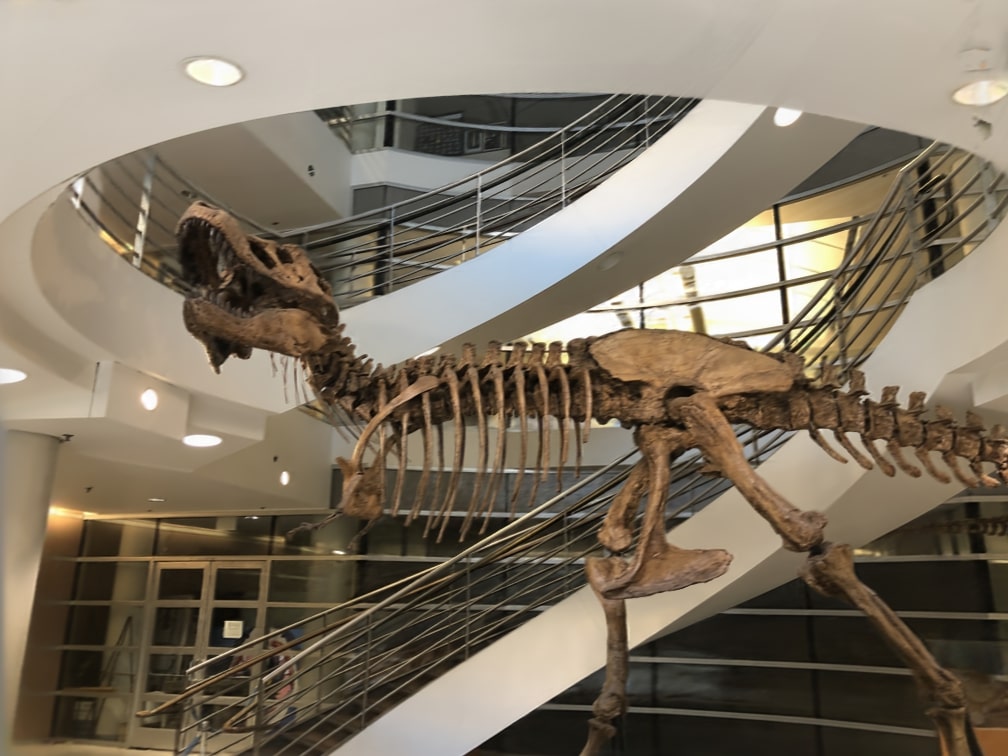}
  \caption{Plenoxels}
\end{subfigure}
  \caption{\textbf{Forward-facing scenes.} We show a random view from each of the forward-facing scenes, comparing the ground truth, JAXNeRF~\cite{mildenhall2020nerf, jaxnerf2020github}, and our Plenoxels. Note that these two methods have different behaviors in unsupervised regions (\eg the bottom right in the orchids view): JAXNeRF fills in plausible textures whereas Plenoxels default to gray.}
  \label{fig:fullforward}
\end{figure*}

%% file: figures_tex/full360.tex
\newcommand{\plothalf}[1]{\adjincludegraphics[trim={{0.5\width} {0} {0} {0}}, clip, width=\linewidth]{#1}}

\begin{figure*}[t]
  \centering
  \begin{subfigure}[b]{\thirdwidth}
    \plotzoom{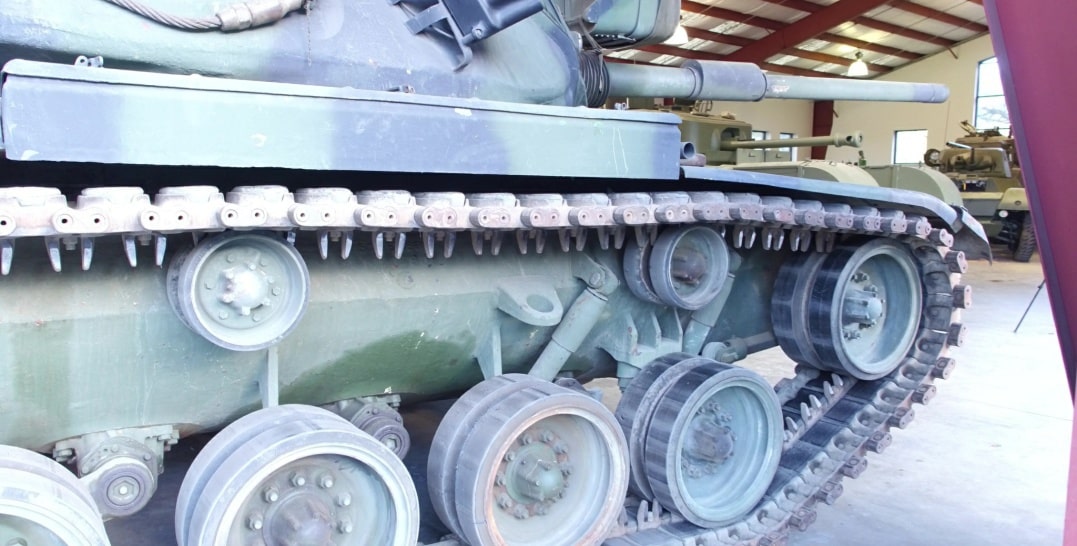}
    \plotzoom{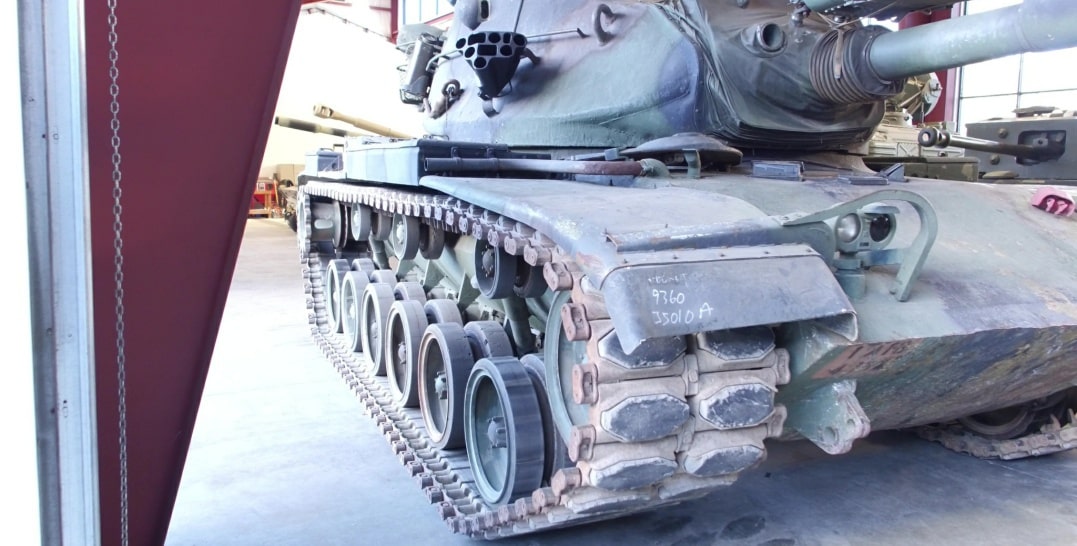}
  \plotzoom{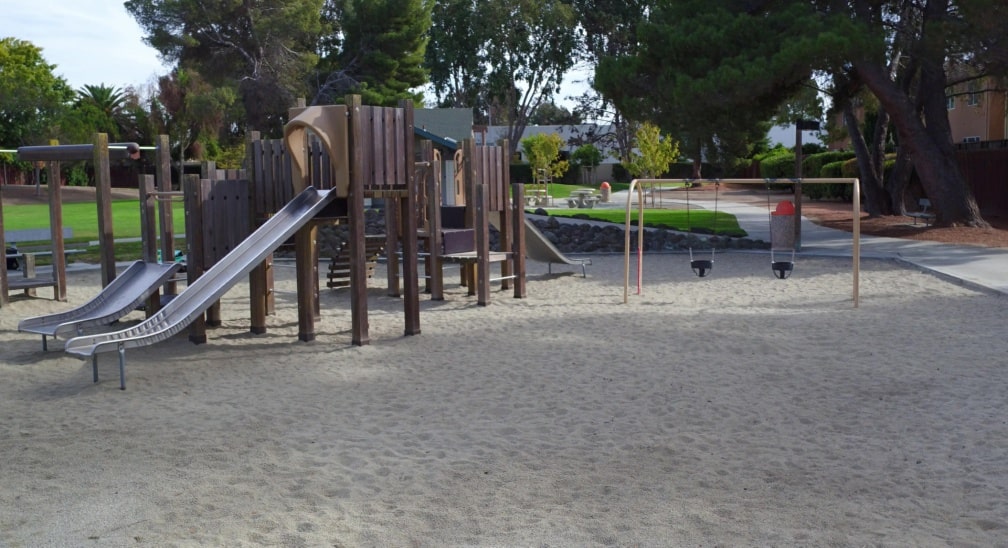}
  \plotzoom{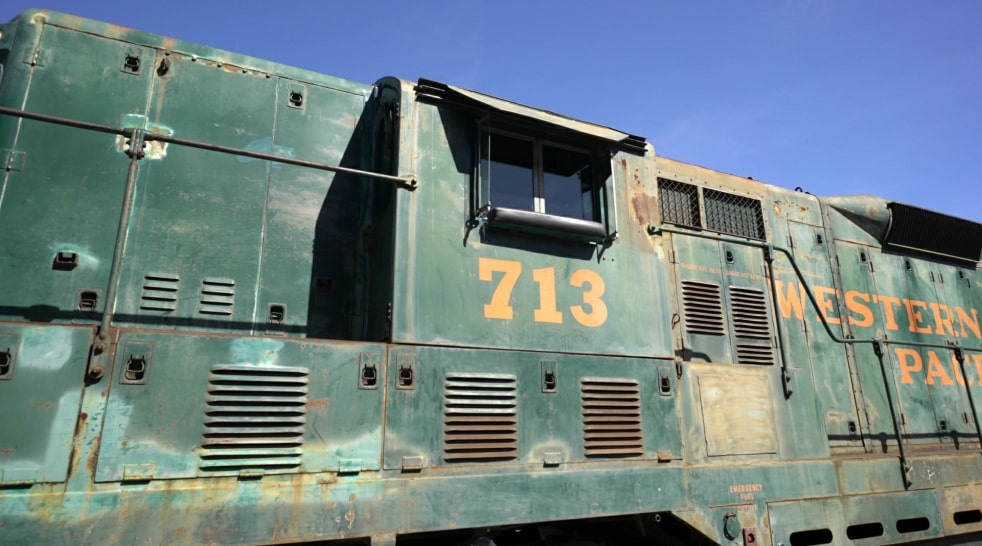}
  \plotzoom{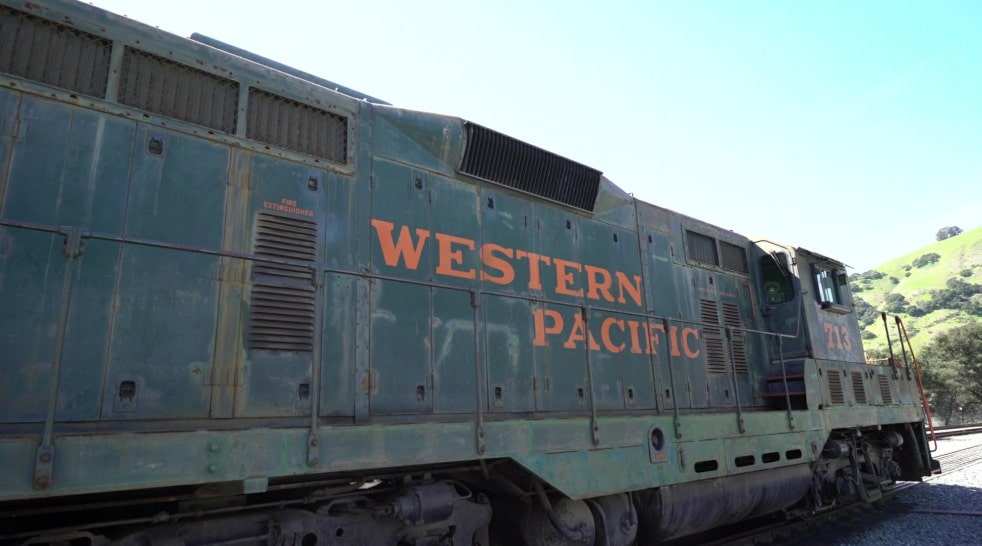}
  \plotzoom{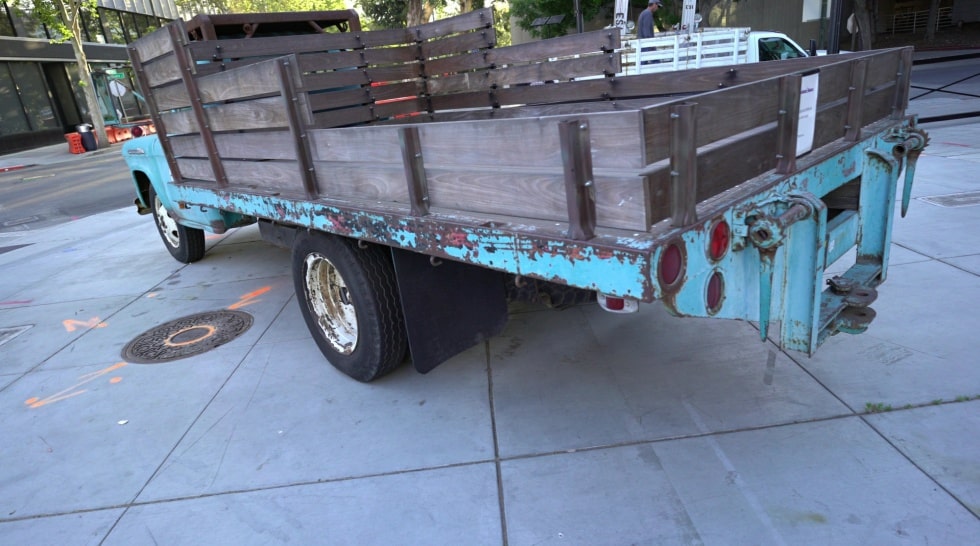}
  \caption{Ground Truth}
\end{subfigure}
  \begin{subfigure}[b]{\thirdwidth}
     \plotzoom{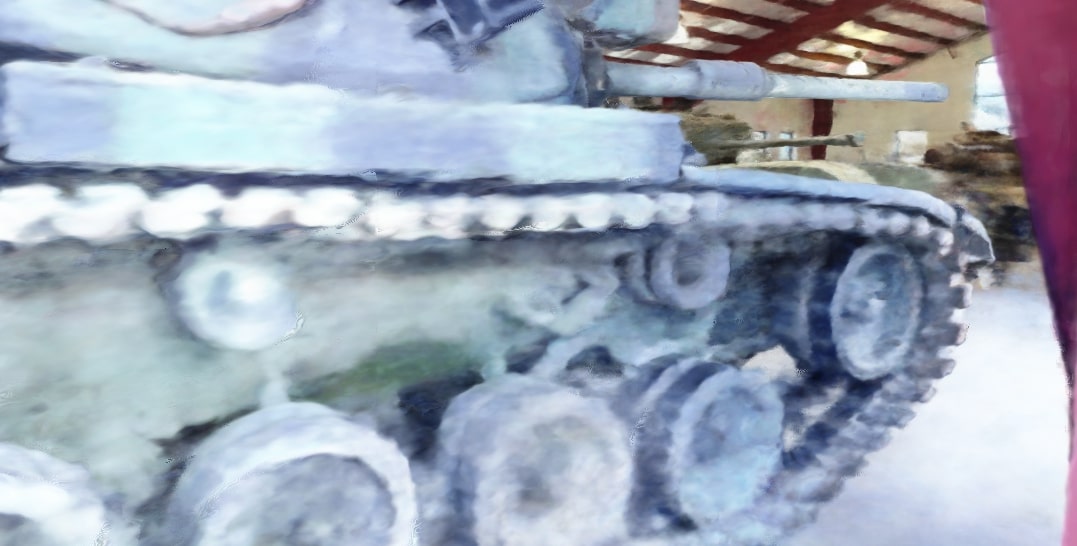}
     \plotzoom{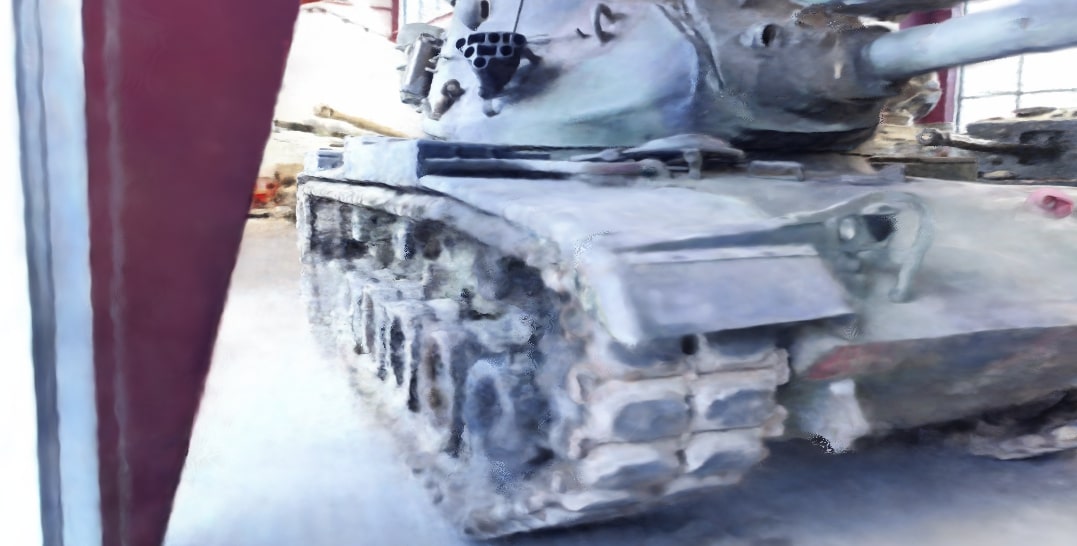}
   \plotzoom{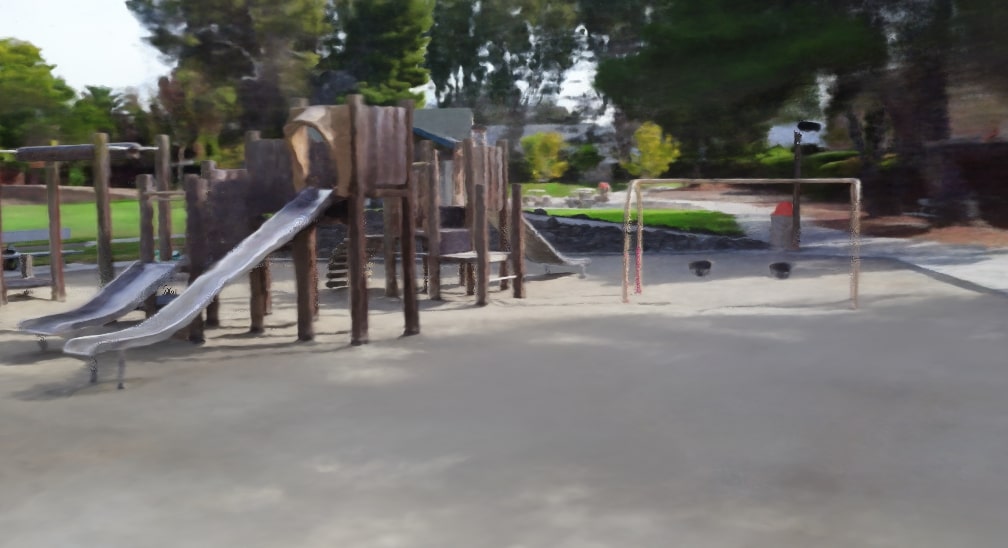}
   \plotzoom{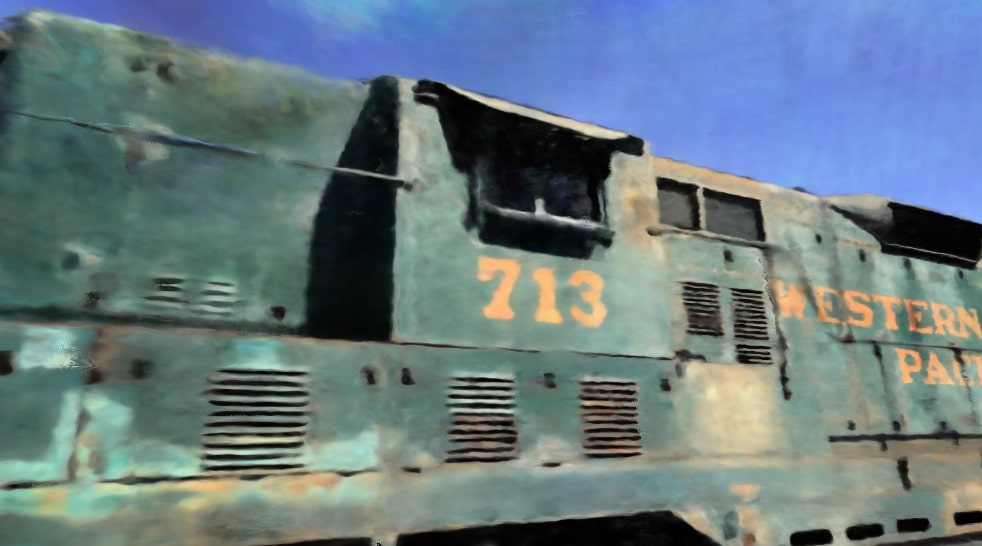}
   \plotzoom{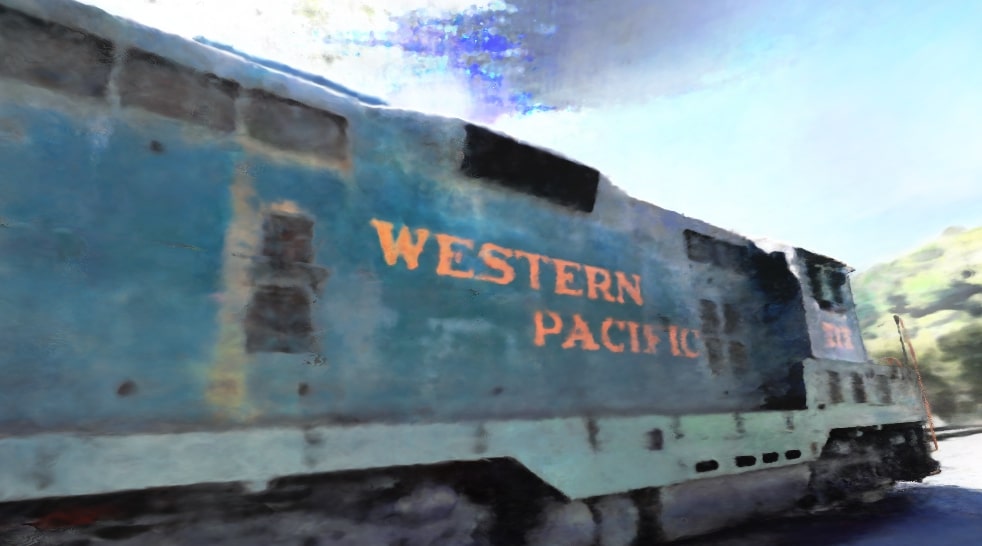}
   \plothalf{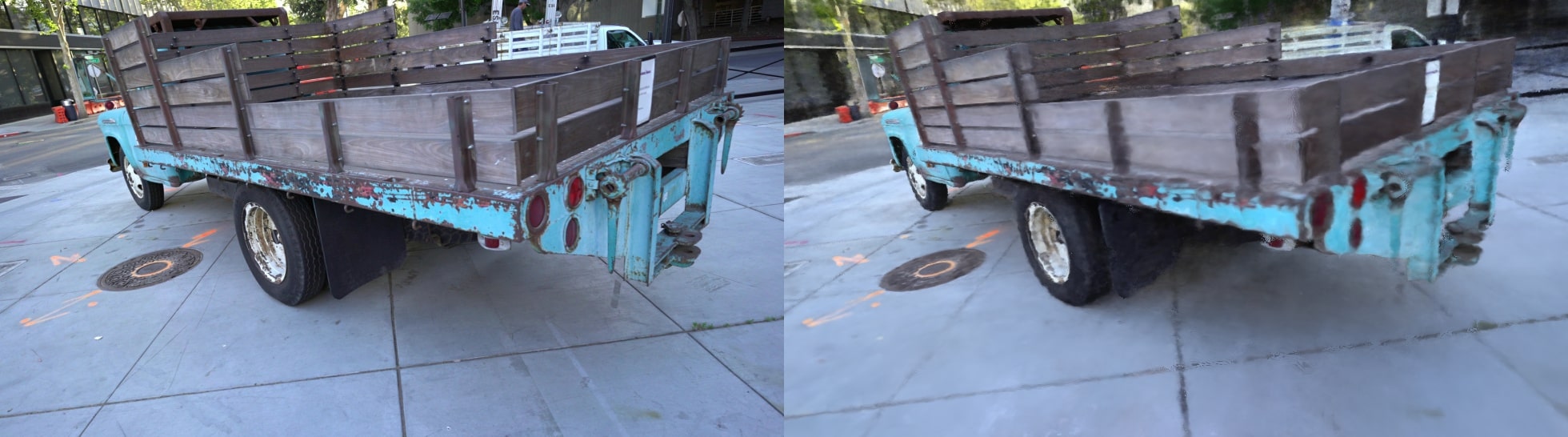}
  \caption{NeRF++}
\end{subfigure}
  \begin{subfigure}[b]{\thirdwidth}
    \plotzoom{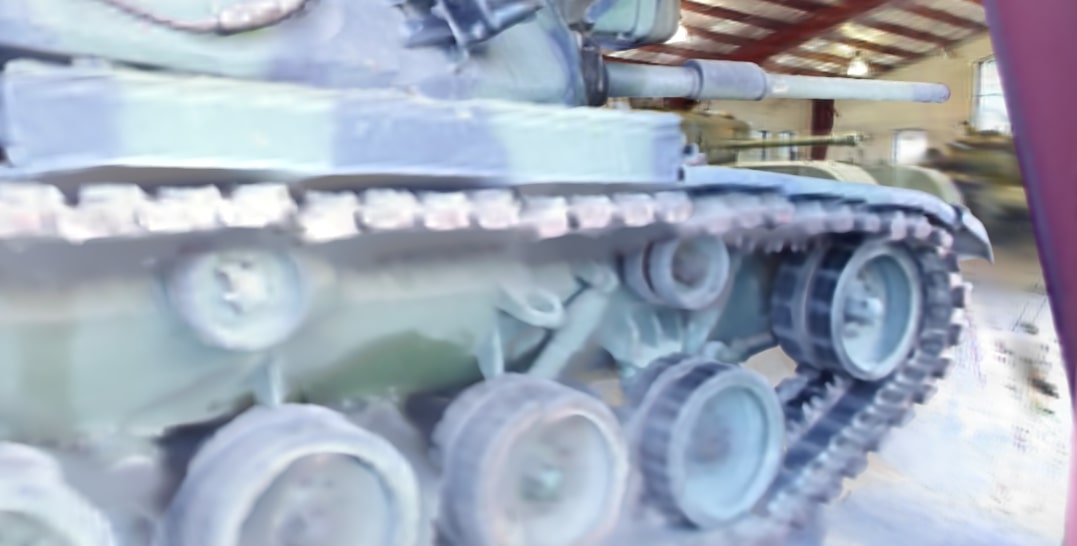}
    \plotzoom{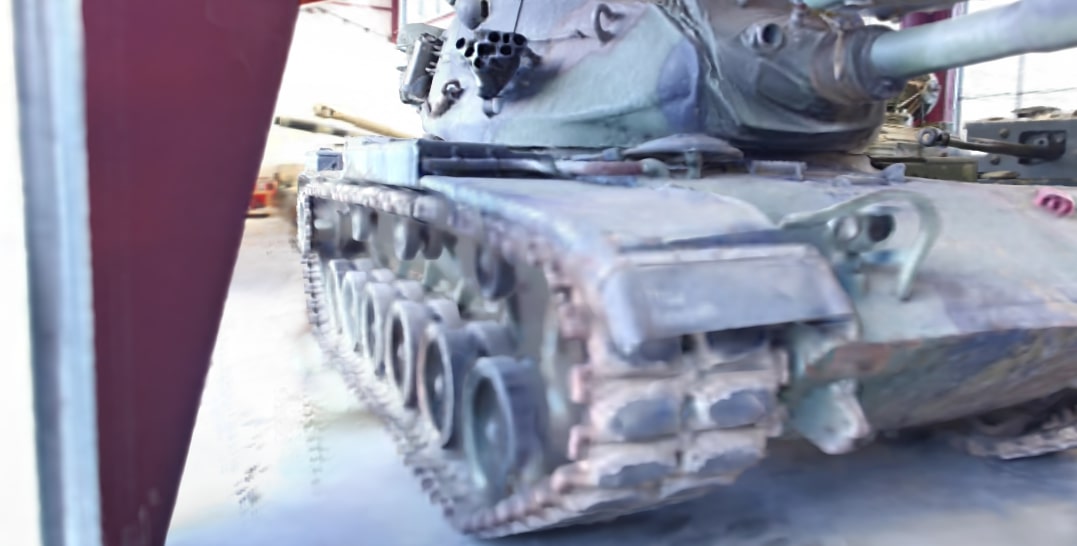}
  \plotzoom{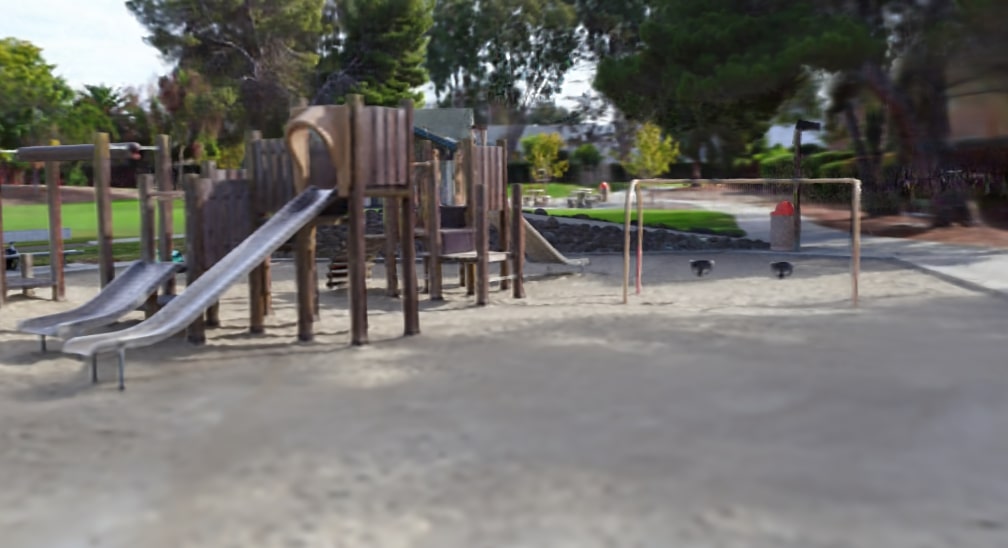}
  \plotzoom{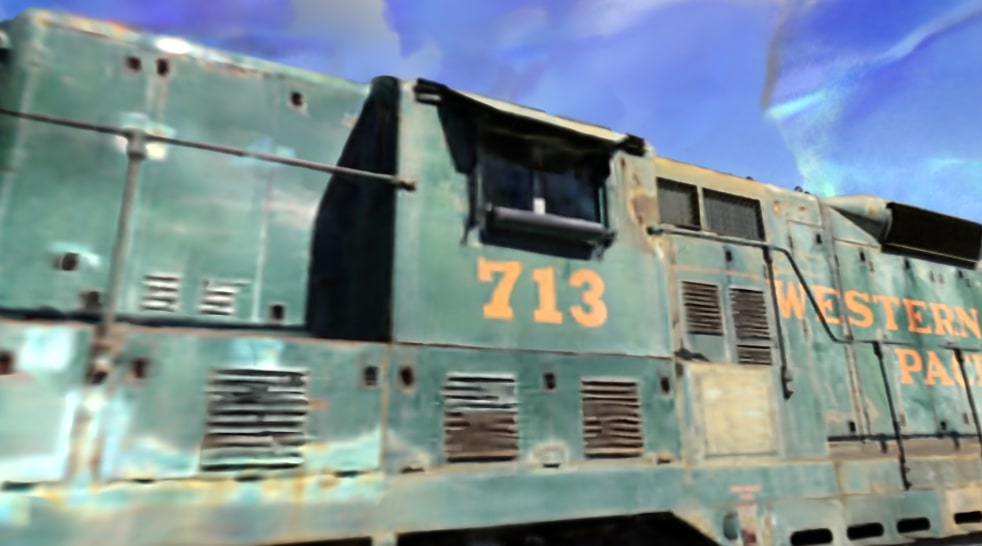}
  \plotzoom{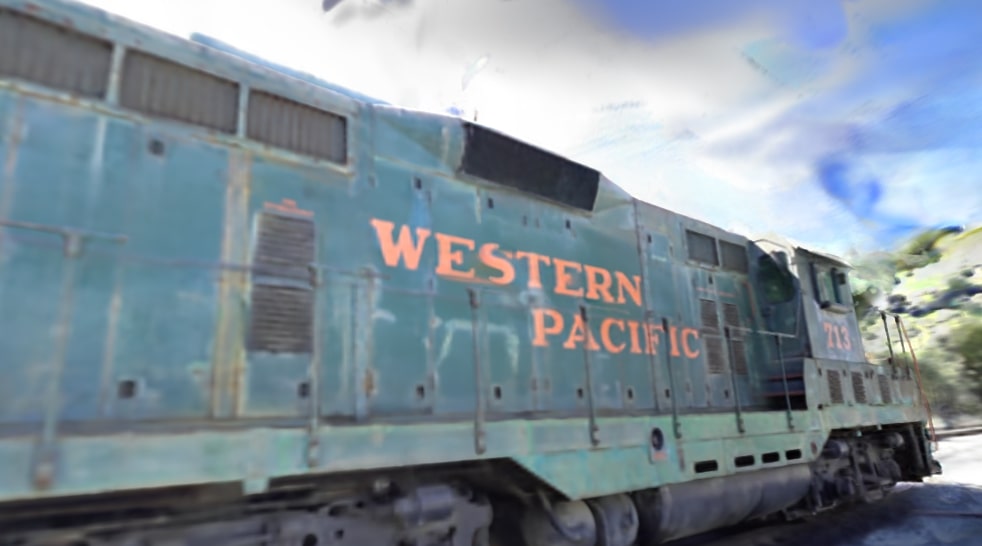}
  \plotzoom{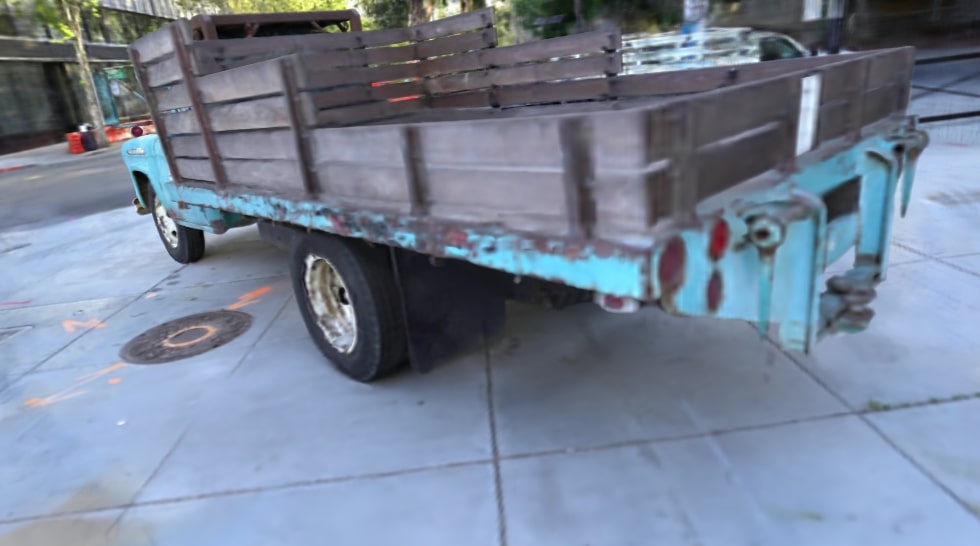}
  \caption{Plenoxels}
\end{subfigure}
\caption{\textbf{$360^\circ$ scenes.} We show a random view from each of the Tanks and Temples scenes, comparing the ground truth, NeRF++~\cite{zhang2020nerf}, and our Plenoxels. We include two random views each for the M60 and train scenes, since the playground and truck scenes were shown in the main text.}
  \label{fig:full360}
\end{figure*}